\documentclass[letterpaper]{article} 
\usepackage{aaai2026}  
\usepackage{times}  
\usepackage{helvet}  
\usepackage{courier}  
\usepackage[hyphens]{url}  
\usepackage{graphicx} 
\usepackage{natbib}  
\usepackage{caption} 
\usepackage{amsmath}
\usepackage{amssymb}
\usepackage{amsthm}
\usepackage{booktabs}
\usepackage[capitalize,noabbrev]{cleveref}
\usepackage{comment}
\usepackage{enumitem}
\usepackage{fontawesome5}
\usepackage{ifthen}
\usepackage{longtable}
\usepackage{multirow}
\usepackage{pifont}
\usepackage[group-separator={,},group-minimum-digits={3}]{siunitx}
\usepackage{stackengine}
\usepackage{subcaption}
\usepackage{tikz}
\usetikzlibrary{arrows.meta,automata,calc,fit,matrix,patterns,positioning,shadows,shapes.arrows,shapes.geometric,svg.path}
\usepackage{xspace}

\newcommand{\realnums}{\mathbb{R}}
\newcommand{\simplex}[1]{\Delta(#1)}
\newcommand{\expect}{\mathbb{E}}
\newcommand{\card}[1]{|#1|}
\newcommand{\concat}{\oplus}
\newcommand{\normaldist}{\mathcal{N}}
\newcommand{\variance}{\sigma^2}
\newcommand{\unidist}{\mathcal{U}}

\newcommand{\mdp}{\mathcal{M}}
\newcommand{\obs}{O}
\newcommand{\actions}{A}
\newcommand{\paramset}{\Theta}
\newcommand{\states}{S}
\newcommand{\transitionfn}{\mathcal{T}}
\newcommand{\obsfn}{\mathcal{I}}
\newcommand{\rewardfn}{\mathcal{R}}
\newcommand{\disc}{\gamma}
\newcommand{\gae}{\lambda}
\newcommand{\params}{\theta}
\newcommand{\policy}{\pi}
\newcommand{\policyopt}{\policy^\ast}
\newcommand{\valuefn}{V}
\newcommand{\reward}{r}
\newcommand{\horizon}{T}
\newcommand{\rmax}{R_{\max}}

\newcommand{\rmname}{M}
\newcommand{\rmstates}{U}
\newcommand{\propset}{P}
\newcommand{\rmtxstate}{\delta}
\newcommand{\rmtxreward}{\mathcal{R}}
\newcommand{\rmstate}{u}
\newcommand{\rmidx}[1]{\rmstate_{#1}}
\newcommand{\rminit}{\rmidx{0}}
\newcommand{\rmacc}{\rmidx{A}}
\newcommand{\proppowerset}{2^\propset}
\newcommand{\labfunc}{\mathcal{L}}
\newcommand{\proplabel}{L}

\newcommand{\propstr}[1]{\mathtt{#1}}
\newcommand{\front}{\propstr{front}}
\newcommand{\carrying}{\propstr{carrying}}
\newcommand{\nextto}{\propstr{next}}
\newcommand{\obj}{\propstr{o}}
\newcommand{\ball}{\propstr{ball}}
\newcommand{\key}{\propstr{key}}
\newcommand{\squareobj}{\propstr{square}}
\newcommand{\door}{\propstr{door}}
\newcommand{\objcol}{\propstr{c}}
\newcommand{\red}{\propstr{red}}
\newcommand{\green}{\propstr{green}}
\newcommand{\blue}{\propstr{blue}}
\newcommand{\purple}{\propstr{purple}}
\newcommand{\yellow}{\propstr{yellow}}
\newcommand{\gray}{\propstr{gray}}
\newcommand{\objstate}{\propstr{s}}
\newcommand{\open}{\propstr{open}}
\newcommand{\closed}{\propstr{closed}}
\newcommand{\locked}{\propstr{locked}}

\newcommand{\graph}{G}
\newcommand{\node}{v}
\newcommand{\nodes}{\mathcal{V}}

\newcommand{\edges}{\mathcal{E}}
\newcommand{\edgefeat}[2]{\mathbf{e}_{#1,#2}}
\newcommand{\nodefeat}{\mathbf{h}}
\newcommand{\nodefeatl}[2]{\mathbf{h}^{(#1)}_{#2}}
\newcommand{\layer}{l}
\newcommand{\numlayers}{L}
\newcommand{\activ}{\sigma}
\newcommand{\layerweights}[1]{\mathbf{W}^{(#1)}}
\newcommand{\layerweightsself}[1]{\layerweights{#1}_\textnormal{self}}
\newcommand{\layerweightsneigh}[1]{\layerweights{#1}_\textnormal{neigh}}
\newcommand{\lastweights}{\mathbf{W}_\textnormal{last}}
\newcommand{\neigh}[1]{\mathfrak{N}(#1)}
\newcommand{\identity}{\mathbf{I}}
\newcommand{\numnodefeats}{n}
\newcommand{\numedgefeats}{m}

\newcommand{\addrooms}{\textsc{AddRooms}}
\newcommand{\rmrooms}{\textsc{RemoveRooms}}
\newcommand{\addobj}{\textsc{AddObject}}
\newcommand{\rmobj}{\textsc{RemoveObject}}
\newcommand{\mvagent}{\textsc{MoveAgent}}
\newcommand{\mvobj}{\textsc{MoveObject}}
\newcommand{\rpdoor}{\textsc{ReplaceDoor}}
\newcommand{\rpnondoor}{\textsc{ReplaceNonDoor}}
\newcommand{\switchprop}{\textsc{SwitchProposition}}
\newcommand{\addstate}{\textsc{AddState}}
\newcommand{\remstate}{\textsc{RemoveState}}
\newcommand{\hindpred}{\textsc{ExtractPreceding}}
\newcommand{\hindsucc}{\textsc{ExtractSucceeding}}

\newcommand{\rwhstruct}{T}
\newcommand{\rwhstructv}{\mathcal{V}_\rwhstruct}
\newcommand{\rwhstructe}{\mathcal{E}_\rwhstruct}
\newcommand{\rwnumrms}{m}
\newcommand{\rwlocalrm}{G}
\newcommand{\rwlocalrmv}{\mathcal{V}}
\newcommand{\rwlocalrme}{\mathcal{E}}
\newcommand{\rwcalllabfun}{\mathcal{C}}
\newcommand{\rwedlabfun}{\mathcal{P}}
\newcommand{\rwrmtransition}{\mathbf{M}}
\newcommand{\rwavgconnect}{\bar{k}}
\newcommand{\rwconstmatrix}{\mathbf{C}}
\newcommand{\rwoutlabelings}{\mathbf{p}_\text{out}}
\newcommand{\rwinlabelings}{\mathbf{p}_\text{in}}
\newcommand{\rwnumouttrans}{l}
\newcommand{\rwnumintrans}{k}
\newcommand{\rwcatprior}{\boldsymbol{\rho}_\textnormal{prop}}
\newcommand{\rwcallcompat}{\mathbf{K}}
\newcommand{\rwcallcatprior}{\boldsymbol{\rho}_\textnormal{call}}
\newcommand{\rwinittransmatrix}{\mathbf{P}_\textnormal{init}}

\tikzset{
	in place/.style={
		auto=false,
		fill=white,
		inner sep=2pt,
	},
	cascaded/.style = {%
		general shadow = {%
			shadow scale = 1,
			shadow xshift = -1ex,
			shadow yshift = 1ex,
			draw,
			fill = white},
		general shadow = {%
			shadow scale = 1,
			shadow xshift = -.5ex,
			shadow yshift = .5ex,
			draw,
			fill = white},
		fill = white, 
		draw,
		minimum width = 1.5cm,
		minimum height = 2cm
	},
	none/.style = {
		inner sep=0mm
	},
	every loop/.append style={-Latex}
}

\newcommand{\rplr}{$\textnormal{PLR}^\bot$\xspace}
\newcommand{\methodfullname}{Aligning Tasks and Levels for Autocurricula of Specifications}
\newcommand{\methodabbrv}{\textsc{ATLAS}}
\newcommand{\ourmethod}[1]{%
	\ifthenelse{\equal{#1}{full}}{\methodfullname}{}%
	\ifthenelse{\equal{#1}{abbrv}}{\methodabbrv}{}%
	\ifthenelse{\equal{#1}{both}}{\methodabbrv{} (\methodfullname{})}{}%
}
\newcommand{\ourmethoddr}{DR\xspace}
\newcommand{\ourmethodrplr}{\rplr}
\newcommand{\ourmethodaccel}{ACCEL\xspace}
\newcommand{\ourmethodaccelzero}{ACCEL-0\xspace}

\theoremstyle{definition}

\newtheorem{example}{Example} 
\theoremstyle{remark}

\newcommand{\cmark}{\ding{51}}  
\newcommand{\xmark}{\ding{55}}  

\definecolor{icred}{HTML}{e64545}
\definecolor{icgreen}{HTML}{45e645}
\definecolor{icblue}{HTML}{4545e6}
\definecolor{icpurple}{HTML}{8b4cbf}
\definecolor{icyellow}{HTML}{f0c040}
\definecolor{icgray}{HTML}{646464}  
\DeclareRobustCommand{\iccircle}{\scriptsize\faCircle}
\DeclareRobustCommand{\icsquare}{\scriptsize\faSquareFull}
\DeclareRobustCommand{\ickey}{\scriptsize\faKey}
\DeclareRobustCommand{\icdooropen}{\scriptsize\faDoorOpen}
\DeclareRobustCommand{\icdoorclosed}{\scriptsize\faDoorClosed}
\DeclareRobustCommand{\icdoor}{
	\tikz[baseline=-0.0ex, trim left=-1.5pt]{%
		\node[inner sep=0pt, outer sep=0pt, anchor=base] (door) {\icdoorclosed};
		\node[inner sep=0pt, font=\scriptsize\bfseries, color=white, yshift=0.5pt, overlay] at (door.center) {?};
	}%
}
\DeclareRobustCommand{\icdoorlocked}{%
	\tikz[baseline=-0.0ex, inner sep=0pt, outer sep=0pt]{%
		\node[inner sep=0pt, outer sep=0pt, anchor=base] (door) {\icdoorclosed};
		\useasboundingbox (door.south west) rectangle (door.north east);
		\node[inner sep=0pt, font=\tiny\bfseries, color=white, xshift=1pt, yshift=0.25pt] at (door.center) {\scalebox{0.4}{\faLock}};
	}%
}
\DeclareRobustCommand{\iccarrying}[1]{
	\tikz[baseline=-0.0ex, inner sep=0pt, outer sep=0pt]{%
		\node[inner sep=0pt] (base) {\small\faHandHolding};
		\node[inner sep=0pt, yshift=6.5pt] at (base.center) {\scriptsize{#1}};
	}%
}
\DeclareRobustCommand{\icnexthor}[2]{{#1}\,{#2}}
\DeclareRobustCommand{\icnext}[2]{\shortstack{\makebox[1em]{#1}\\\makebox[1em]{#2}}}
\DeclareRobustCommand{\iccolor}[2]{{\color{#2}#1}}
\DeclareRobustCommand{\icnocolor}[1]{%
	\stackinset{c}{}{c}{}{%
		\tikz{\fill[pattern=north east lines, pattern color=white, opacity=1.0] (0,0) rectangle (0.65em,0.65em);}%
	}{#1}%
}

\newif\ifaaai
\newif\ifarxiv
\aaaifalse 
\ifaaai
\arxivfalse
\else
\arxivtrue
\fi

\newcommand{\appref}[2]{%
	\ifarxiv
	\cref{#1}
	\else
	Appendix~#2
	\fi
}

\title{Beyond Fixed Tasks: Unsupervised Environment Design for Task-Level Pairs}
\author {
    Daniel Furelos-Blanco\textsuperscript{\rm 1}, 
    Charles Pert\textsuperscript{\rm 1},
    Frederik Kelbel\textsuperscript{\rm 1},
    Alex F. Spies\textsuperscript{\rm 1},\\
    Alessandra Russo\textsuperscript{\rm 1},
    Michael Dennis\textsuperscript{\rm 2}\thanks{Contributed in an advisory capacity.}
}
\affiliations {
    \textsuperscript{\rm 1}Imperial College London\\
    \textsuperscript{\rm 2}Google DeepMind\\
    \{d.furelos-blanco18, c.pert22, frederik.kelbel20, afspies, a.russo\}@imperial.ac.uk, dennismi@google.com
}

\begin{document}

\maketitle

\begin{abstract}
	Training general agents to follow complex instructions (\emph{tasks}) in intricate environments (\emph{levels}) remains a core challenge in reinforcement learning. Random sampling of task-level pairs often produces unsolvable combinations, highlighting the need to co-design tasks and levels. While unsupervised environment design~(UED) has proven effective at automatically designing level curricula, prior work has only considered a fixed task. We present \ourmethod{both}, a novel method that generates \emph{joint autocurricula} over tasks and levels. Our approach builds upon UED to automatically produce solvable yet challenging task-level pairs for policy training. To evaluate \ourmethod{abbrv} and drive progress in the field, we introduce an \emph{evaluation suite} that models tasks as reward machines in Minigrid levels. Experiments demonstrate that \ourmethod{abbrv} vastly outperforms random sampling approaches, particularly when sampling solvable pairs is unlikely. We further show that mutations leveraging the structure of both tasks and levels accelerate convergence to performant policies.
\end{abstract}

\begin{links}
    \link{Code}{https://github.com/spike-imperial/atlas}
    \ifaaai
    \link{Extended version}{https://arxiv.org/abs/2511.12706}
    \fi
\end{links}

\section{Introduction}
Training generally-capable agents that follow diverse instructions (\emph{tasks}) in varied environments (\emph{levels}) is a central challenge in reinforcement learning~\cite{OELTeam21}. This dual complexity emerges across domains---from cooking agents executing recipes in different kitchens, to navigation agents following directives through unfamiliar cities. To generate this open-ended complexity, methods like EUREKA~\cite{MaLWHBJZFA24}, OMNI-EPIC~\cite{FaldorZCC25}, and others~\cite{ZhangLSC24,KlissarovHRSVZBPMO25} have often relied on LLM-driven code-generation.

However, there has been growing interest in an alternative approach of expressing tasks through \emph{formal languages}, including temporal logics~\cite{KuoKB20,VaezipoorLIM21,QiuMZ23,JackermeierA25} and finite-state machines~\cite{YalcinkayaLVS23,YalcinkayaLVS24}. Unlike natural language~\cite{LuketinaNFFAGWR19}, formal languages offer unambiguous semantics and precise progress tracking, making them well-suited for generalization across task-level pairs. These approaches typically train agents by sampling from uninformed task-level distributions, a strategy known as \emph{domain randomization} \cite[DR;][]{SadeghiL17,TobinFRSZA17}.

DR can succeed when sampled task-level pairs are mostly solvable. However, as the number of task-level combinations grows, the proportion of solvable pairs tends to decrease. Even among solvable pairs, DR generates samples of \emph{arbitrary difficulty}, causing unsolvable or overly challenging levels to dominate training. This raises the question: \emph{How can we train agents effectively from solvable yet appropriately challenging task-level pairs?}

Unsupervised environment design~\cite[UED;][]{DennisJVBRCL20}  has demonstrated success in automatically generating \emph{level} curricula, producing solvable and progressively more challenging levels that enable effective generalization. However, existing UED methods only generate curricula over levels for a \emph{fixed task}, yet general agents require both.

We introduce \ourmethod{both}, which extends UED's principled curriculum generation to both tasks and levels, ensuring that agents train on solvable yet appropriately challenging pairs. We express tasks as \emph{reward machines}~\cite[RMs;][]{ToroIcarteKVM18}---finite-state machines that compactly represent reward functions. Our \emph{contributions} include:
\begin{description}
	\item[Joint task-level autocurricula] \ourmethod{abbrv} co-designs tasks and levels to generate aligned curricula, ensuring solvable yet challenging pairs. A policy network, conditioned on graph embeddings of the RMs, learns effectively from the resulting curriculum.
	\item[Structure-aware task mutations] We leverage RM structure to guide task mutations while co-evolving levels, achieving faster learning than approaches that solely curate solvable but challenging random samples.
	\item[Evaluation] We introduce an evaluation suite combining RM tasks with Minigrid levels~\cite{Chevalier-Boisvert23} that, unlike previous work, explicitly targets settings where task-level pairs are rarely solvable.
\end{description}

Our \emph{experiments} show that \ourmethod{abbrv} generates joint task-level curricula that enable agents to master increasingly complex levels (with more rooms and objects) and tasks (with more RM states). By prioritizing solvable pairs at the frontier of agent capability, \ourmethod{abbrv} consistently outperforms DR across diverse test instances. Interestingly, forming curricula by repeatedly mutating tasks and levels from a simple starting point can result in policies with comparable performance to those based on random sampling alone.

\section{Background}
\label{sec:background}
We review unsupervised environment design and the use of reward machines in reinforcement learning.

\subsection{Unsupervised Environment Design}
\label{sec:ued}
Unsupervised environment design~\cite[UED;][]{DennisJVBRCL20} aims to generate training environments that adapt to an agent's capabilities, inducing an \emph{autocurriculum}~\cite{LeiboHLG19,NarvekarPLSTS20,PortelasCWHO20}. The key is to \emph{parameterize} RL environments (e.g.,~grid size) and adapt these parameters during training. The goal is to train agents that generalize across all possible parameterizations.

Formally, UED problems are \emph{underspecified partially observable Markov decision processes}~\cite[UPOMDPs;][]{DennisJVBRCL20}. A standard POMDP is a tuple $\mdp=\langle\actions, \obs,\states,\transitionfn,\obsfn,\rewardfn,\disc\rangle$ where $\actions$ is a set of actions, $\obs$ is a set of observations, $\states$ is a set of states, $\transitionfn:\states\times \actions \to \simplex{\states}$ is the transition function, $\obsfn:\states\to\obs$ is the observation function, $\rewardfn:\states\to\realnums$ is the reward function, and $\disc$ is the discount factor. UPOMDPs augment POMDPs with a \emph{parameter space} $\paramset$, which represents the space of all environment configurations. Each parameterization $\params\in\paramset$ instantiates a fully specified POMDP $\mdp^\params$, often called \emph{level}~\cite{CobbeHHS20,JiangGR21}. The expected discounted return (or \emph{value}) of a \emph{policy} $\policy$ in level $\mdp^\params$ is $\valuefn^\params(\policy)=\expect[\sum^\horizon_{t=0}\disc^t\reward_t]$, where $\horizon$ is a horizon. The goal is to produce a sequence of distributions over $\paramset$ that maximizes the policy's value across levels.

\subsection{Methods for UED}
\label{sec:ued_methods}
UED methods often frame curriculum design as a game between two players: a \emph{teacher}, which proposes levels, and a \emph{student}, which trains on them. The teacher generates levels by maximizing a \emph{utility score}. We review two scoring strategies: \emph{constant} and \emph{regret-based} scoring.

\subsubsection{Constant.}
All levels have equal utility, ignoring the student's performance. Parameters $\params$ are uniformly sampled from the parameter space $\paramset$. This strategy is commonly known as \emph{domain randomization}~\cite[DR;][]{SadeghiL17,TobinFRSZA17}.

\subsubsection{Regret-Based.}
The regret for a level $\mdp^\params$ is  the value gap $\valuefn^\params(\policyopt)-\valuefn^\params(\policy)$ between the optimal policy $\policyopt$ and the current policy $\policy$. Unlike DR, regret scoring deprioritizes unsolvable levels and focuses on those at the frontier of the agent's capabilities. Theoretically, a teacher that maximizes regret induces a minimax regret student policy at equilibrium~\cite{DennisJVBRCL20}.

Since finding the optimal policy is intractable, UED methods resort to regret approximations. For example, MaxMC~\cite{JiangDPFGR21} uses the highest undiscounted return seen so far, $\rmax^\params$, as a proxy for optimal performance in $\mdp^\params$ and computes regret as $(1/\horizon)\sum^\horizon_{t=0}\rmax^\params-\valuefn^\params(o_t)$, where $\valuefn^\params(o_t)$ is the value of the observation at time $t$.

\emph{Prioritized Level Replay}~\cite[PLR;][]{JiangGR21} curates a \emph{buffer} of high-regret levels. At each episode, PLR chooses between (i)~with probability $p$, replaying levels from the buffer and updating their estimates, and (ii)~with probability $1-p$, sampling new levels via DR and adding them to the buffer. In both cases, the student policy trains on the selected levels.

In this paper, we consider two extensions of PLR. \emph{Robust PLR}~\cite[\rplr;][]{JiangDPFGR21} trains the student only on replayed levels. \emph{ACCEL}~\cite{Parker-HolderJ022} extends \rplr by \emph{mutating} the last replayed levels (e.g., moving an object) before adding them to the buffer, thereby exploring the neighborhood of high‑regret levels rather than solely relying on random sampling.

\subsection{Reward Machines}
\label{sec:rms}
Reward machines~\cite[RMs;][]{ToroIcarteKVM18} are finite-state machines for specifying reward functions using high-level propositional events. They provide a flexible task representation, supporting derivations from other formal languages~\cite{CamachoIKVM19} and mappings from natural language~\cite{TuliLVKSM22,LiuYILSTS23,LiKWAM25}.

Formally, an RM is a tuple $\langle\rmstates,\propset,\rmtxstate,\rmtxreward,\rminit,\rmacc \rangle$, where $\rmstates$ is a finite set of states; $\propset$ is a finite set of propositions that forms the \emph{alphabet} of the RM; $\rmtxstate:\rmstates\times\proppowerset\to\rmstates$ is the state-transition function, which maps an RM state and a subset of propositions into an RM state; $\rmtxreward:\rmstates\times\rmstates\to\realnums$ is the reward-transition function, which maps an RM state pair into a reward; $\rminit\in\rmstates$ is the initial state of the RM; and $\rmacc\in\rmstates$ is the accepting state of the RM, which represents the successful completion of the task.

We refer to proposition sets $\proplabel\in\proppowerset$ as \emph{labels}. These are produced by a \emph{labeling function} $\labfunc:\obs\to\proppowerset$ from environment observations. Given an RM state $\rmstate$ and a label $\proplabel$, the RM transitions to $\rmstate'=\rmtxstate(\rmstate,\proplabel)$ and emits reward $\rmtxreward(\rmstate, \rmstate')$.

\section{Task-Level Generalization Framework}
\label{sec:evaluation_suite}
We introduce a framework for problem-conditioned RL, where a \emph{problem specification} (hereafter, \emph{problem}) is a tuple consisting of a \emph{task}, an instruction the agent must follow, and a \emph{level}, the environment instance in which the agent acts.

We explore two key \emph{questions} within this framework. First, how agents generalize from \emph{rarely solvable} task-level pairs---a critical but understudied regime. Such pairs arise when tasks are infeasible in a level (e.g., opening a red door in a level with no doors). Second, how to extend UED beyond fixed tasks to settings where tasks and levels are \emph{co-designed}.

We instantiate this framework using RMs for tasks and Minigrid environments~\cite{Chevalier-Boisvert23} for levels, with samplers that generate problems in both rarely solvable and frequently solvable regimes. We also contribute 150 hand-designed problems that test capabilities such as long-term planning, implicit subgoals, and exploration. See \appref{app:evaluation_suite}{A} for details.

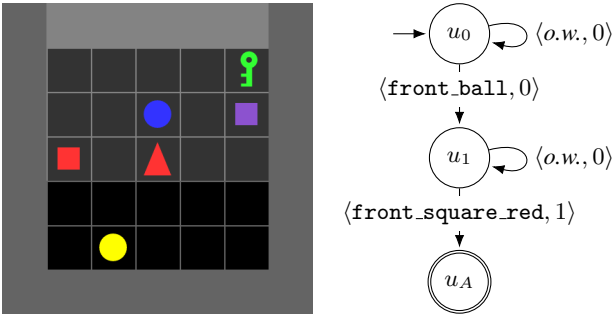
\begin{figure}
	\begin{subfigure}{0.49\linewidth}
		\centering
		\includegraphics[width=\linewidth]{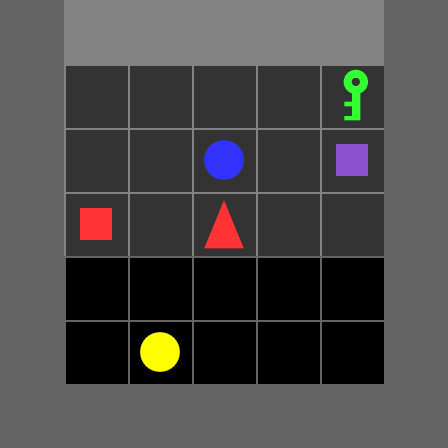}
	\end{subfigure}
	\hfill
	\begin{subfigure}{0.49\linewidth}
		\centering
		\begin{tikzpicture}[shorten >=1pt,node distance=1.65cm,on grid,auto,every initial by arrow/.style ={-Latex},state/.style={circle,draw,minimum size=0.8cm,inner sep=0pt}]
			\node[state,initial,initial text=] (u_0)   {\small$\rminit$};
			\node[state] (u_1) [below = of u_0]   {\small$\rmidx{1}$};
			\node[state,accepting] (u_acc) [below = of u_1]  {\small$\rmacc$};
			
			\path[-Latex] (u_0) edge [loop right] node {\small$\langle \textit{o.w.}, 0\rangle$} ();
			\path[-Latex] (u_1) edge [loop right] node {\small$\langle \textit{o.w.}, 0\rangle$} ();
			\path[-Latex] (u_0) edge node[in place, pos=0.4] {\small$\langle\front\_\ball,0\rangle$} (u_1);
			\path[-Latex] (u_1) edge node[in place, pos=0.4] {\small$\langle\front\_\squareobj\_\red,1\rangle$} (u_acc);
		\end{tikzpicture}
	\end{subfigure}
	\caption{A \emph{problem} consisting of a Minigrid \emph{level} and an RM \emph{task} for \emph{``go to a ball, then go to a red square''}.}
	\label{fig:rm_example}
\end{figure}

\subsection{Levels}
\label{sec:evaluation_suite_levels}
We instantiate \emph{levels} as Minigrid environments~\cite{Chevalier-Boisvert23}. A level consists of a grid containing \emph{objects} (keys, squares, balls, doors) with different \emph{colors} (red, green, blue, purple, yellow, gray). Doors have three \emph{states}: locked, open, or closed, with locked doors requiring keys of matching color to open. The agent observes a $5\times5$ region in front of it.

\begin{example}
	\cref{fig:rm_example} (left) shows a Minigrid level with the agent (red triangle) and its observation (highlighted area).
\end{example}

\subsubsection{Sampling.}
We develop samplers that generate levels with a random number of rooms and objects. All objects are randomly determined. Agent and non-door objects are randomly placed. Rooms are connected via doors.

\subsection{Tasks}
\label{sec:evaluation_suite_tasks}
\emph{Tasks} are instantiated as RMs that encode BabyAI instructions~\cite{Chevalier-BoisvertBLWSNB19}. These instructions consist of \emph{go to}, \emph{open}, \emph{pick up}, and \emph{put next} commands, each applied to specific objects, optionally conditioned on color and state. The RM \emph{alphabet} is defined by mapping these commands to propositions:
\begin{itemize}
	\item $\front\_\langle\obj_1\rangle\_\langle\objcol_1\rangle\_\langle\objstate_1\rangle$ indicates the agent is in front of object $\obj_1$ with color $\objcol_1$ and state $\objstate_1$,
	\item $\carrying\_\langle\obj_1\rangle\_\langle\objcol_1\rangle\_\langle\objstate_1\rangle$ indicates the agent is carrying object $\obj_1$ with color $\objcol_1$ and state $\objstate_1$, and
	\item $\nextto\_\langle\obj_1\rangle\_\langle\objcol_1\rangle\_\langle\objstate_1\rangle\_\langle\obj_2\rangle\_\langle\objcol_2\rangle\_\langle\objstate_2\rangle$ indicates that object $\obj_1$ with color $\objcol_1$ and state $\objstate_1$ is next to object $\obj_2$ with color $\objcol_2$ and state $\objstate_2$ within the agent's visual field,
\end{itemize}
where $\obj_i\in\{\ball, \squareobj, \key,\door\}$, $\objcol_i\in\{\red,\allowbreak\green,\allowbreak\blue,\allowbreak\purple,\allowbreak\yellow,\allowbreak\gray,\allowbreak\epsilon\}$, and $\objstate_i\in\{\open,\allowbreak\closed,\allowbreak\locked,\allowbreak\epsilon\}$ for $i\in\{1,2\}$. The state $\objstate_i$ is unspecified ($\epsilon$) if $\obj_i\neq \door$. The propositions capture each instruction type: $\front$ for \emph{go to} and \emph{open}, $\carrying$ for \emph{pick up}, and $\nextto$ for \emph{put next}. Crucially, RMs enable sequencing and alternating formulas over these propositions, mirroring the connectors \emph{then} and \emph{and} in BabyAI instructions.

\begin{example}
	\cref{fig:rm_example}~(right) shows an RM for the instruction \emph{``go to a ball, then go to a red square''}. A reward of 1 is given upon completing the task, and 0 otherwise. Given the level on its left, the observation of the agent is mapped into the label below, which satisfies the transition from $\rminit$ to $\rmidx{1}$: 
	{\footnotesize
		\begin{align*}
			\left\lbrace\begin{matrix}
				\front\_\ball & \nextto\_\squareobj\_\purple\_\key\_\green\\
				\front\_\ball\_\blue & \nextto\_\squareobj\_\key\_\green\\
				\nextto\_\squareobj\_\key & \nextto\_\squareobj\_\purple\_\key
			\end{matrix}\right\rbrace.
		\end{align*}
	}
\end{example}

\subsubsection{Sampling.}
We introduce two RM sampling strategies:
\begin{itemize}
	\item \emph{Sequential.} Generates RMs with a single path from $\rminit$ to $\rmacc$, where the path length is sampled from a predefined range. \cref{fig:rm_example} is a length 2 example.
	\item \emph{Random Walk-Based.} Generates RMs with one or more (possibly cyclic) paths from $\rminit$ to $\rmacc$, subsuming the sequential case. Transitions are determined by a random walk over a uniformly initialized Markov transition matrix. See \appref{app:random_walk}{A.2} for details.
\end{itemize}
In both strategies, each transition is labeled with a proposition $p$ sampled uniformly from the alphabet, while its negation $\neg p$ is added to all other outgoing transitions from the same state to enforce \emph{determinism}.

Sampled RMs support a variety of \emph{reward functions}, including sparse rewards (1 only when reaching $\rmacc$), step-wise rewards \cite[1 on each transition to a new state, e.g.][]{TuliLVKSM22}, and shaped rewards based on the distance to $\rmacc$ \cite[e.g.,][]{CamachoCSM17,Furelos-BlancoLJBR21}.

\subsection{Problem Sampling}
\label{sec:evaluation_suite_sampling}
There are two main strategies to sample a problem:
\begin{description}
	\item[Independent.] Tasks and levels are sampled separately.
	\item[Level-Conditioned.] Task generation is constrained by the objects in the sampled level. Only propositions involving those objects may label the edges in the RM.
\end{description}
Level-conditioning increases the likelihood of sampling \emph{solvable} problems, but does not guarantee it. For example, a task may require unlocking a door with an existing but unreachable key. Guaranteeing solvability is challenging, as it may require solving the problem during generation.

\begin{figure*}
	\centering
	\includegraphics[width=\linewidth]{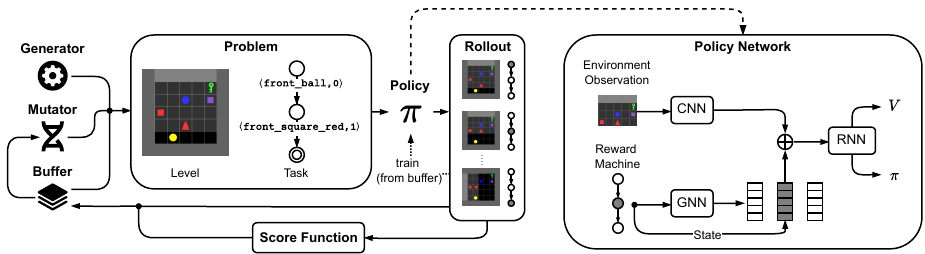}
	\caption{Overview of \ourmethod{abbrv} instantiated with \rplr and ACCEL. The UED loop \textbf{(left)} samples problems---i.e.,~task-level pairs---from either a generator or a buffer of high-regret problems. \ourmethodaccel provides problems that result from mutating selected buffer problems. The policy network \textbf{(right)} processes observations via a convolutional neural network (CNN) and RM tasks via a graph neural network (GNN). The GNN produces representations for all RM states. The current state's embedding is concatenated with the CNN features and passed through a recurrent neural network (RNN) to capture history. The resulting representation is used to generate actions and value estimates. Policy rollouts are used to train the network (for buffer-sourced problems), and to compute regret scores, which determine if new problems enter the buffer or update existing ones. Unlike \ourmethodrplr and \ourmethodaccel, \ourmethoddr trains policies only from problems produced by the generator.
	}
	\label{fig:method_overview}
\end{figure*}

\section{Problem-Conditioning via Autocurricula}
\label{sec:method}
We introduce \ourmethod{both}, a method for learning policies that \emph{generalize} across problem specifications by leveraging \emph{autocurricula}---automatically adapting the training problems to match the agent's capabilities. This is critical in settings where \emph{solvable} problems are rarely sampled.

Our approach extends UED beyond level generation to \emph{jointly} adapt over both tasks and levels. By co-designing these, regret-based UED methods generate problems that are not only solvable (i.e., the task is feasible within the corresponding level) but also challenging.

\cref{fig:method_overview} illustrates \ourmethod{abbrv}, which comprises two components: a problem-conditioned policy network and a UED-driven curriculum generation loop. We describe the components of \ourmethod{abbrv} in the following sections.

\subsection{Policy Architecture}
\label{sec:method_arch}
To generalize across task-level combinations, we use an actor-critic architecture conditioned on both environment observations and RM task states. The \emph{key} design choice is to encode RMs---labeled directed graphs---using a \emph{graph neural network}~(GNN). Unlike fixed embeddings (e.g.,~one-hot encodings of the RM state index), GNNs naturally handle varying RM topologies and edge labelings, supporting generalization across RMs. The policy is trained using PPO~\cite{SchulmanWDRK17}. See \appref{app:architecture}{C} for details.

\subsection{Problem Autocurricula via UED}
\label{sec:method_ued}
We extend UED approaches to support joint generalization over both tasks and levels. For brevity, we refer to \ourmethod{abbrv} instantiations of these methods as \ourmethoddr, \ourmethodrplr, and \ourmethodaccel, corresponding to the underlying UED approach. In the latter two, generalization is driven by automatically induced curricula over both levels and RM tasks.

Unlike the original ACCEL~\cite{Parker-HolderJ022}, which only applied level mutations, \ourmethod{abbrv}' instantiation incorporates \emph{task-aware mutations} leveraging the RM structure. This enables exploring a broader neighborhood of problems by modifying both tasks and levels.

A mutation consists of a sequence of edits, where the number of edits (sampled from a predefined range) and each edit type are selected uniformly at random. We define three \emph{types of edits} below (see \appref{app:mutations}{D} for examples):

\paragraph{Level Edits.} 
Edits applied to the environment. In Minigrid, this includes moving the agent, adding/removing rooms, and adding/removing/replacing/moving an object.

\paragraph{Task Edits.} 
Edits applied to the RM structure. These include switching a proposition and adding/removing a state.

\paragraph{Hindsight Edits.}
Inspired by \emph{hindsight relabeling} \cite{AndrychowiczCRS17}, these edits generate new subproblems from partial progress. When a rollout ends in an intermediate RM state $\rmstate$ (neither initial nor accepting), there are two possible edits:
\begin{itemize}
	\item \emph{Preceding edits.} Keep the original level but set $\rmstate$ as the accepting RM state, yielding a simpler problem.
	\item \emph{Succeeding edits.} Use the reached environment state as the new level and set $\rmstate$ as the initial RM state, creating a continuation problem.
\end{itemize}
Since these edits depend on the original problem, they can only be applied as the first step in a mutation sequence.

\section{Experiments}
\label{sec:experiments}
We address the following research questions:
\begin{enumerate}[label=\textbf{RQ\arabic*},itemsep=0.2em, parsep=0pt, leftmargin=2.5em] 
	\item\label{rq:curric_effectiveness}\textbf{Joint Curriculum Effectiveness.} Do approaches that generate curricula over both levels and tasks (\ourmethodrplr, \ourmethodaccel) outperform \ourmethoddr?
	\item\label{rq:mutation_benefits}\textbf{Mutation Benefits.} Does incorporating mutations (\ourmethodaccel) offer advantages over curation-only approaches (\ourmethodrplr)?
	\item\label{rq:curric_emergence}\textbf{Joint Curriculum Emergence.} Does \ourmethod{abbrv} induce autocurricula over both levels and tasks?
	\item\label{rq:mutation_analysis}\textbf{Mutation Analysis.} Which mutation types become most prevalent during training?
\end{enumerate}
We address these questions in our main results, supported by key ablations below and extended analysis in \appref{app:experiments}{E}.

\subsection{Experimental Setup}
\label{sec:experiments_setup}
We describe the \emph{default} training setup used in our experiments. Additional details are provided in \appref{app:experiments_setup}{E.2}.

\subsubsection{Problem Generation.}
Levels and tasks are sampled \emph{independently}, creating a challenging setting in which most sampled problems are not solvable. \emph{Levels} are generated by sampling (i)~a number of rooms in $\{1,2,4,6\}$, (ii)~a number of objects from a grid-dependent range, (iii)~the objects themselves, and (iv)~the agent position. \emph{Tasks} are specified as RMs generated via the sequential sampler, with path lengths randomly chosen between 1 and 5. Although our approach supports various reward functions (see \cref{sec:evaluation_suite_tasks}), we use sparse rewards---1 on transitions to $\rmacc$ and 0 otherwise---as they are general yet challenging, making them well-suited for assessing performance and curricula emergence.

\subsubsection{Algorithms.}
We evaluate \ourmethod{abbrv} instantiated with \ourmethoddr, \ourmethodrplr, and \ourmethodaccel. For \ourmethodaccel, we distinguish between (i)~\ourmethodaccel, where problems are sampled using the setup above; and (ii)~\ourmethodaccelzero, where sampled problems consist of single-room levels with one object and RMs with one transition, and complexity emerges solely through mutations. We apply the \emph{mutations} described in \cref{sec:method_ued}, with sequence lengths sampled uniformly in the range 7--10.

\subsubsection{Metrics.}
We evaluate performance using two complementary metrics, averaged over five seeds with 95\% confidence intervals. Each problem is evaluated 10 times per seed. First, we compute the \emph{conditional value at risk} \cite[CVaR;][]{RutherfordBWLHF24}, which quantifies \emph{robustness} by reporting the solve rate for the $\alpha\%$ worst-performing problems in a large sampled set. Second, we report \emph{test performance} using the aggregate \emph{inter-quartile mean}~\cite[IQM;][]{AgarwalSCCB21} of solve rates on our hand-designed evaluation set, assessing both in-distribution and out-of-distribution generalization.

\subsection{Main Results}
\label{sec:experiments_main_results}
We present our core experimental results addressing the research questions outlined earlier.

\subsubsection{Performance (\ref{rq:curric_effectiveness}, \ref{rq:mutation_benefits}).}
\begin{figure}
	\centering
	\begin{subfigure}[b]{\linewidth}
		\centering
		\includegraphics[width=0.7\linewidth]{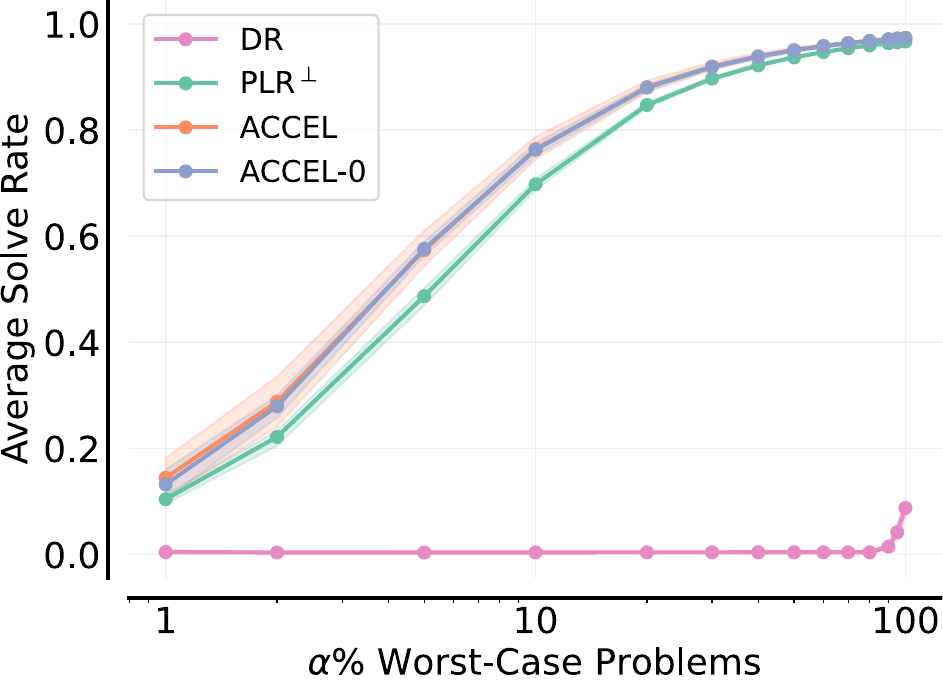}
		\caption{CVaR of the solve rate.}
		\label{fig:val_acc}
	\end{subfigure}
	\hfill
	\begin{subfigure}[b]{\linewidth}
		\centering
		\includegraphics[width=0.7\linewidth]{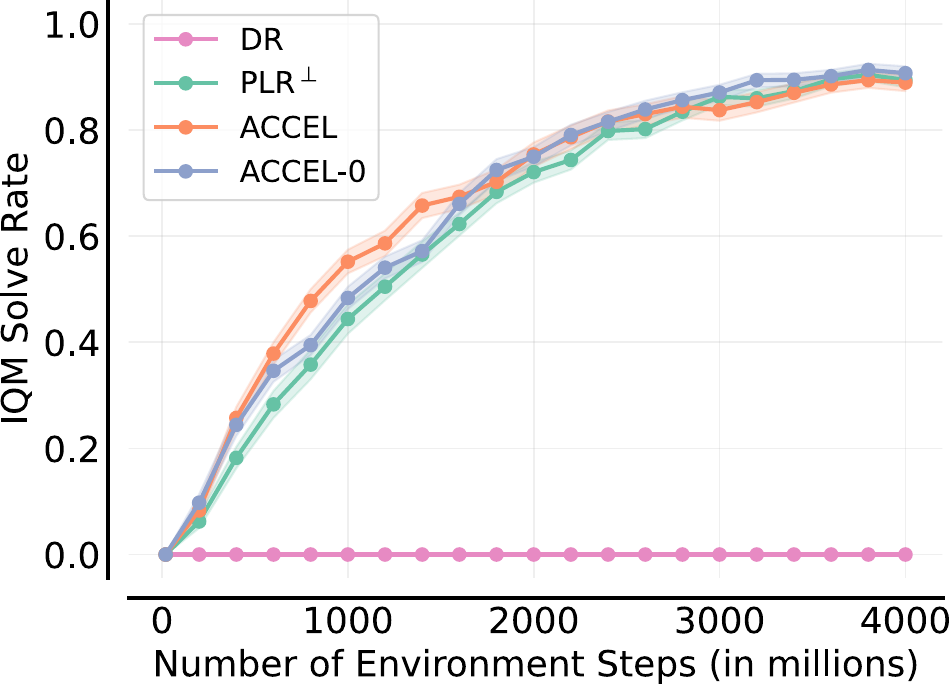}
		\caption{Aggregate zero-shot performance on the hand-designed test set.}
		\label{fig:iqm_solve_rate}
	\end{subfigure}
	\caption{Performance of \ourmethod{abbrv} variants on (a)~the worst-case problems and (b)~a challenging hand-designed test set.}
	\label{fig:ued_cvar_and_test_set}
\end{figure}

\cref{fig:val_acc} displays CVaR results measuring agent \emph{robustness}. \ourmethodaccel variants consistently outperform \rplr across most $\alpha$ values, particularly for worst-case scenarios (low $\alpha$). At $\alpha=100\%$ (average performance), both methods perform similarly and substantially outperform DR, which solves almost no problems.

\cref{fig:iqm_solve_rate} shows zero-shot performance over time on our hand-designed evaluation set. While \ourmethodaccel variants and \rplr achieve comparable final results, \ourmethodaccel converges faster in early training. Notably, \ourmethodaccelzero achieves strong performance, demonstrating that problem complexity can emerge purely through targeted mutations, without relying on initial problem diversity. DR again underperforms due to the scarcity of solvable training problems.

\begin{figure*}
	\begin{subfigure}[b]{0.31\textwidth}
		\begin{minipage}[b]{0.59\linewidth}
			\includegraphics[width=\linewidth]{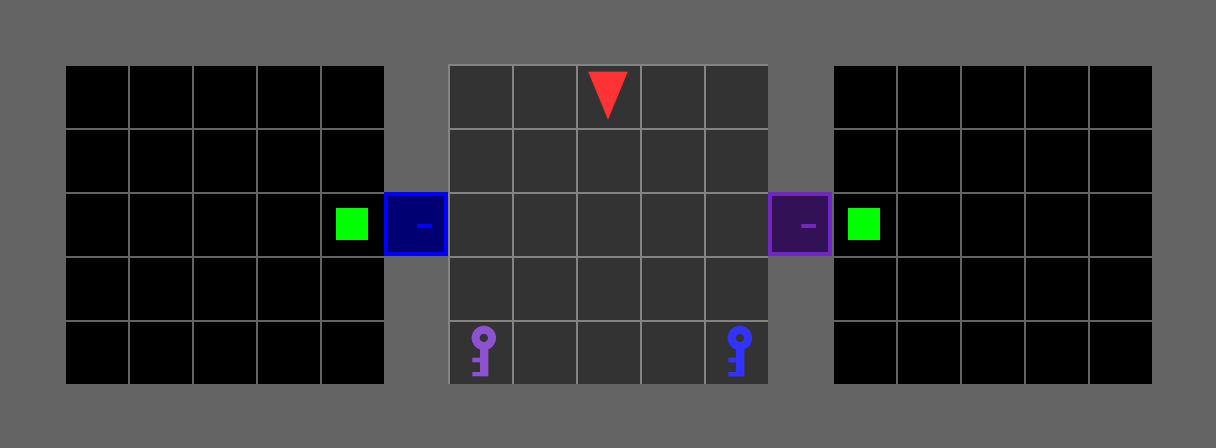}
		\end{minipage}
		\begin{minipage}[b]{0.39\linewidth}
			\includegraphics[width=\linewidth]{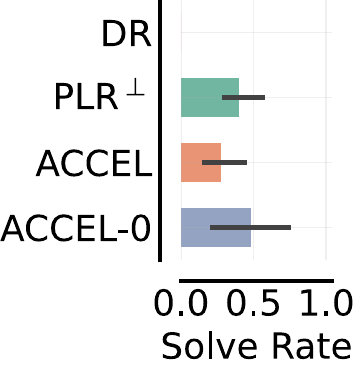}
		\end{minipage}\\[0.5em]
		\centering
		\begin{tikzpicture}[shorten >=1pt,node distance=1.1cm,on grid,auto,every initial by arrow/.style ={-Latex}, state/.style={circle,draw,minimum size=0.2cm,inner sep=0pt}]
			\node[state] (u_0)   {};
			\node[state] (u_1) [right = of u_0]   {};
			\node[state,accepting] (u_acc) [right = of u_1]  {};
			
			\path[-Latex] (u_0) edge node[in place,pos=0.4] {\icnocolor{\icsquare}} (u_1);
			\path[-Latex] (u_1) edge node[in place,pos=0.4] {\iccolor{\icdoorlocked}{icblue}} (u_acc);
		\end{tikzpicture}
		\caption{\textsc{Myopic}}
	\end{subfigure}
	\hfill
	\begin{subfigure}[b]{0.31\textwidth}
		\begin{minipage}[b]{0.59\linewidth}
			\centering
			\includegraphics[width=0.7\linewidth]{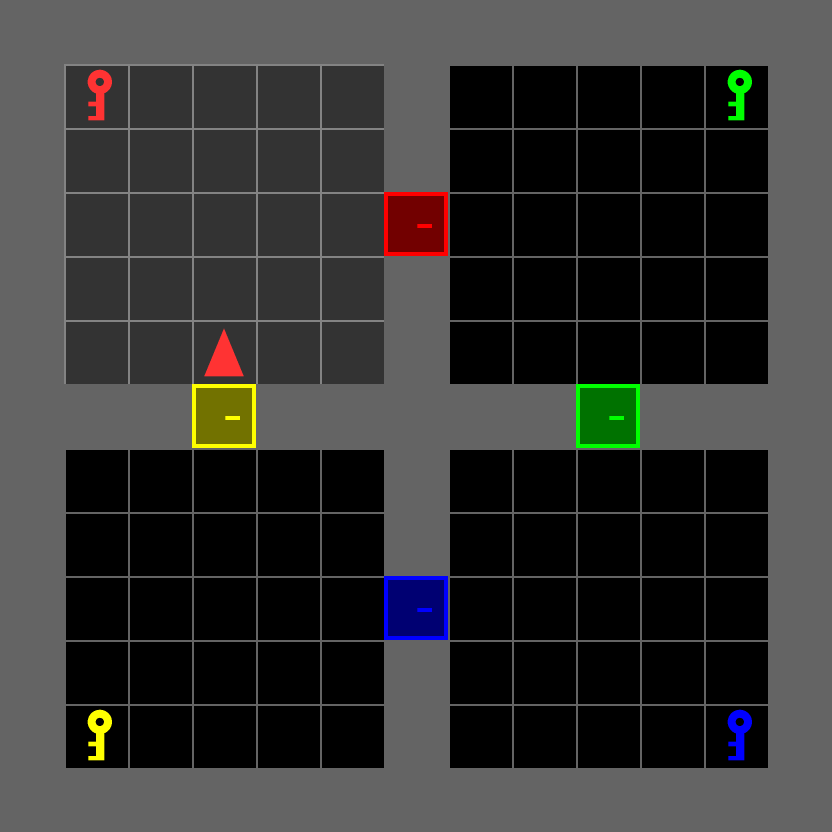}
		\end{minipage}
		\begin{minipage}[b]{0.39\linewidth}
			\includegraphics[width=\linewidth]{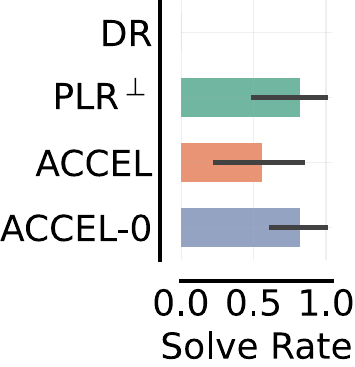}
		\end{minipage}\\[0.5em]
		\centering
		\begin{tikzpicture}[shorten >=1pt,node distance=1.1cm,on grid,auto,every initial by arrow/.style ={-Latex}, state/.style={circle,draw,minimum size=0.2cm,inner sep=0pt}]
			\node[state] (u_0)   {};
			\node[state] (u_1) [right = of u_0]   {};
			\node[state] (u_2) [right = of u_1]   {};
			\node[state] (u_3) [right = of u_2]   {};
			\node[state,accepting] (u_acc) [right = of u_3]  {};
			
			\path[-Latex] (u_0) edge node[in place,pos=0.4] {\iccolor{\icdooropen}{icred}} (u_1);
			\path[-Latex] (u_1) edge node[in place,pos=0.4] {\iccolor{\icdooropen}{icgreen}} (u_2);
			\path[-Latex] (u_2) edge node[in place,pos=0.4] {\iccolor{\icdooropen}{icblue}} (u_3);
			\path[-Latex] (u_3) edge node[in place,pos=0.4] {\iccolor{\icdooropen}{icyellow}} (u_acc);
		\end{tikzpicture}
		\caption{\textsc{Patrol}}
	\end{subfigure}
	\hfill
	\begin{subfigure}[b]{0.31\textwidth}
		\begin{minipage}[b]{0.59\linewidth}
			\centering
			\includegraphics[width=\linewidth]{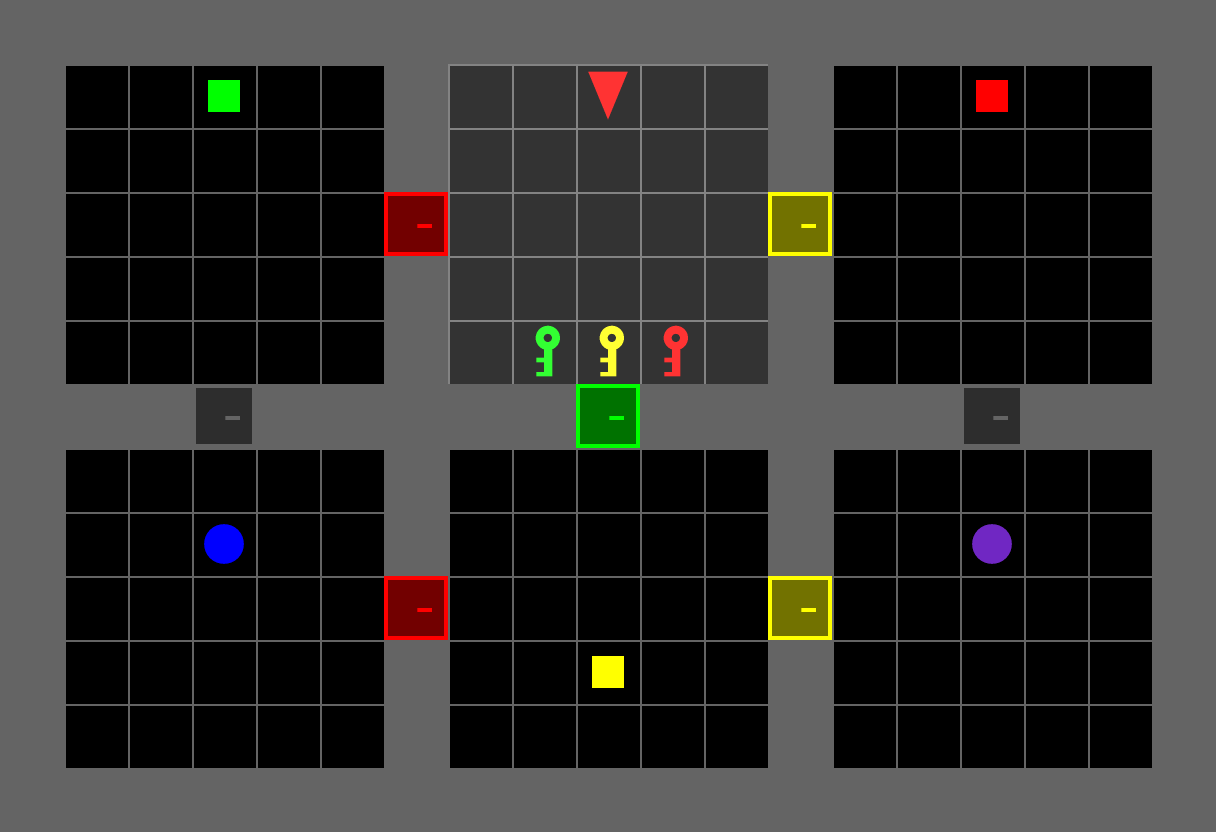}
		\end{minipage}
		\begin{minipage}[b]{0.39\linewidth}
			\includegraphics[width=\linewidth]{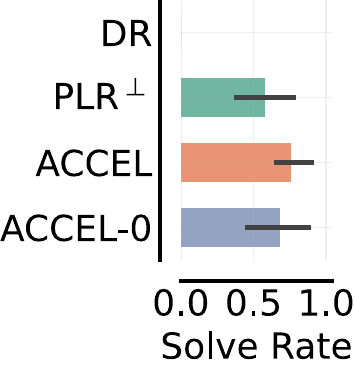}
		\end{minipage}\\[0.5em]
		\centering
		\begin{tikzpicture}[shorten >=1pt,node distance=1.8cm,on grid,auto,every initial by arrow/.style ={-Latex}, state/.style={circle,draw,minimum size=0.2cm,inner sep=0pt}]
			\node[state] (u_0)   {};
			\node[state] (u_1) [right =1.1cm of u_0]   {};
			\node[state] (u_2) [right =1.4cm of u_1]   {};
			\node[state] (u_3) [right =1.1cm of u_2]   {};
			\node[state,accepting] (u_acc) [right =1.4cm of u_3]  {};
			
			\path[-Latex] (u_0) edge node[in place,pos=0.4] {\iccolor{\icdooropen}{icyellow}} (u_1);
			\path[-Latex] (u_1) edge node[in place,pos=0.4] {\icnexthor{\iccolor{\icsquare}{icred}}{\iccolor{\icdoorlocked}{icgreen}}} (u_2);
			\path[-Latex] (u_2) edge node[in place,pos=0.4] {\iccolor{\icdooropen}{icgreen}} (u_3);
			\path[-Latex] (u_3) edge node[in place,pos=0.4] {\icnexthor{\iccolor{\icsquare}{icred}}{\iccolor{\icdoorlocked}{icyellow}}} (u_acc);
		\end{tikzpicture}
		\caption{\textsc{Choice}}
	\end{subfigure}
	\caption{Zero-shot performance of \ourmethod{abbrv} on hand-designed problems. Symbols represent balls ({\iccircle}), squares ({\icsquare}), keys~({\ickey}), and closed/locked/open/unspecified doors ({\icdoorclosed}/{\icdoorlocked}/{\icdooropen}/{\icdoor}). Unspecified doors can match any state. Single symbols indicate $\front$ propositions, pairs indicate $\nextto$ propositions. Striped patterns represent an unspecified color (e.g.,~{\icnocolor{\icsquare}} stands for \small$\front\_\squareobj$).}
	\label{fig:test_problem_subset_performance}
\end{figure*}
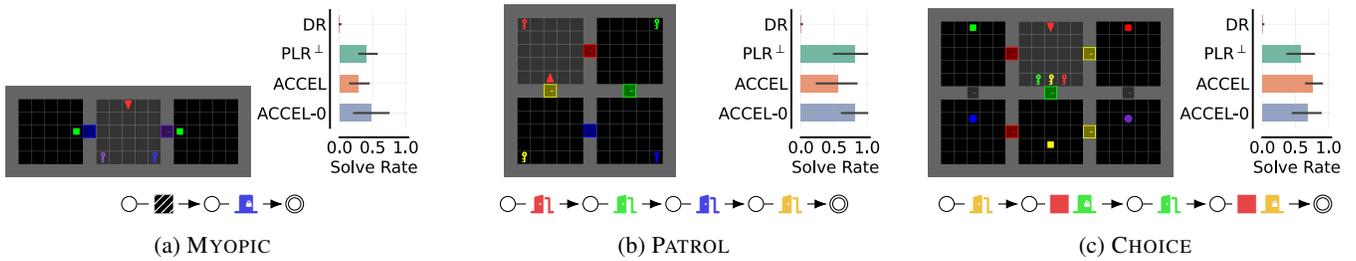

\cref{fig:test_problem_subset_performance} breaks down performance on three challenging hand-designed scenarios. \textsc{Myopic} requires long-term planning: the agent must face a square \emph{but} it cannot be the one behind the locked blue door, as unlocking that door renders the problem unsolvable. \textsc{Patrol} involves navigating with underspecified instructions (i.e., carrying the keys is required to unlock the doors, but not mentioned). \textsc{Choice} combines both: agents must explore the grid while ensuring some doors remain locked. In all cases, regret-based methods outperform DR by a wide margin.

The success of \rplr and \ourmethodaccel stems from filtering solvable problems. Independent sampling yields only 2.7\% solvable problems per batch (see \appref{app:experiments_percentage_solvable}{E.3}), making task-level co-design essential. Both methods curate a buffer of high-regret problems, resulting in the generation of challenging yet solvable problems. Indeed, the fraction of solvable buffer problems steadily grows, eventually nearing 100\% (see \appref{app:experiments_main_results}{E.4})---interestingly, the growth is faster in \ourmethodaccel than for \ourmethodrplr. \ourmethoddr, in contrast, trains on predominantly unsolvable problems throughout.

\subsubsection{Curriculum Analysis (\ref{rq:curric_emergence}).}
\begin{figure*}
	\centering
	\includegraphics[width=\linewidth]{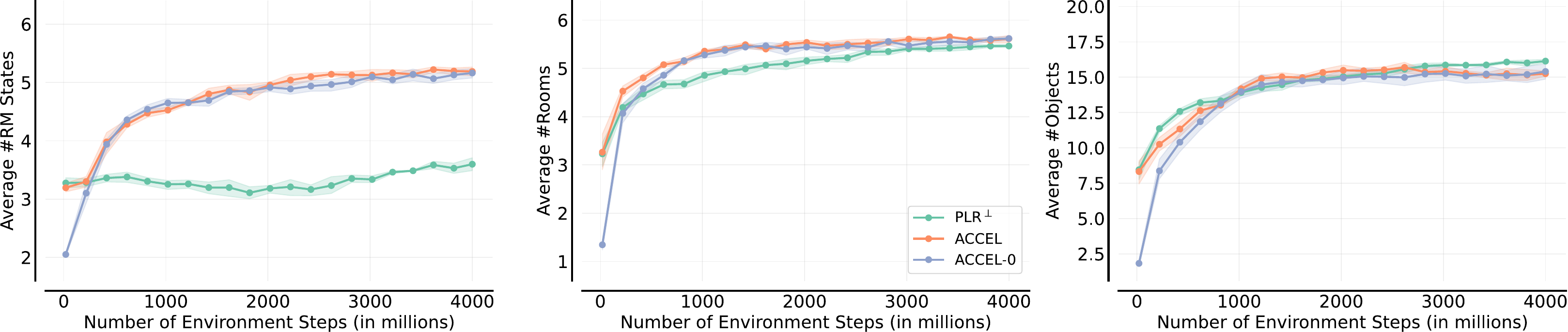}
	\caption{Emergent complexity metrics for problems in the buffer.}
	\label{fig:seq_sampling_curriculum}
\end{figure*}

We analyze how curricula emerge across both tasks and levels. \cref{fig:seq_sampling_curriculum} illustrates the evolution of buffer problems during training. Initially, \ourmethodrplr and \ourmethodaccel curate buffers with simple problems---few rooms, objects, and RM states. Over time, problem complexity increases along all dimensions. \ourmethodaccel variants reach higher RM state counts than \ourmethodrplr, reflecting a stronger emphasis on task complexity. As expected, \ourmethodaccelzero starts with the simplest problems (two states, one room, one object) as it only relies on mutations from these.

\begin{figure*}
	\centering
	\begin{subfigure}[b]{0.14\textwidth}
		\centering
		\includegraphics[height=4em]{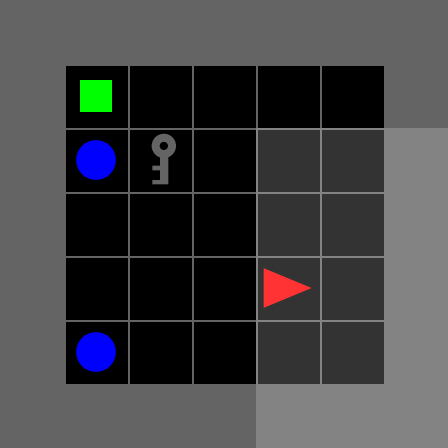}\\[0.2em]
		\begin{tikzpicture}[shorten >=1pt,node distance=1.1cm,on grid,auto,every initial by arrow/.style ={-Latex}, state/.style={circle,draw,minimum size=0.2cm,inner sep=0pt}]
			\node[state] (u_0)   {};
			\node[state] (u_1) [right = of u_0]   {};
			\node[state,accepting] (u_acc) [right = of u_1]  {};
			
			\path[-Latex] (u_0) edge node[in place,pos=0.4] {\icnext{\iccolor{\icsquare}{icred}}{\iccolor{\icdoor}{icgray}}} (u_1);
			\path[-Latex] (u_1) edge node[in place,pos=0.4] {\icnext{\iccolor{\icsquare}{icblue}}{\iccolor{\icdoor}{icgray}}} (u_acc);
		\end{tikzpicture}
	\end{subfigure}
	\hfill
	\begin{subfigure}[b]{0.23\textwidth}
		\centering
		\includegraphics[height=8em]{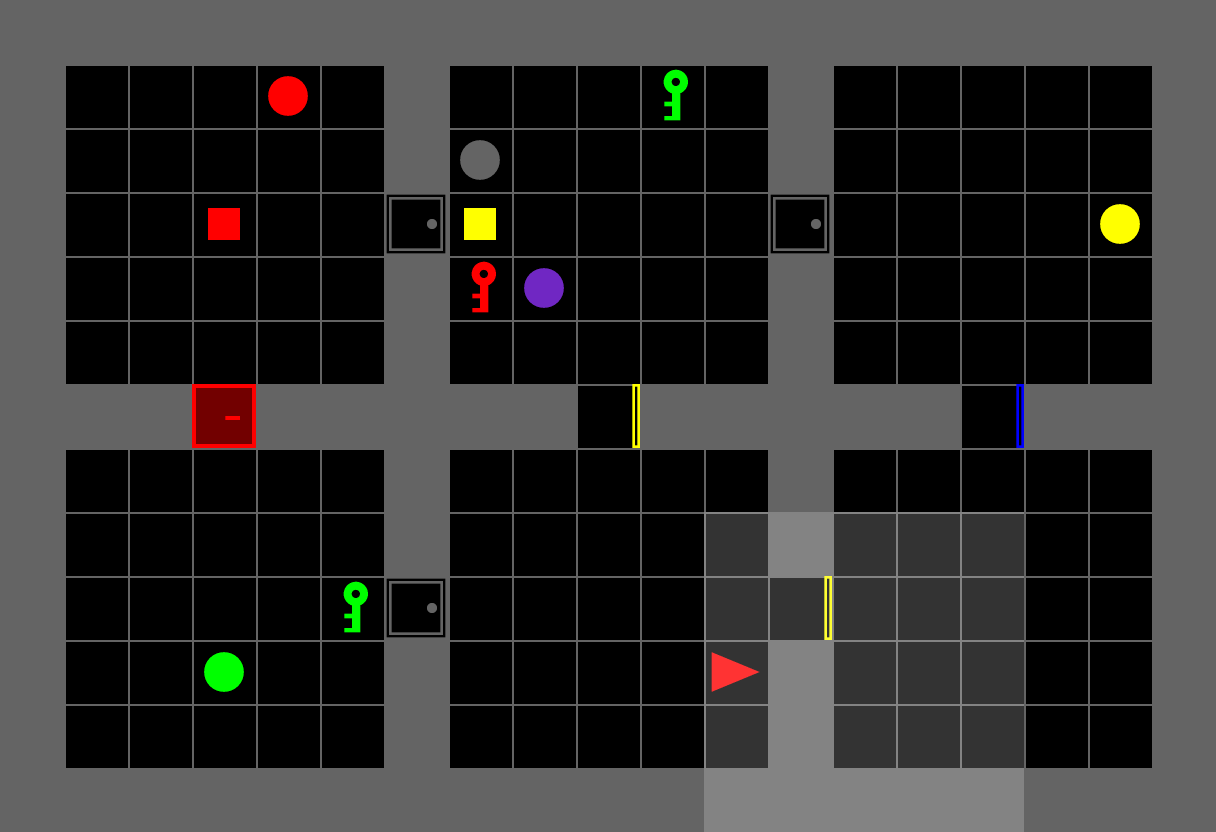}\\[0.2em]
		\begin{tikzpicture}[shorten >=1pt,node distance=1.1cm,on grid,auto,every initial by arrow/.style ={-Latex}, state/.style={circle,draw,minimum size=0.2cm,inner sep=0pt}]
			\node[state] (u_0)   {};
			\node[state] (u_1) [right = of u_0]   {};
			\node[state] (u_2) [right = of u_1]   {};
			\node[state,accepting] (u_acc) [right = of u_2]  {};
			
			\path[-Latex] (u_0) edge node[in place,pos=0.45] {\icnext{\iccolor{\ickey}{icgreen}}{\iccolor{\icdoorclosed}{icred}}} (u_1);
			\path[-Latex] (u_1) edge node[in place,pos=0.45] {\icnext{\iccolor{\iccircle}{icyellow}}{\icnocolor{\iccircle}}} (u_2);
			\path[-Latex] (u_2) edge node[in place,pos=0.45] {\icnext{\iccolor{\iccircle}{icred}}{\iccolor{\icdooropen}{icred}}} (u_acc);
		\end{tikzpicture}
	\end{subfigure}
	\hfill
	\begin{subfigure}[b]{0.27\textwidth}
		\centering
		\includegraphics[height=8em]{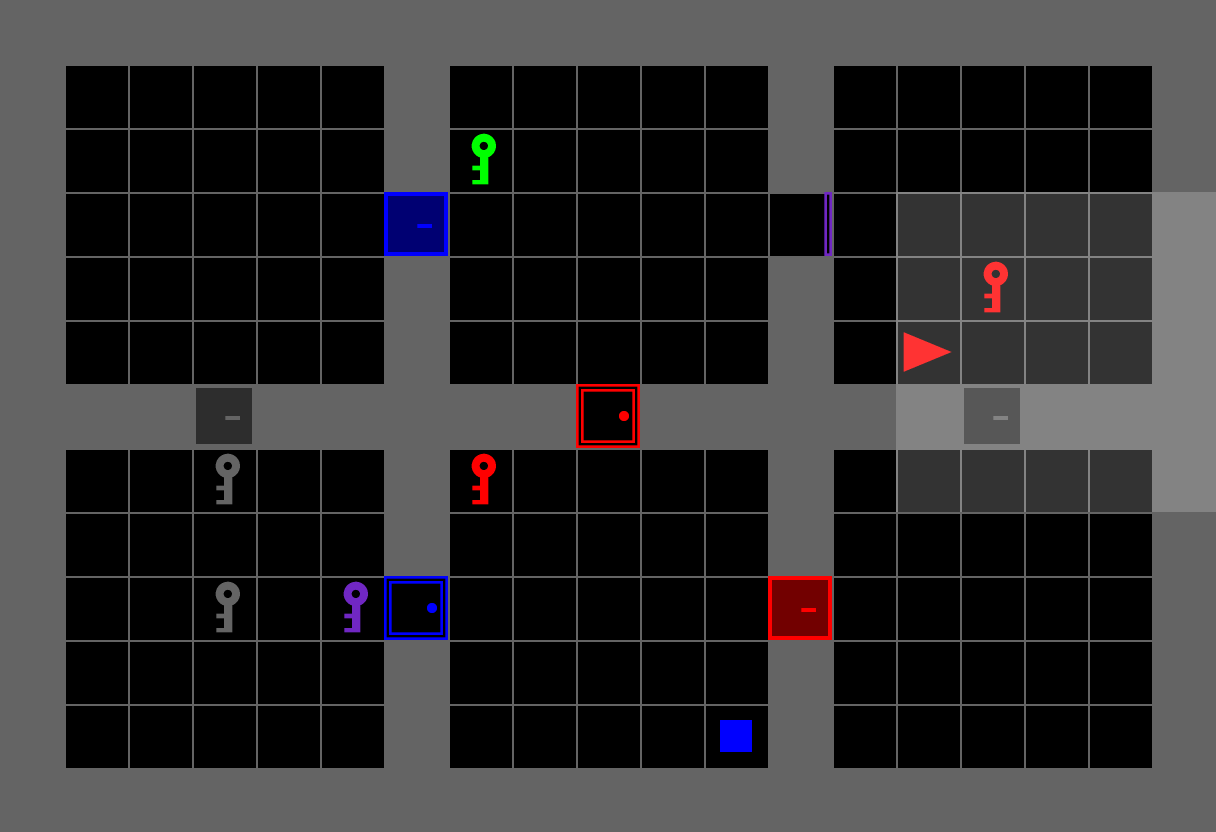}\\[0.2em]
		\begin{tikzpicture}[shorten >=1pt,node distance=1.1cm,on grid,auto,every initial by arrow/.style ={-Latex}, state/.style={circle,draw,minimum size=0.2cm,inner sep=0pt}]
			\node[state] (u_0)   {};
			\node[state] (u_1) [right = of u_0]   {};
			\node[state] (u_2) [right = of u_1]   {};
			\node[state] (u_3) [right = of u_2]   {};
			\node[state,accepting] (u_acc) [right = of u_3]  {};
			
			\path[-Latex] (u_0) edge node[in place,pos=0.45] {\icnext{\icnocolor{\icsquare}}{\iccolor{\icdoorlocked}{icgray}}} (u_1);
			\path[-Latex] (u_1) edge node[in place,pos=0.45] {\icnext{\iccolor{\ickey}{icgreen}}{\iccolor{\icdoor}{icpurple}}} (u_2);
			\path[-Latex] (u_2) edge node[in place,pos=0.45] {\icnext{\iccolor{\icsquare}{icblue}}{\iccolor{\icdoorlocked}{icgray}}} (u_3);
			\path[-Latex] (u_3) edge node[in place,pos=0.45] {\icnext{\iccolor{\icsquare}{icblue}}{\icnocolor{\ickey}}} (u_acc);
		\end{tikzpicture}
	\end{subfigure}
	\hfill
	\begin{subfigure}[b]{0.34\textwidth}
		\centering
		\includegraphics[height=8em]{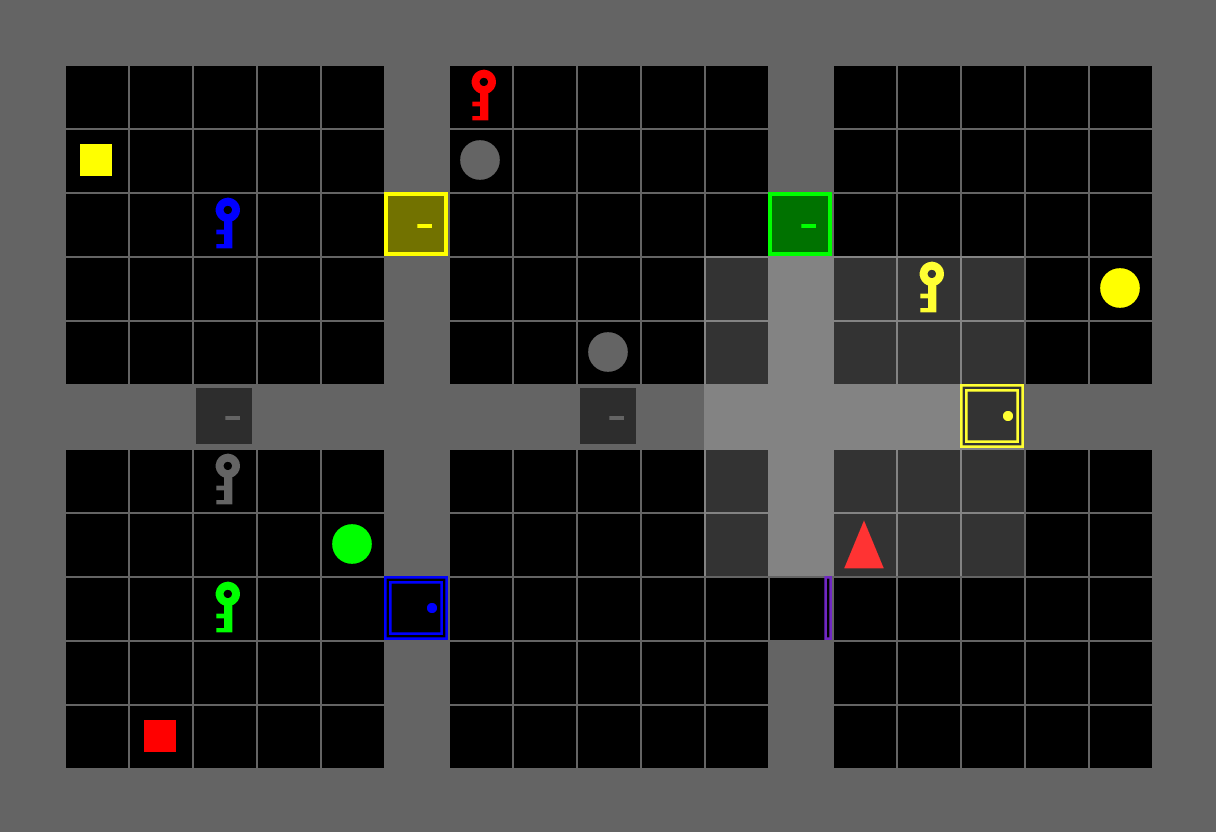}\\[0.2em]
		\begin{tikzpicture}[shorten >=1pt,node distance=1.1cm,on grid,auto,every initial by arrow/.style ={-Latex}, state/.style={circle,draw,minimum size=0.2cm,inner sep=0pt}]
			\node[state] (u_0)   {};
			\node[state] (u_1) [right = of u_0]   {};
			\node[state] (u_2) [right = of u_1]   {};
			\node[state] (u_3) [right = of u_2]   {};
			\node[state] (u_4) [right = of u_3]   {};
			\node[state,accepting] (u_acc) [right = of u_4]  {};
			
			\path[-Latex] (u_0) edge node[in place,pos=0.45] {\icnext{\iccolor{\icsquare}{icyellow}}{\iccolor{\icdoorclosed}{icgray}}} (u_1);
			\path[-Latex] (u_1) edge node[in place,pos=0.45] {\icnext{\iccolor{\iccircle}{icyellow}}{\icnocolor{\ickey}}} (u_2);
			\path[-Latex] (u_2) edge node[in place,pos=0.45] {\iccolor{\icdooropen}{icgreen}} (u_3);
			\path[-Latex] (u_3) edge node[in place,pos=0.45] {\icnext{\iccolor{\ickey}{icgreen}}{\iccolor{\icdooropen}{icgray}}} (u_4);
			\path[-Latex] (u_4) edge node[in place,pos=0.45] {\icnext{\iccolor{\iccircle}{icgreen}}{\iccolor{\icdoorclosed}{icgreen}}} (u_acc);
		\end{tikzpicture}
	\end{subfigure}
	\caption{Generated problems by \ourmethod{abbrv} (\ourmethoddr, \ourmethodrplr, \ourmethodaccel, \ourmethodaccelzero). See \cref{fig:test_problem_subset_performance} for symbol details.}
	\label{fig:generated_samples}
\end{figure*}

\cref{fig:generated_samples} provides examples of generated training problems near the end of training. For \ourmethodrplr and \ourmethodaccel, these are sampled from their respective buffers, while \ourmethoddr examples are drawn directly from the generator. Both \ourmethodrplr and \ourmethodaccel prioritize dense six-room layouts, but \ourmethodaccel favors RMs with more states. The problems generated by \ourmethodrplr and \ourmethodaccel are complex, requiring efficient exploration, implicit door unlocking, and even maintaining doors locked to preserve solvability (e.g.,~the gray doors in \ourmethodaccel). In contrast, \ourmethoddr trains on mostly unsolvable or overly complex problems, which the agent struggles to learn from.

\subsubsection{Mutation Analysis (\ref{rq:mutation_analysis}).}
We observe that the \emph{average number of edits per problem} increases over training. This indicates that (i)~longer edit sequences are beneficial and (ii)~the buffer composition shifts from primarily randomly generated problems to predominantly mutated ones. This trend shows that \ourmethodaccel successfully compounds complexity over time, rather than relying on random sampling alone.

We analyze how different \emph{edit types} shape the curriculum by measuring their frequency in the buffer, weighted by the sampling probabilities of the problems they generate. Task and level edits contribute roughly equally. In contrast, hindsight edits are rare, likely due to limited applicability: they require rollouts to end in intermediate RM states (neither $\rminit$ nor $\rmacc$) and can only appear as the first edit. Exploring more general hindsight edits is an interesting direction for future work. See \appref{app:experiments_main_results}{E.4} for illustrations.

\subsection{Problem Sampling Ablations}
\label{sec:experiments_problem_sampling_ablation_results}
In the default setup, levels and RM tasks are sampled \emph{independently}, yielding only 2.7\% solvable problems per batch. In this setting, \ourmethodrplr and \ourmethodaccel vastly outperformed \ourmethoddr. To test the hypothesis that this gap stems from the scarcity of solvable problems, we perform an ablation with \emph{level-conditioned} problem sampling (see \cref{sec:evaluation_suite_sampling}), which increases solvability to 83.4\%.

We make three key observations. First, as hypothesized, \ourmethoddr improves substantially, matching the CVaR and test performance ($\approx$91.6\%) of \ourmethodrplr and \ourmethodaccel variants. Second, \ourmethodrplr and \ourmethodaccel see only marginal gains ($\approx$1--3\%) over the independent setting, highlighting their robustness when solvable problems are rare. Third, a curriculum still emerges: for \ourmethodrplr, it becomes more pronounced, with RM tasks containing 4--5 states increasingly prioritized. See \appref{app:experiments_problem_sampling_ablation_results}{E.5} for further analysis.

\subsection{Task Sampling Ablations}
To evaluate generalization under more complex training distributions, we replace the sequential task sampler with the \emph{random walk-based} task sampler, restricted to generate RMs as \emph{directed acyclic graphs} (see \cref{sec:evaluation_suite}). This class subsumes sequential tasks and introduces multiple acyclic paths from $\rminit$ to $\rmacc$, creating richer temporal dependencies. As in the default setup, the maximum number of states is 6. We restrict comparisons to \ourmethoddr and \ourmethodrplr under independent problem sampling.

We find that \ourmethodrplr continues to substantially outperform \ourmethoddr for both CVaR and zero-shot performance on the hand-designed set. However,  solve rates on the hand-designed evaluation set drop by $\approx$35\% compared to the sequential setting, indicating that increased task complexity can affect overall performance. See \appref{app:experiments_task_sampling_ablation_results}{E.6} for full results.

\subsection{Mutation Ablations}
\label{sec:experiments_mutation_ablation_results}
We assess \ourmethodaccel's performance under two types of ablations: (i)~\emph{disabling specific edit types}, and (ii)~\emph{varying the edit sequence length}. Disabling either level or task edits significantly reduces performance, while combining both types yields the best results. This highlights the necessity of joint task-level mutations for effective training.

\ourmethodaccel is sensitive to sequence length: short sequences severely hinder performance, with single-edit sequences causing a $\approx$40\% drop and 3-edit sequences leading to a $\approx$15\% drop. In contrast, long sequences (20 edits) have minimal effect ($\approx$2\% variation). See \appref{app:experiments_mutation_ablation_results}{E.7} for details.

\section{Related Work}
\label{sec:rw}

We summarize key related work here, with an extended discussion in \appref{app:rw}{B}.

\subsubsection{Unsupervised Environment Design (UED).}
To the best of our knowledge, prior UED research has focused on level generation for a fixed task. \ourmethod{abbrv} is a UED-based approach that explores the joint co-design of tasks and levels.

Seminal UED approaches rely on regret-based utility scores~\cite{DennisJVBRCL20,JiangDPFGR21}, which we also employ in \ourmethod{abbrv}. However, recent work identifies three key limitations: high-regret levels need not be diverse~\cite{LiVL23}; regret can become irreducible, causing training to stagnate~\cite{BeukmanCMFJDF24}; and common regret approximations correlate with success rate rather than true regret~\cite{RutherfordBWLHF24}. To address these issues, alternative scores---diversity~\cite{LiVL23}, novelty~\cite{TeohLV24}, and learnability~\cite{RutherfordBWLHF24}---have been proposed. We hypothesize irreducible floors may emerge under any scoring function. Our method supports future research using task-level variations to evaluate these limitations.

Alternative approaches to automatic level design beyond UED have been proposed. POET~\cite{WangLCS19} trains populations of specialist agents using evolutionary strategies, while \ourmethod{abbrv} trains general agents from random and co-evolved task-level pairs. PCGRL~\cite{KhalifaBET20} frames level design as an RL problem, where levels are incrementally edited to optimize a given quality objective.

\subsubsection{Formal Language Conditioning.}
Most closely related to our work, \citet{YalcinkayaLVS23,YalcinkayaLVS24} condition policies on GNN embeddings of \emph{automata} representing reach-avoid task sequences. \citet{YalcinkayaLVS23} derive training automata from observed proposition trajectories (akin to our hindsight edits), while \citet{YalcinkayaLVS24} apply task mutations. Unlike these approaches, \ourmethod{abbrv} targets high-regret tasks for curriculum generation and co-evolves \emph{both} tasks and levels rather than tasks alone.

Linear temporal logic~\cite[LTL;][]{Pnueli77} approaches typically train using formulas sampled from context-free grammars. \citet{KuoKB20} and \citet{VaezipoorLIM21} encode task structure by embedding the formula's syntax tree through compositional RNNs and GNNs, respectively. \citet{QiuMZ23} and \citet{JackermeierA25}, unlike previous work and \ourmethod{abbrv}, tackle \emph{infinite-horizon} tasks by mapping formulas into equivalent automata whose paths determine the conditioning---the former is sequentially conditioned on each subtask along a path, while the latter is conditioned on derived reach-avoid sequences.

These approaches implement DR, sometimes with fixed curricula~\cite{JackermeierA25}, which may collapse when solvable problems are rarely sampled. In contrast, \ourmethod{abbrv} leverages regret-based UED to generate \emph{autocurricula} of solvable yet challenging problems. We hypothesize prior work with DR succeeds because their domains ensure high solvability: all propositions are observable across levels, making most tasks completable. We address this evaluation gap through settings with naturally low solvability rates, revealing some limitations of DR.

\section{Conclusions}
\label{sec:conclusions}
We introduce \ourmethod{abbrv}, a novel method for generating joint autocurricula over problems (task-level pairs). By co-designing tasks and levels via regret-based UED, \ourmethod{abbrv} achieves robust generalization even when most sampled problems are unsolvable---a setting in which domain randomization fails. Additionally, \ourmethodaccelzero achieves strong performance and builds rich curricula by repeatedly mutating initially simple problems. We contribute 150 hand-designed test problems on which we verify that these findings hold, validating our approach and providing a foundation for further research on task-level generalization.

\subsubsection{Future Work.} Our framework opens several promising research directions. First, extending the approach to other temporal formalisms---including LTL, programs, or hierarchies of RMs~\cite{FurelosBlancoLJBR23}---would demonstrate the broad applicability of joint curriculum generation via UED. Second, developing scoring functions and mutation operators that better exploit task structure could yield even more effective curricula. Finally, scaling to domains like Craftax~\cite{MatthewsBESJCF24} would test the effectiveness of joint curricula in richer settings.

\section*{Acknowledgements}
We thank the reviewers and Roko Para\'{c} for their feedback, Isabella Pearce for her help in designing test problems, and Dugan Witherick for his support in setting up the resources provided by the Imperial College Research Computing Service (\url{http://doi.org/10.14469/hpc/2232}).

This work is partially supported by DEVCOM Army Research Lab under grant W911NF2220243 and EPSRC projects EP/X040518/1 and EP/Y037421/1. CP is supported by UK EPSRC grant 2760033. FK is supported by UK EPSRC grant 2757464. 

\bibliography{aaai2026}

\ifarxiv
\appendix
\onecolumn
\section{Evaluation Suite Details}
\label{app:evaluation_suite}
In this section, we describe additional information about the evaluation suite introduced in \cref{sec:evaluation_suite}. We organize the section as follows: \cref{app:evaluation_suite_implementation} describes some implementation details and features omitted in the main paper, \cref{app:random_walk} outlines random walk-based sampling, and  \cref{app:evaluation_suite_hand_designed} provides some details about the structure of the problems in the hand-designed set. Illustrative example problems are given throughout the paper and the appendix (e.g.,~see \cref{sec:method,sec:experiments,app:mutations,app:experiments_task_sampling_ablation_results}).

\subsection{Implementation Details}
\label{app:evaluation_suite_implementation}
The suite is implemented in JAX~\cite{jax2018github} to enable seamless execution across CPUs and hardware accelerators (GPUs, TPUs). The implementation of Minigrid is based on that by \citet{NikulinKZASK24}.

\subsubsection{Alphabet.}
Following the work of \citet{MuZRJGRG22}, who also build upon BabyAI instructions, we do not encode propositions for commands involving objects relative to the agent's position, e.g.~\emph{``go to a ball on your left''}. None of the commands imply the uniqueness of an object, e.g.~\emph{``pick up \underline{the} yellow square''}. The \emph{alphabet size} consists of 889 propositions, which is almost two orders of magnitude larger than the alphabets considered in previous work \cite{KuoKB20,VaezipoorLIM21,YalcinkayaLVS24,JackermeierA25}. The alphabet does not contain symmetric propositions (e.g.,~only contains one of $\nextto\_\ball\_\key$ and $\nextto\_\key\_\ball$) and is derived under the assumption that two doors cannot be next to each other (one of the objects in a $\nextto$ proposition must be movable).

\subsubsection{Hierarchies of RMs.}
Our implementation supports hierarchies of RMs~\cite[HRMs;][]{FurelosBlancoLJBR23}, which are composed by enabling RMs to call each other. This modular structure supports task decomposition and reusability across HRMs. For such HRMs, a single RM is designated as the root, initiating execution and terminating upon reaching its accepting state; other RMs return control to their caller upon completion. Our random walk-based sampler (described in \cref{app:random_walk}) can generate hierarchical structures; however, we focus on flat HRMs (i.e., standard RMs) for simplicity and defer the use of hierarchies to future work.

\subsubsection{Task Sampling.}
For both the sequential and the random walk-based HRM samplers, there is a single edge between each pair of states. We provide an in-depth description of the random walk-based sampler in \cref{app:random_walk}.

\subsubsection{Problem Sampling.}
We implement a \emph{task-conditioned} sampler that complements the independent and level-conditioned sampling strategies (see \cref{sec:evaluation_suite_sampling}). The task-conditioned sampler generates levels conditioned on the sampled HRM's propositions, ensuring that objects related to the HRM's propositions are present in the level. For example, given the proposition $\nextto\_\ball\_\blue\_\squareobj$, the level will be guaranteed to contain a blue ball and a square of any color. Upon choosing a conditioning strategy for the ablations (see \cref{sec:experiments_problem_sampling_ablation_results}), we selected the level-conditioned sampler as it better reflects real-world problem specification. In practice, the physical environment determines the agent's affordances (i.e., what actions and interactions are possible), which then constrains the meaningful range of tasks that can be specified. This mirrors how humans naturally assign tasks based on environmental context and available objects.

\subsection{Random Walk-Based HRM Generation}
\label{app:random_walk}
We here describe the method for sampling HRM tasks via random walks, briefly outlined in \cref{sec:evaluation_suite}. \cref{fig:rw_example} illustrates an HRM sampled via random walks. Unlike other hierarchies depicted throughout the paper, these HRMs may consist of multiple RMs, and each constituent RM need not be sequential. We specify HRMs using the following components, where $\rwnumrms$ is the number of RMs in the hierarchy:
\begin{itemize}
	\item a hierarchical graph structure \( \rwhstruct = \langle \rwhstructv, \rwhstructe \rangle \), where $\rwhstructv$ is a set of nodes for each constituent RM, and $\rwhstructe\subseteq\rwhstructv\times\rwhstructv$ is a set of edges between these nodes;
	\item local RM graph structures \( \rwlocalrm_i = \langle \rwlocalrmv_i, \rwlocalrme_i \rangle \) for $i\in[1,\rwnumrms]$, where $\rwlocalrmv_i$ is a set of nodes for each RM state, and $\rwlocalrme_i\subseteq\rwlocalrmv_i\times\rwlocalrmv_i$ is a set of edges between these nodes;
	\item the call-labeling function \( \rwcalllabfun: \rwlocalrme_i \rightarrow \{\rwlocalrm_j\}_{j=1}^\rwnumrms \), mapping each edge in RM $i$ to an RM graph; and
	\item the edge-labeling function \( \rwedlabfun: \rwlocalrme_i \rightarrow \propset \), mapping each edge in RM $i$ to a proposition.
\end{itemize}
This specification can be easily transformed into the HRMs defined by \citet{FurelosBlancoLJBR23}. In the following paragraphs, we describe how each of these components is sampled.

\subsubsection{Hierarchical Graph Structure Sampling.}
The hierarchical graph structure $\rwhstruct$ determines which RMs may be invoked from a given RM. Since the number of RMs in the hierarchy is $\rwnumrms$, the number of non-leaf nodes (i.e., RM nodes calling at least one RM) is at most \( \lfloor \rwnumrms / 2 \rfloor \). To generate the tree structure, we sample from a categorical distribution \( \text{Cat}(p(\lfloor \rwnumrms / 2 \rfloor), q) \) over the ordered integer partitions \( p(\lfloor \rwnumrms / 2 \rfloor) \), with associated categorical weights \( q \). Here, \emph{ordered} means that the order of summands matters. The value of the \( i \)-th summand in a partition specifies the number of children of the \( i \)-th node. Edges are connected sequentially based on these values. For balanced hierarchies, we recommend using a uniform distribution over the partitions, ensuring that all nodes have equal branching factors. Alternatively, the weights \( q \) can be tuned to bias the hierarchy toward flatter or deeper structures.

\begin{figure*}[t]
	\begin{minipage}[t]{0.38\linewidth}
		\centering
		$M_0$\\[0.5em]
		\begin{tikzpicture}[shorten >=1pt,node distance=2.7cm,on grid,auto,every initial by arrow/.style ={-Latex}]
			\node[state,initial,initial text=] (u_0)   {$\rminit$};
			\node[state] (u_1) [below =2cm of u_0]   {$\rmidx{1}$};
			\node[state] (u_2) [left = of u_1]   {$\rmidx{2}$};
			\node[state,accepting] (u_acc) [below =2cm of u_1]  {$\rmacc$};
			
			\path[-Latex] (u_0) edge node[in place] {\icnexthor{\iccolor{\ickey}{icyellow}}{\icnocolor{\icdoor}}} (u_1);
			\path[-Latex] (u_1) edge[bend right] node[in place] {\small$M_1 \mid$ \icnexthor{\iccolor{\icsquare}{icpurple}}{\iccolor{\ickey}{icyellow}} } (u_2);
			\path[-Latex] (u_2) edge[bend right] node[in place] {\small$M_1 \mid$ \icnexthor{\iccolor{\iccircle}{icyellow}}{\iccolor{\icdoorclosed}{icblue}} } (u_1);
			\path[-Latex] (u_1) edge node[in place] {$M_1 \mid $\icnexthor{\iccolor{\icsquare}{icpurple}}{~\iccolor{\icdoor}{icpurple}}} (u_acc);
		\end{tikzpicture}
	\end{minipage}
	\begin{minipage}[t]{0.38\linewidth}
		\centering
		$M_1$\\[0.5em]
		\begin{tikzpicture}[shorten >=1pt,node distance=2.7cm,on grid,auto,every initial by arrow/.style ={-Latex}]
			\node[state,initial,initial text=] (u_0)   {$\rminit$};
			\node[state] (u_1) [below =2cm of u_0]   {$\rmidx{1}$};
			\node[state] (u_2) [above left =1.5cm and 3cm of u_1]   {$\rmidx{2}$};
			\node[state] (u_3) [below left =1.5cm and 3cm of u_1]   {$\rmidx{3}$};
			\node[state,accepting] (u_acc) [below =2cm of u_1]  {$\rmacc$};
			
			\path[-Latex] (u_0) edge node[in place] {\icnexthor{\iccolor{\icsquare}{icyellow}}{\iccolor{\icdoorclosed}{icgray}}} (u_1);
			\path[-Latex] (u_1) edge node[in place] {\small$M_2 \mid$ \icnexthor{\iccolor{\iccircle}{icgreen}}{~\iccolor{\icdoor}{icblue}} } (u_acc);
			\path[-Latex] (u_1) edge node[in place] {\small$M_2 \mid$ \icnexthor{\iccolor{\iccircle}{icpurple}}{\iccolor{\ickey}{icyellow}} } (u_2);
			\path[-Latex] (u_2) edge node[in place] {\small$M_2 \mid$ \icnexthor{\iccolor{\icsquare}{icpurple}}{\iccolor{\icdooropen}{icblue}} } (u_3);
			\path[-Latex] (u_3) edge node[in place] {\small$M_2 \mid$ \icnexthor{\iccolor{\iccircle}{icred}}{\iccolor{\icdoorclosed}{icyellow}} } (u_1);
		\end{tikzpicture}
	\end{minipage}
	\begin{minipage}[t]{0.2\linewidth}
		\centering
		$M_2$\\[0.5em]
		\begin{tikzpicture}[shorten >=1pt,node distance=2cm,on grid,auto,every initial by arrow/.style ={-Latex}]
			\node[state,initial,initial text=] (u_0)   {$\rminit$};
			\node[state,accepting] (u_acc) [below = of u_0]  {$\rmacc$};
			
			\path[-Latex] (u_0) edge node[in place] {\icnexthor{\iccolor{\iccircle}{icgreen}}{\iccolor{\icsquare}{icgreen}}} (u_acc);
		\end{tikzpicture}
	\end{minipage}
	\caption{Example of an HRM generated via the random-walk based sampler. Symbols follow the descriptions in \cref{fig:test_problem_subset_performance}. Calls to other RMs are denoted with $M_i\mid\varphi$, where $M_i$ is the called RM and $\varphi$ a formula that must be satisfied for the call to be started. Negative propositions in the cases where there are multiple outgoing transitions from a given state are omitted, e.g.~the edges from $\rmstate_1$ in $M_0$ are actually labeled with the formulas $\text{\icnexthor{\iccolor{\icsquare}{icpurple}}{\iccolor{\ickey}{icyellow}}}\land \neg \text{\icnexthor{\iccolor{\icsquare}{icpurple}}{~\iccolor{\icdoor}{icpurple}}}$ and $\text{\icnexthor{\iccolor{\icsquare}{icpurple}}{~\iccolor{\icdoor}{icpurple}}} \land \neg\text{\icnexthor{\iccolor{\icsquare}{icpurple}}{\iccolor{\ickey}{icyellow}}}$.}
	\label{fig:rw_example}
\end{figure*}
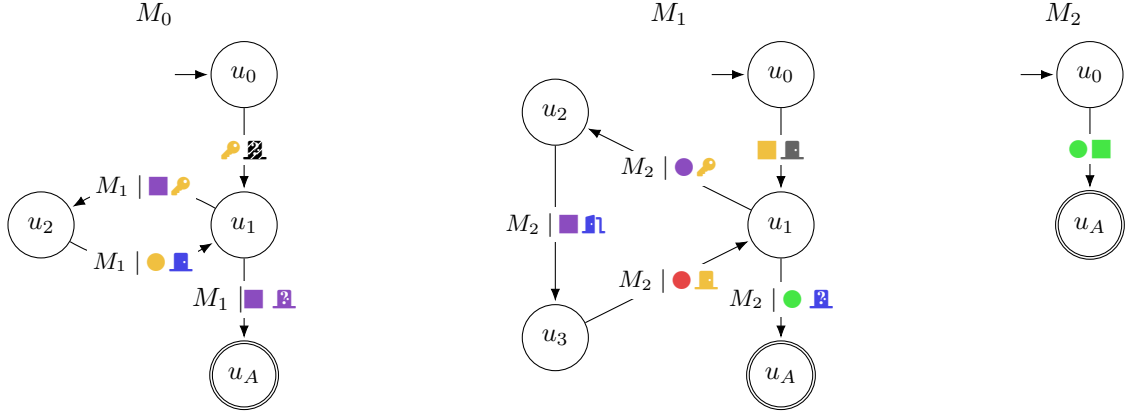

\subsubsection{RM Sampling.}
The RM sampling is parallelized. Each RM \( i \) is instantiated via a realization of a Markov chain defined by a transition matrix \( \rwrmtransition_i \in \realnums^{\card{\rmstates} \times \card{\rmstates}} \). For sequential RMs, \( \rwrmtransition_i \) is a diagonal matrix with a unit diagonal shift. For directed acyclic RMs, \( \rwrmtransition_i \) is lower triangular with a diagonal shift of one. Each matrix \( \rwrmtransition_i \) is sampled as \( \rwrmtransition_i \sim \mathcal{D} \), where \( \mathcal{D} \) is typically chosen as a uniform distribution to avoid introducing bias, though it can be tuned to induce other desirable structures. The sampled matrix \( \rwrmtransition_i \) is then normalized to be right-stochastic. Let \( \rwavgconnect \) denote the average node connectivity in the sampled graph. Given \( \card{\rmstates} \) as the number of states, the RM structure \( \rwlocalrm_i \) is generated from the connectivity distribution induced by a random walk of length \( \rwavgconnect \card{\rmstates} \), i.e., \( \rwlocalrm_i \sim \text{RW}_{\rwavgconnect \card{\rmstates}}(\rwrmtransition_i) \). After sampling a structure \( \rwlocalrm_i \), the random walk may be restarted with a fixed probability, and the final graph is obtained by taking the union of edges from all sampled trajectories. This mechanism enables control over the distribution of paths per RM. In the case of cyclic graphs, it imposes a lower bound on the number of distinct paths. For non-root RMs, we introduce one modification: we set \( \rwrmtransition_i(0, j) = 0 \) for all \( j \), ensuring that the initial state has exactly one outgoing edge. The motivation for this constraint will become clear in the following sections.

\subsubsection{Proposition Labeling.} 
Labeling must be performed conditionally at random to avoid pathological cases. There are two such undesirable cases.  
First, a labeling is \emph{tautological} if an incoming transition is labeled with a proposition \( p_k \) such that \( p_k \rightarrow p_l \equiv \top \), where \( p_l \) is the label of an outgoing transition.  
Second, a labeling is \emph{non-deterministic} if two outgoing transitions from the same node are labeled with propositions \( p_{l_1} \) and \( p_{l_2} \) such that either \( p_{l_1} \rightarrow p_{l_2} \equiv \top \) or \( p_{l_1} \leftarrow p_{l_2} \equiv \top \). For every node, each of the outgoing $\rwnumouttrans$ transition labelings is represented as a one-hot encoded vector \( \rwoutlabelings^{(l)} \in \{0,1\}^\propset \) over the proposition set sampled from the conditional categorical distribution:
\begin{align*}
	\rwoutlabelings^{(l)} \sim \textnormal{Cat}\left(\propset, 
	\left( \prod_{i=1}^\rwnumintrans \rwconstmatrix \cdot \rwinlabelings^{(i)} \right) \odot 
	\left( \prod_{i=1}^{\rwnumouttrans-1} \rwconstmatrix \cdot \rwoutlabelings^{(i)} \right) \odot 
	\rwcatprior \right),
\end{align*}
where \( \rwconstmatrix \) is a binary constraint matrix encoding the undesirable cases above and defined by
\begin{align*}
	\rwconstmatrix_{ij} = 
	\begin{cases}
		0 & \text{if } p_j \rightarrow p_i \equiv \top, \\
		1 & \text{otherwise},
	\end{cases}
\end{align*}
$\rwnumintrans$ is the number of incoming transitions, $\rwinlabelings^{(i)}\in\{0,1\}^\propset$ is a one-hot vector representing the labeling of the $i$-th incoming transition, and \( \rwcatprior \in \{0,1\}^\propset \) is a categorical prior over the proposition set. To avoid introducing bias into the labeling, labels are sequentially assigned by sampling uniformly at random from the available outgoing edges \( \rwlocalrme_i \). Finally, to ensure determinism, each label must be conjoined with the negation of its neighboring labels (i.e.,~labels on the transitions from the same node).

\subsubsection{Call Labeling.}
Call labeling follows principles similar to those used for proposition labeling. However, in this case, the labels correspond to calls determined during hierarchy sampling; that is, the hierarchy defines the call alphabet for each RM. As with proposition labels, call labels are subject to consistency constraints. Since any call condition is conjoined with the initial transition labeling of the RM being called, the two must not be contradictory. This dependency also motivates the design choice of restricting RMs in the hierarchy to have exactly one initial transition without any call label. This restriction ensures tractability over call labelings. Let \( \rwcallcompat \) be a binary compatibility matrix defined by:
\begin{align*}
	\rwcallcompat_{ij} = 
	\begin{cases}
		0 & \text{if } p_j \land p_i \equiv \bot, \\
		1 & \text{otherwise},
	\end{cases}
\end{align*}
where \( p_i \) and \( p_j \) are propositions. Then, for each edge \( \mathcal{E}_i^{(j)} \), the call label \( c_j \) is sampled from the conditional categorical distribution:
\begin{align*}
	c_j \sim \textnormal{Cat}\left(C_i \cup \{\top\},\, \left(\rwcallcompat \cdot \rwinittransmatrix^{(i)}\right) \cdot \rwcallcatprior^{(i)}\right),
\end{align*}
where \( C_i \subset \{\rwlocalrm_j\}_{j=1}^\rwnumrms \) is the set of callable RMs as defined by the hierarchy $\rwhstruct$ from RM $i$, and \( \rwinittransmatrix^{(i)} \) is a matrix whose columns correspond to the proposition vectors associated with the initial transitions of the RMs in \( C_i \). The vector \( \rwcallcatprior^{(i)} \) specifies the categorical prior over call probabilities.

\subsection{Hand-Designed Evaluation Set}
\label{app:evaluation_suite_hand_designed}
The hand-designed evaluation set consists of 150 problems. \emph{Levels} mostly follow the grid morphologies used for training (see \cref{app:experiments_setup}), which divide the grid into different equally-sized rooms. Unlike random generation, we distribute objects across the grid in challenging ways, e.g.~sparsely across different rooms, surrounded by several distractor objects, or enforcing door unlocking to reach certain areas. A small fraction of the problems (6) consist of levels without the specified room morphology. \emph{Tasks} are all non-hierarchical RMs, most of which are \emph{sequential}, following our default training scenario; however, some consist of multiple paths from $\rminit$ to $\rmacc$, or longer paths than the training default (5). The RMs are designed to assess performance in scenarios where considering further ahead than the current subgoal (i.e., the literals labeling an outgoing edge from the current RM sate) is needed, or where the tasks are fully specified (i.e.,~contain all required subgoals) or underspecified (i.e.,~omit some subgoals, such as picking up a key to open a door). Examples of these hand-designed problems are shortly described in \cref{sec:experiments_main_results}.

\section{Extended Related Work}
\label{app:rw}
In this section, we provide additional details on the related work covered in \cref{sec:rw}.

\subsection{Unsupervised Environment Design (UED)}
UED is closely related to the broader field of \emph{autocurricula} in reinforcement learning, as it aims to generate training environments that adapt to an agent's capabilities. We briefly describe some alternative approaches that induce autocurricula. \citet{FlorensaHGA18} propose using a generator network to produce tasks at the frontier of the agent's capabilities and optimize it via adversarial training. The UED methods \ourmethod{abbrv} builds upon do not train a generator, but rely on selective sampling to train from solvable yet challenging problems. \citet{MatiisenOCS20} propose algorithms that sample problems based on progress, which is defined as the slope of the learning curve. Unlike the UED approaches considered here, these algorithms assume access to all problems at the start of training. Similar methods in the context of formal language conditioning are outlined in \cref{app:rw_formal_lang_cond}. Asymmetric self-play~\cite{SukhbaatarLKSSF18,OpenAI21} consists of one agent producing a trajectory and another agent having to reach the final state of that trajectory as a goal. Since the goal depends on an agent's trajectory, goals are guaranteed to be attainable; in contrast, goals may not be feasible in the DR setting we consider. Other approaches focus on generating increasingly complex intrinsic goals that encourage the agent to explore and facilitate the achievement of extrinsic environment goals~\cite[e.g.,][]{CamperoRKTRG21,MuZRJGRG22}. For a comprehensive treatment of autocurricula, we refer to the position paper by \citet{LeiboHLG19} and the surveys by \citet[Section~4.1]{NarvekarPLSTS20}, \citet{PortelasCWHO20}, \citet[Section~6]{ColasKSO22}, and \citet[Section~4.8]{KlissarovBLKPM25}.

Adaptive UED methods often rely on regret estimates to determine the utility of levels. In this work, we employed MaxMC~\cite{JiangDPFGR21}; however, there are other methods such as the \emph{positive value loss}~\cite[PVL;][]{JiangDPFGR21}, which we experiment with in \cref{app:experiments_scoring_fn_ablation_results}. Both MaxMC and PVL rely on a single student. In contrast, the seminal UED approach, PAIRED~\cite{DennisJVBRCL20}, computes regret as the value difference between two student policies; specifically, the teacher aims to exploit the weaknesses of one student (the protagonist) in relation to the other student (the antagonist), producing an emergent curriculum of levels at the frontier of the protagonist's capabilities. We outline some known limitations of regret estimation in \cref{sec:rw}.

PAIRED, as highlighted by \citet{JiangDPFGR21}, is slow to adapt to changes in student policies since the teacher adapts through gradient updates. PLR~\cite{JiangGR21} takes a different approach: the teacher does not generate levels, but curates a buffer of previously seen levels. Crucially, the teacher adapts faster than in PAIRED since it only implements a scoring mechanism. \rplr~\cite{JiangDPFGR21} extends PLR by training from buffer levels only, which improves the convergence to Nash equilibria. ACCEL~\cite{Parker-HolderJ022} extends PLR methods by mutating previously sampled high-regret levels, showing that complexity can emerge starting from simple levels. Recent directions in UED explore environment generation via diffusion~\cite{ChungLKKO24} and world models~\cite{AhmetJLRFBP25}, the distribution shift in adaptive curricula~\cite{GarcinDGLA24}, the application in multi-agent scenarios~\cite{SamvelyanK0JPFR23,RuhdorferBPB25}, and stronger theoretical guarantees~\cite{MonetteLBJRGF25}.

Prior UED approaches have primarily focused on level curricula with fixed tasks. Closer to \ourmethod{abbrv}, \citet{RutherfordBWLHF24} recently evaluated different UED algorithms in XLand-Minigrid~\cite{NikulinKZASK24}, an environment that enables generalization across tasks. However, these goals are specified as single objectives, whereas RMs encode explicit multi-step subgoal sequences. Furthermore, we propose mutations that leverage RM structure.

\subsection{Formal Language Conditioning}
\label{app:rw_formal_lang_cond}
We provide further details on the core components of the methods that learn policies conditioned by formal language task specifications, mainly LTL and automata. \cref{tab:goal_cond_comp} provides an overview of the main distinguishing factors, which we elaborate on in the following paragraphs and \cref{sec:rw}.

\begin{table*}[t]
	\centering
	\resizebox{\textwidth}{!}{%
		\begin{tabular}{lccccccccc}
			\toprule
			\multicolumn{1}{c}{\textbf{Algorithm}} & \textbf{Conditioning} & \textbf{Curriculum Target} & \textbf{Autocurric.} & \textbf{Sampling} & \textbf{Pretraining} & \textbf{Alphabet Size} & \textbf{Unsolv. Problems} \\
			\midrule
			\textbf{\citet{KuoKB20}} & LTL/Network-operators & LTL complexity & \xmark & DR & \xmark & $\leq 9 $ & \xmark \\ 
			\textbf{\citet{VaezipoorLIM21}} & LTL syntax tree & No curriculum & \xmark & DR & \cmark & $\leq 12 $ & \xmark \\
			\textbf{\citet{QiuMZ23}}                            & Propositions    & No curriculum & \xmark            & DR (prop.) & \xmark & $\leq 16$ & \xmark \\
			\textbf{\citet{YalcinkayaLVS23} }                             & DFA  & No curriculum & \xmark        & DR & \xmark &  $3$    & \xmark \\
			\textbf{\citet{YalcinkayaLVS24}} & cDFA & No curriculum & \xmark & DR (RAD) & \cmark & $\leq 12 $  & \xmark \\
			\textbf{\citet{JackermeierA25}} & Reach-avoid seq. & LTL complexity & \xmark  & DR & \xmark  & $\leq 20 $ & \xmark \\
			\textbf{\ourmethod{abbrv} (ours)} & RMs & RM \& level complexity & \cmark  & PLR, ACCEL & \xmark & $\leq 889$ & \cmark \\
			\bottomrule
		\end{tabular}
	}
	\caption{Comparison of problem-conditioned RL methods using formal task representations.}
	\label{tab:goal_cond_comp}
\end{table*}

\subsubsection{Sampling and Mutations.}
\emph{LTL-conditioned} methods~\cite[e.g.,][]{KuoKB20,VaezipoorLIM21,JackermeierA25} typically generate tasks by procedurally sampling LTL formulas combining logical and temporal operators from a context-free grammar. These methods build rules into the grammar that prevent logically unsatisfiable tasks. \citet{QiuMZ23} propose a method that trains from randomly sampled propositions instead of LTL tasks---the policies are proposition-conditioned and then ensembled together to satisfy LTL specifications.

In the context of \emph{automata-conditioned} methods, \citet{YalcinkayaLVS23} sample deterministic finite automata (DFA) representing reach-avoid tasks. The path length is specified within a range similar to ours~(3--7). The agents are trained from both random DFAs and DFAs derived from sequencing propositions observed in previous trajectories---a technique that could complement our existing hindsight edits. In a similar vein, \citet{OlivieriLGP25} use RM-based counterfactual reasoning to produce synthetic hindsight experiences. \citet{YalcinkayaLVS24} propose a family of DFAs called \emph{reach-avoid derived}~(RAD). The authors pretrain DFA encoders on this class and demonstrate zero-shot generalization to other DFAs. The key insight is that any path through a DFA can be seen as a series of reach-avoid sub-tasks. RAD DFAs are generated through mutations from reach-avoid DFAs. Each mutation involves randomly changing a transition, converting the accepting state into a sink, and minimizing the resulting DFA to ensure simplicity. A filtering step rejects resulting trivial, uninteresting tasks (e.g.,~single-state DFAs); hence, the training distribution consists of non-trivial, solvable tasks. \ourmethod{abbrv} does not perform the mutations on random samples, but on high-regret ones; further, it indirectly performs the filtering through regret scoring. Our random walk-based sampling strategy acts as an alternative to performing mutations on reach-avoid DFAs, still permitting the application of mutations over the samples. The authors also introduce cDFAs, a parallel composition of two DFAs that enables executing the DFAs in an interleaved manner. HRMs, the formalism underlying our implementation (see \cref{app:evaluation_suite}), sequentially composes automata at arbitrarily many hierarchical levels rather than in parallel. Previous work has considered parallel compositions of RMs in the multi-agent setting~\cite{NearyXWT21}.

Both \citet{VaezipoorLIM21} and \citet{YalcinkayaLVS24} rely on environment-agnostic \emph{pretraining} schemes to improve sample efficiency; that is, they pretrain graph embeddings leveraging task structure only, without interacting with the actual environment. In contrast, \ourmethod{abbrv} does not employ pretraining. We hypothesize that such pretraining could also help improve DR's performance in settings with low solvability and/or large alphabets. Since pretraining is orthogonal to curriculum generation via UED, examining their combination is an interesting avenue for future work.

\subsubsection{Curriculum.} 
The LTL-conditioned methods proposed by \citet{KuoKB20} and \citet{JackermeierA25} use a handcrafted curriculum that starts with simple tasks and gradually progresses to more complex tasks (e.g., longer formulas) as satisfactory performance is achieved. Similar curricula, combined with a small predefined set of tasks, have been previously used with policy sketches~\cite{AndreasKL17}, neural programs~\cite{PierrotLRS0LKBF19}, and hierarchies of reward machines~\cite{FurelosBlancoLJBR23}. In contrast, \ourmethod{abbrv} leverages regret-based UED to generate autocurricula of solvable yet challenging problems; further, unlike previous approaches, \ourmethod{abbrv} performs curricula over both tasks and levels rather than tasks alone.

\subsubsection{Evaluation.}
There are two key differences between \ourmethod{abbrv} and previous work. First, the problems tackled by previous LTL- and automata-conditioned methods often consist of small propositional alphabets, with the maximum being 20 in the work by \citet{JackermeierA25}; in contrast, \ourmethod{abbrv} employs an alphabet of 889 propositions. Second, as mentioned in \cref{sec:rw}, prior work employs levels where any proposition in the alphabet can be observed. For example, the commonly used \textsc{LetterWorld} domain contains 12 letters in a $7\times7$ grid, each corresponding to a proposition and appearing twice. Similarly, the also typical \textsc{Zones} domain determines four colors (i.e., four propositions) and any level has two regions for each of these colors. On the other hand, the problems we consider do not guarantee that the objects mentioned in an RM will appear or be reachable.

\section{Architecture Details}
\label{app:architecture}
The following paragraphs provide details on the problem-conditioned architecture introduced in \cref{sec:method_arch}.

\subsection{Environment Observation Encoding}
In the Minigrid implementation by \citet{NikulinKZASK24}, observations are $V \times V \times 2$, where $V=5$ is the size of the agent's field of view, the first dimension contains the object identifiers, and the second dimension contains the colors. If the agent is carrying an object, the object appears in the agent's position.

Following the example implementation by \citet{NikulinKZASK24}, the object and color identifiers in the observation are separately embedded into 16-dimensional vectors. The object and color embeddings are concatenated. The resulting $V\times V \times 32$ tensor is processed by a 3-layer convolutional neural network (CNN) with ReLU activations and 32, 64, and 64 filters, respectively. All kernels are $2\times 2$ with a stride of 1 and zero-padding.

\subsection{Reward Machine Encoding}
Given an RM $\rmname=\langle\rmstates,\propset,\rmtxstate,\rmtxreward,\rminit,\rmacc \rangle$ and one of its states $\rmstate\in\rmstates$, the encoding is performed with a \emph{graph neural network}~(GNN). The section is organized as follows. First, we describe the GNN architecture we use. Second, we describe how the graph passed to the GNN is constructed from the input RM. Finally, we describe the values of the GNN-related hyperparameters. Our \emph{implementation} is based on that from the $\mathtt{jraph}$ library~\cite{jraph2020github}.

\subsubsection{Graph Convolutional Networks \cite[GCNs;][]{KipfW17}.}
Given a graph $\graph=\langle\nodes, \edges\rangle$ with nodes $\node_i\in\nodes$ and edges $\langle\node_i,\node_j \rangle\in \edges$, GCNs update node representations $\nodefeat_i\in\realnums^\numnodefeats$ by aggregating those from neighboring nodes. Each node $\node_i$ is randomly initialized with features $\nodefeat^{(0)}_i\sim\normaldist(0,\variance \identity)$, where $\variance=0.1$~\citep{LiuKBKS19,AbboudCGL21}. Our GCN updates the node features through $\numlayers$ layers of message passing:
\begin{align*}
	\nodefeatl{\layer+1}{i}=\activ\left(\text{LN}^{(l)}\left(\layerweightsself{\layer} \nodefeatl{\layer}{i} + \layerweightsneigh{\layer}\sum_{j\in \neigh{i}}\frac{\nodefeatl{\layer}{j}\concat\edgefeat{j}{i}}{\card{\neigh{i}}}\right)\right),
\end{align*}
where $\neigh{i}$ denotes the neighbors of node $\node_i$, $\layerweightsself{\layer}\in\realnums^{\numnodefeats \times \numnodefeats}$ and $\layerweightsneigh{\layer}\in\realnums^{\numnodefeats \times (\numnodefeats + \numedgefeats)}$ are learnable parameters, $\edgefeat{j}{i}\in\realnums^\numedgefeats$ encodes the formula of edge $\langle\node_j,\node_i\rangle$, $\activ$ is an activation function, LN denotes layer normalization~\cite{BaKH16}, and $\concat$ denotes concatenation. The parameters $\layerweightsself{\layer}$ are used to capture the same node's features in the previous layer, while $\layerweightsneigh{\layer}$ are used for the neighboring nodes. We separate them since RM states do not have self-loops labeled by formulas. R-GCNs~\cite{SchlichtkrullKB18} handle this similarly by having a set of parameters for different relation types and specifying self-loops as a type of relation. The resulting features $\nodefeat^{(\numlayers)}_i$ undergo a last linear transformation $\lastweights\in\realnums^{\numnodefeats\times\numnodefeats}$.

\subsubsection{Graph Construction from a Reward Machine.}
The \emph{input graph} is constructed from an RM by reversing the edges, enabling each state to be contextualized by future states reachable within $\numlayers$ transitions. This design enables considering what needs to be accomplished far into the future, rather than being myopic (i.e.,~looking one step ahead). Previous approaches for conditioning policies on LTL formulas~\cite[e.g.,][]{VaezipoorLIM21} and finite-state machines~\cite[e.g.,][]{YalcinkayaLVS23,YalcinkayaLVS24} follow the same reasoning.

The \emph{edge features} $\edgefeat{j}{i}$ are built by (i)~embedding each literal (proposition or its negation) of the formula, (ii)~aggregating the literal embeddings by summing them, and (iii)~applying a linear transformation $\mathbf{W}_{\text{lit}}\in\realnums^{\numedgefeats\times\numedgefeats}$. We consider two methods for constructing the \emph{literal embeddings}:
\begin{itemize}
	\item \emph{Domain Independent.} Literals are embedded independently through an embedding matrix with $2N+1$ rows, where $N$ is the alphabet size. Negative literals correspond to rows $0,\ldots, N-1$; $\top$ (the value \emph{true}) corresponds to the $N$-th row; and positive literals correspond to rows $N+1,\ldots,2N$.
	\item \emph{Domain Dependent.} The proposition's location and object information are decomposed, embedded separately, and aggregated. \cref{fig:literal_emebdding} illustrates and describes how each literal embeddings are constructed.
\end{itemize}
In both cases, the literal embeddings are learnable.

\begin{figure}
	\begin{subfigure}[b]{0.5\linewidth}
		\centering
		\includegraphics[width=\linewidth]{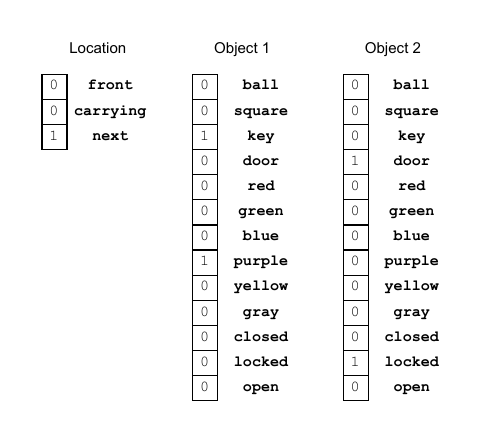}
		\caption{Encoding decomposition for the $\nextto\_\key\_\purple\_\door\_\locked$ proposition.}
	\end{subfigure}
	\hfill
	\begin{subfigure}[b]{0.46\linewidth}
		\centering
		\includegraphics[width=\linewidth]{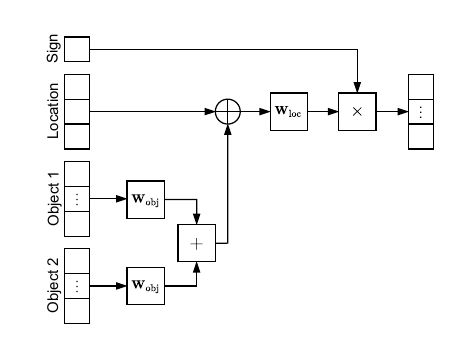}
		\caption{Construction of a literal embedding from the decomposed proposition encoding.}
	\end{subfigure}
	\caption{Phases in the literal embedding construction. \textbf{(a)} A binary encoding of the proposition is first constructed. \textbf{(b)} The literal embedding is built using the proposition decomposition. First, the binary encodings of the objects undergo a linear transformation $\mathbf{W}_\text{obj}\in\realnums^{\numedgefeats\times13}$ and their resulting representations are aggregated using an order-invariant operation (sum). Second, the location binary encoded is appended to the object representation and linearly transformed with $\mathbf{W}_\text{loc}\in\realnums^{\numedgefeats\times (\numedgefeats+3)}$. Finally, the sign of the literal (\num{1} if positive, \num{-1} if negative) is applied to produce the final embedding.}
	\label{fig:literal_emebdding}
\end{figure}

\subsubsection{Hyperparameters.}
The GCN consists of $\numlayers=5$ layers, chosen after the maximum number of transitions from the initial state $\rminit$ to the accepting state $\rmacc$ that can be sampled in our training setting. The number of node features is $\numnodefeats=128$ and the number of edge features is $\numedgefeats=64$. The activation functions are ReLUs.

\subsection{Encoding Aggregation and Actor-Critic Heads}
Given the encodings for the current observation and the current RM state, we concatenate them with a 16-dimensional embedding of the previous action. The resulting embedding is processed by a gated recurrent unit~\cite[GRU;][]{ChoMGBBSB14} producing 512 features. RMs implicitly encode a history over propositions, so the application of the GRU over the RM state encoding could be deemed unnecessary; however, we found that it experimentally works better.

The embedding---aggregating the observation, RM state, and action information---is then processed by the actor and critic heads, each consisting of two hidden layers with 256 rectifier units. The actor's output layer produces a logit for each action, while the critic's output layer produces a scalar estimating the value for the input observation, RM state, and action.

\section{Mutation Details}
\label{app:mutations}
In this section, we describe the preconditions and effects of each edit introduced in \cref{sec:method_ued}. These edits are conceptually applicable to any problem, i.e.~task-level pair. However, we focus on our main \emph{training} setting where (i)~RM tasks are sequential with at most 6 states, and (ii)~levels are structured into $\{1,2,4,6\}$ rooms of size $7\times 7$. One-room levels are $7\times 7$ and contain 1--5 objects, two-room levels are $7\times 13$ and contain 1--10 objects, four-room levels are $13\times13$ and contain 4--15 objects, and six-room levels are $13\times19$ and contain 7--20 objects. The minimum number of objects is determined as the maximum between 1 and the number of doors in the level.

\subsection{Level Edits}
There are eight types of level edits, illustrated in \cref{fig:level_edit_examples}:
\begin{description}
	\item[\addrooms] Applicable if the source level has 1, 2, or 4 rooms. Transforms one-room levels into two-room levels, two-room levels into four-room levels, and four-room levels into six-room levels. The extended side of the source level is determined uniformly at random---that is, given a one-room level, a room is added to the left or the right; given a two-room level, two rooms are added above or below; and given a six-room level, two rooms are added to the left or the right.
	\item[\rmrooms] Applicable if the source level has 2, 4, or 6 rooms. Transforms levels following the inverse order of \textsc{AddRooms}. The removed rooms cannot have the agent in them---given a two-room level, remove a room; given a four-room level, remove a row of rooms; and given a six-room level, remove the leftmost or rightmost column (if the agent is in the center column, choose one randomly).
	\item[\addobj] Applicable if the source level does not contain the maximum number of objects allowed (room-dependent, see above). A random non-door object (type and color) is determined and placed on a random free position.
	\item[\rmobj] Applicable if the source level does not contain the minimum number of objects. A random non-door object on the grid is removed.
	\item[\mvagent] Applicable if there are free locations, which is guaranteed by the maximum number of objects being lower than the number of locations per level. The agent is moved to a random free location with a randomly determined orientation.
	\item[\mvobj] Applicable if there are free locations. An existing non-door object is randomly chosen and moved to a random free location.
	\item[\rpdoor] Applicable if the level contains doors (i.e., more than one room). An existing door is randomly chosen and replaced with a new door whose color and status (locked, closed, open) are randomly determined.
	\item[\rpnondoor] Applicable if the level contains non-door objects. An existing non-door object is randomly chosen and replaced with a new random non-door object.
\end{description}

\begin{figure}
	\centering
	\begin{subfigure}{\linewidth}
		\centering
		\includegraphics[height=5em]{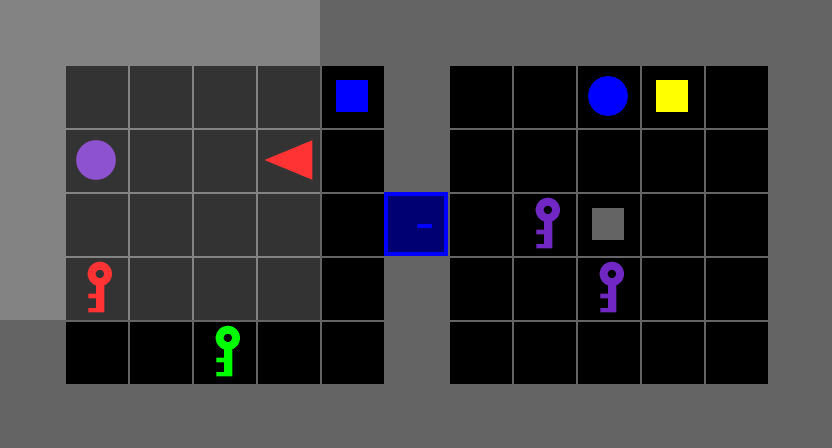}
		\caption{Source level}
	\end{subfigure}\\[1em]
	
	\begin{subfigure}[b]{0.245\linewidth}
		\centering
		\includegraphics[height=10em]{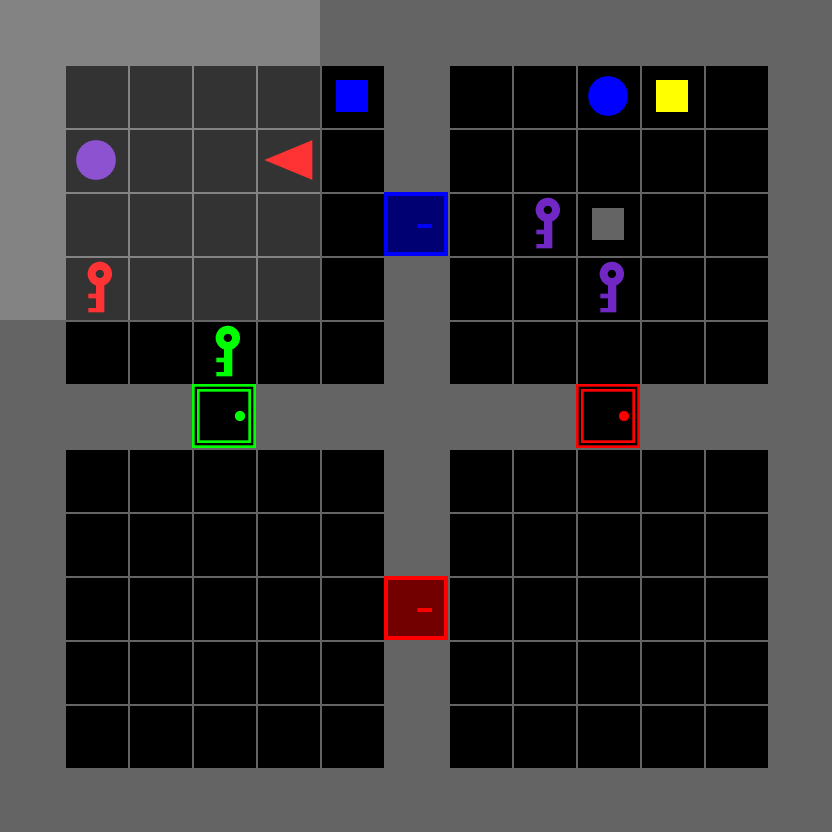}
		\caption{\addrooms}
	\end{subfigure}
	\begin{subfigure}[b]{0.245\linewidth}
		\centering
		\includegraphics[height=5em]{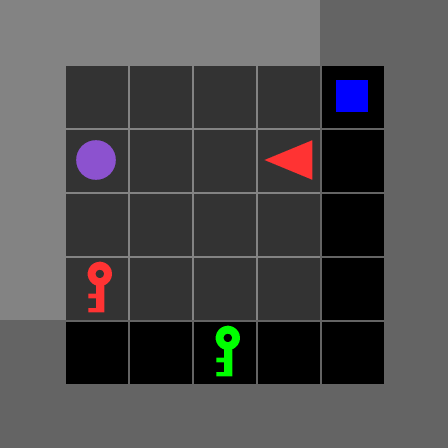}
		\caption{\rmrooms}
	\end{subfigure}
	\begin{subfigure}[b]{0.245\linewidth}
		\centering
		\includegraphics[height=5em]{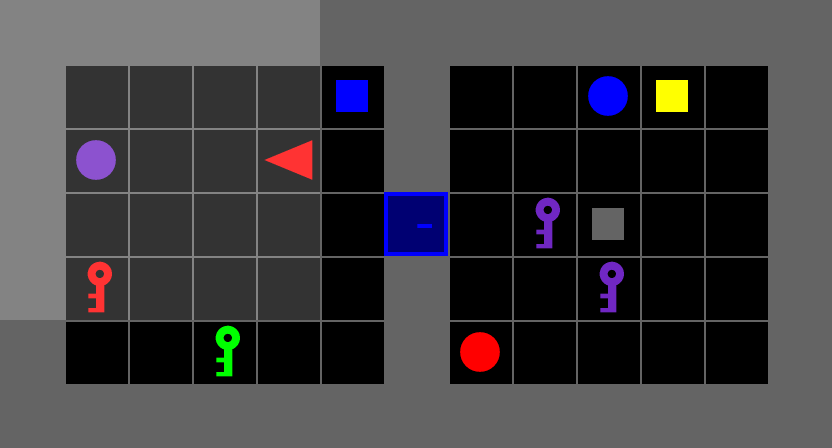}
		\caption{\addobj}
	\end{subfigure}
	\begin{subfigure}[b]{0.245\linewidth}
		\centering
		\includegraphics[height=5em]{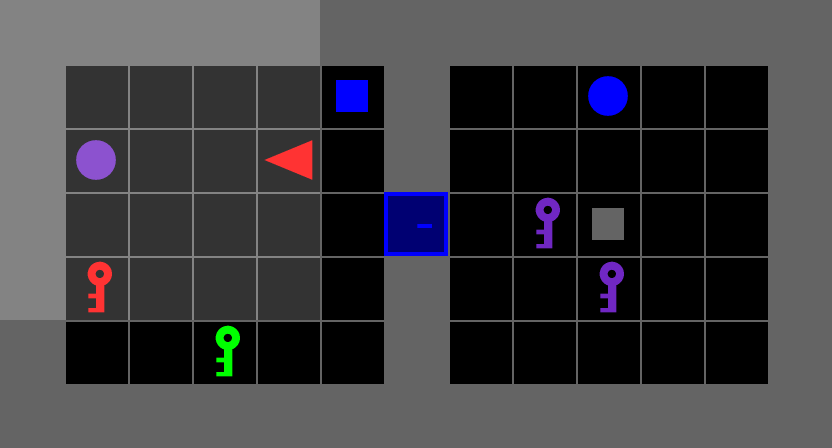}
		\caption{\rmobj}
	\end{subfigure}\\[1em]
	
	\begin{subfigure}[b]{0.245\linewidth}
		\centering
		\includegraphics[height=5em]{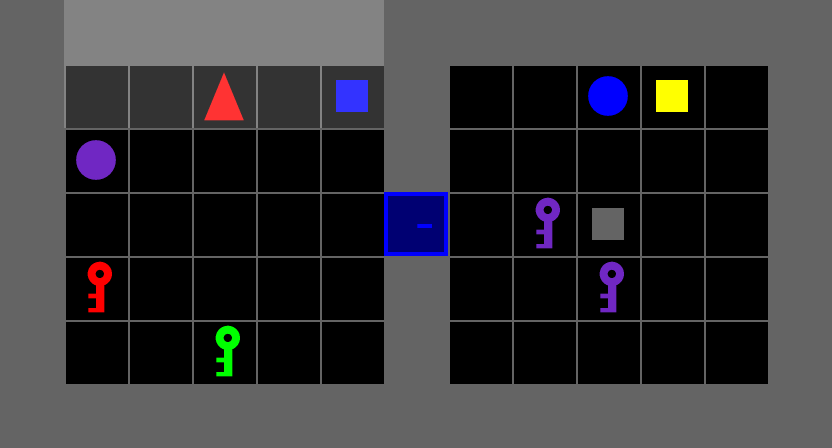}
		\caption{\mvagent}
	\end{subfigure}
	\begin{subfigure}[b]{0.245\linewidth}
		\centering
		\includegraphics[height=5em]{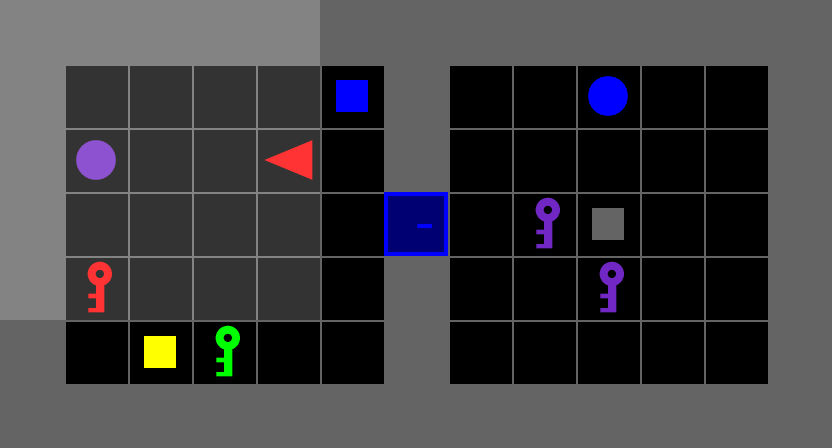}
		\caption{\mvobj}
	\end{subfigure}
	\begin{subfigure}[b]{0.245\linewidth}
		\centering
		\includegraphics[height=5em]{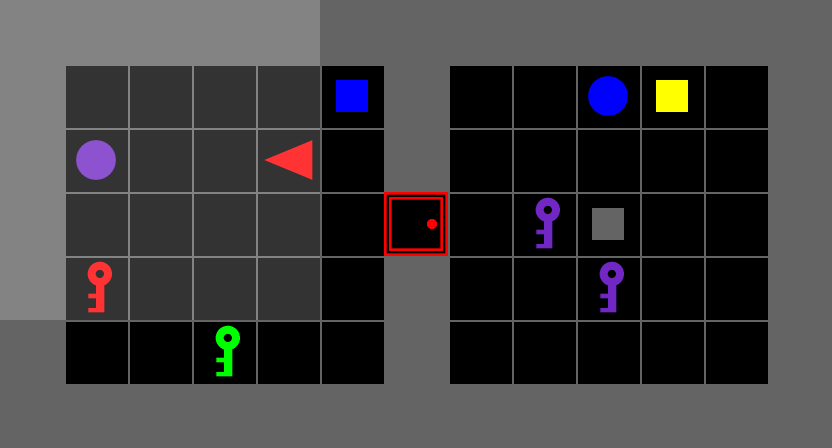}
		\caption{\rpdoor}
	\end{subfigure}
	\begin{subfigure}[b]{0.245\linewidth}
		\centering
		\includegraphics[height=5em]{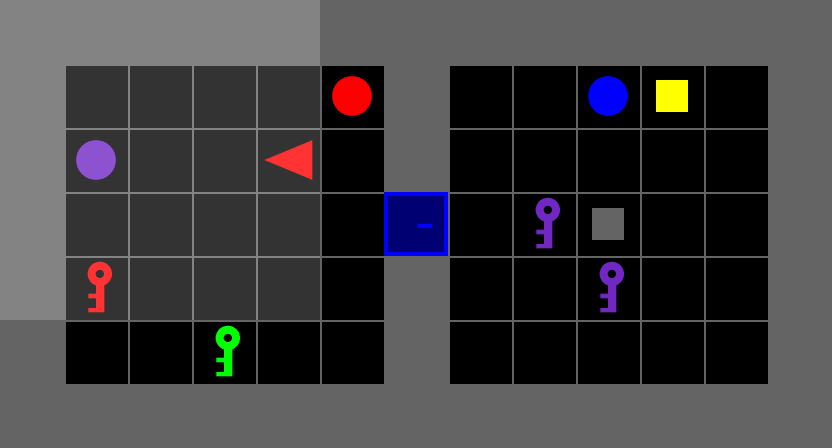}
		\caption{\rpnondoor}
	\end{subfigure}
	\caption{Level edit examples.}
	\label{fig:level_edit_examples}
\end{figure}

\subsection{Task Edits}
There are three types of task edits, illustrated in \cref{fig:task_edit_examples}:
\begin{description}
	\item[\switchprop] Always applicable. An edge is randomly selected, and its proposition is replaced by another one chosen uniformly at random from the alphabet.
	\item[\addstate] Applicable if the number of states is less than the maximum. Adds a new state at the start, middle, or end of the existing state sequence. If inserted in the middle, transitions are rewired to maintain connectivity. The outgoing transition from the new state is labeled with a proposition chosen uniformly at random from the whole alphabet.
	\item[\remstate] Applicable if there are more than two states. Removes one of the states and, if it is a non-accepting state, its outgoing transition as well. If the removed state is the initial state, the state pointed to by it becomes the new initial state. If the removed state is the accepting state, the state that pointed to it becomes the new accepting state.
\end{description}
The reward transition function of the resulting RM is adjusted to make it \emph{sparse} (reward of 1 in transitions to $\rmacc$), which is the training default (see \cref{sec:experiments_setup}).

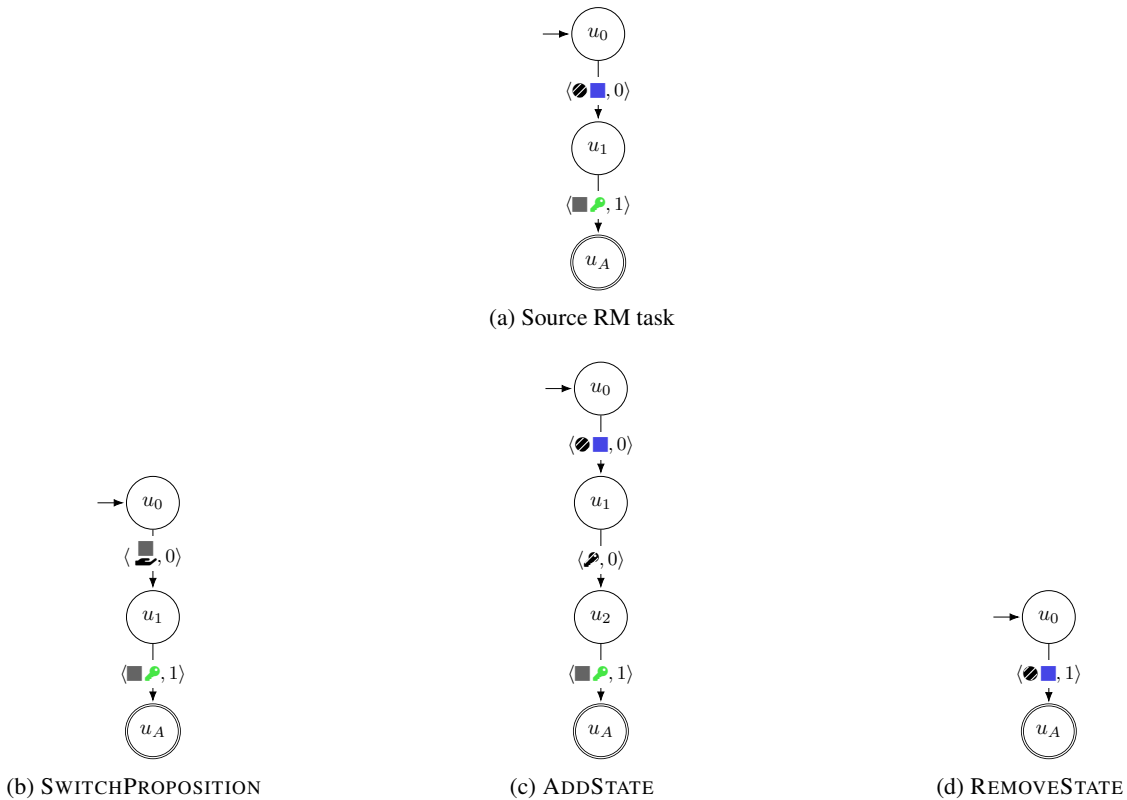
\begin{figure}
	\centering
	\begin{subfigure}{\linewidth}
		\centering
		\scalebox{0.8}{
			\begin{tikzpicture}[shorten >=1pt,node distance=1.9cm,on grid,auto,every initial by arrow/.style ={-Latex}]
				\node[state,initial,initial text=] (u_0)   {$\rminit$};
				\node[state] (u_1) [below = of u_0]   {$\rmidx{1}$};
				\node[state,accepting] (u_acc) [below = of u_1]  {$\rmacc$};
				
				\path[-Latex] (u_0) edge node[in place] {\small$\langle\text{\icnexthor{\icnocolor{\iccircle}}{\iccolor{\icsquare}{icblue}}},0\rangle$} (u_1);
				\path[-Latex] (u_1) edge node[in place] {\small$\langle\text{\icnexthor{\iccolor{\icsquare}{icgray}}{\iccolor{\ickey}{icgreen}}},1\rangle$} (u_acc);
			\end{tikzpicture}
		}
		\caption{Source RM task}
	\end{subfigure}\\[1em]
	
	\begin{subfigure}[b]{0.33\linewidth}
		\centering
		\scalebox{0.8}{
			\begin{tikzpicture}[shorten >=1pt,node distance=1.9cm,on grid,auto,every initial by arrow/.style ={-Latex}]
				\node[state,initial,initial text=] (u_0)   {$\rminit$};
				\node[state] (u_1) [below = of u_0]   {$\rmidx{1}$};
				\node[state,accepting] (u_acc) [below = of u_1]  {$\rmacc$};
				
				\path[-Latex] (u_0) edge node[in place,pos=0.4] {\small$\langle\text{\iccarrying{\iccolor{\icsquare}{icgray}}},0\rangle$} (u_1);
				\path[-Latex] (u_1) edge node[in place] {\small$\langle\text{\icnexthor{\iccolor{\icsquare}{icgray}}{\iccolor{\ickey}{icgreen}}},1\rangle$} (u_acc);
		\end{tikzpicture}}
		\caption{\switchprop}
	\end{subfigure}
	\begin{subfigure}[b]{0.33\linewidth}
		\centering
		\scalebox{0.8}{
			\begin{tikzpicture}[shorten >=1pt,node distance=1.9cm,on grid,auto,every initial by arrow/.style ={-Latex}]
				\node[state,initial,initial text=] (u_0)   {$\rminit$};
				\node[state] (u_1) [below = of u_0]   {$\rmidx{1}$};
				\node[state] (u_2) [below = of u_1]   {$\rmidx{2}$};
				\node[state,accepting] (u_acc) [below = of u_2]  {$\rmacc$};
				
				\path[-Latex] (u_0) edge node[in place] {\small$\langle\text{\icnexthor{\icnocolor{\iccircle}}{\iccolor{\icsquare}{icblue}}},0\rangle$} (u_1);
				\path[-Latex] (u_1) edge node[in place] {\small$\langle\text{\icnocolor{\ickey}},0\rangle$} (u_2);
				\path[-Latex] (u_2) edge node[in place] {\small$\langle\text{\icnexthor{\iccolor{\icsquare}{icgray}}{\iccolor{\ickey}{icgreen}}},1\rangle$} (u_acc);
		\end{tikzpicture}}
		\caption{\addstate}
	\end{subfigure}
	\begin{subfigure}[b]{0.33\linewidth}
		\centering
		\scalebox{0.8}{
			\begin{tikzpicture}[shorten >=1pt,node distance=1.9cm,on grid,auto,every initial by arrow/.style ={-Latex}]
				\node[state,initial,initial text=] (u_0)   {$\rminit$};
				\node[state,accepting] (u_acc) [below = of u_0]  {$\rmacc$};
				
				\path[-Latex] (u_0) edge node[in place] {\small$\langle\text{\icnexthor{\icnocolor{\iccircle}}{\iccolor{\icsquare}{icblue}}},1\rangle$} (u_1);
		\end{tikzpicture}}
		\caption{\remstate}
	\end{subfigure}
	
	\caption{Task edit examples. Zero-reward self-transitions are omitted for simplicity.}
	\label{fig:task_edit_examples}
\end{figure}

\subsection{Hindsight Edits}
These edits are applicable if the last RM state $\rmstate\in\rmstates$ in a rollout is not the initial or the accepting state. Further, they can only be selected as the first step of the edit sequence. There are two edit types, illustrated in \cref{fig:hindsight_edit_examples}, both of which jointly modify the level and the task:
\begin{description}
	\item[\hindpred] Keeps the original level and derives a new RM with $u$ as the accepting state.
	\item[\hindsucc] Derives a new level from the last environment state and a new RM with $\rmstate$ as the initial state.
\end{description}

\begin{figure}
	\centering
	\begin{subfigure}{0.495\linewidth}
		\centering
		
		\includegraphics[height=6em]{figures/level_edits/src_lvl.png}
		~
		\scalebox{0.8}{
			\begin{tikzpicture}[shorten >=1pt,node distance=1.9cm,on grid,auto,every initial by arrow/.style ={-Latex}]
				\node[state,initial,initial text=] (u_0)   {$\rminit$};
				\node[state] (u_1) [below = of u_0]   {$\rmidx{1}$};
				\node[state,accepting] (u_acc) [below = of u_1]  {$\rmacc$};
				
				\path[-Latex] (u_0) edge node[in place] {\small$\langle\text{\icnexthor{\icnocolor{\iccircle}}{\iccolor{\icsquare}{icblue}}},0\rangle$} (u_1);
				\path[-Latex] (u_1) edge node[in place] {\small$\langle\text{\icnexthor{\iccolor{\icsquare}{icgray}}{\iccolor{\ickey}{icgreen}}},1\rangle$} (u_acc);
			\end{tikzpicture}
		}
		\caption{Source problem}
	\end{subfigure}
	\begin{subfigure}{0.495\linewidth}
		\centering
		\includegraphics[height=6em]{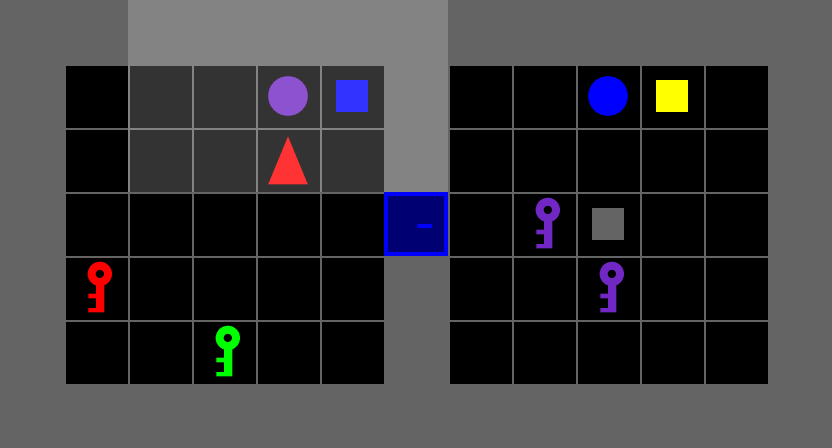}
		~
		\scalebox{0.8}{
			\begin{tikzpicture}[shorten >=1pt,node distance=1.9cm,on grid,auto,every initial by arrow/.style ={-Latex}]
				\node[state,initial,initial text=] (u_0)   {$\rminit$};
				\node[state,fill=lightgray] (u_1) [below = of u_0]   {$\rmidx{1}$};
				\node[state,accepting] (u_acc) [below = of u_1]  {$\rmacc$};
				
				\path[-Latex] (u_0) edge node[in place] {\small$\langle\text{\icnexthor{\icnocolor{\iccircle}}{\iccolor{\icsquare}{icblue}}},0\rangle$} (u_1);
				\path[-Latex] (u_1) edge node[in place] {\small$\langle\text{\icnexthor{\iccolor{\icsquare}{icgray}}{\iccolor{\ickey}{icgreen}}},1\rangle$} (u_acc);
			\end{tikzpicture}
		}
		\caption{Environment state and RM state at the end of a rollout}
	\end{subfigure}\\[1em]
	
	\begin{subfigure}{0.495\linewidth}
		\centering
		
		\includegraphics[height=6em]{figures/level_edits/src_lvl.png}
		~
		\scalebox{0.8}{
			\begin{tikzpicture}[shorten >=1pt,node distance=1.9cm,on grid,auto,every initial by arrow/.style ={-Latex}]
				\node[state,initial,initial text=] (u_0)   {$\rminit$};
				\node[state,accepting] (u_acc) [below = of u_0]  {$\rmacc$};
				
				\path[-Latex] (u_0) edge node[in place] {\small$\langle\text{\icnexthor{\icnocolor{\iccircle}}{\iccolor{\icsquare}{icblue}}},1\rangle$} (u_acc);
			\end{tikzpicture}
		}
		\caption{\hindpred}
	\end{subfigure}
	\begin{subfigure}{0.495\linewidth}
		\centering
		
		\includegraphics[height=6em]{figures/level_edits/end_lvl.png}
		~
		\scalebox{0.8}{
			\begin{tikzpicture}[shorten >=1pt,node distance=1.9cm,on grid,auto,every initial by arrow/.style ={-Latex}]
				\node[state,initial,initial text=] (u_0)   {$\rminit$};
				\node[state,accepting] (u_acc) [below = of u_0]  {$\rmacc$};
				
				\path[-Latex] (u_0) edge node[in place] {\small$\langle\text{\icnexthor{\iccolor{\icsquare}{icgray}}{\iccolor{\ickey}{icgreen}}},1\rangle$} (u_acc);
			\end{tikzpicture}
		}
		\caption{\hindsucc}
	\end{subfigure}
	
	\caption{Hindsight edit examples.}
	\label{fig:hindsight_edit_examples}
\end{figure}

\section{Experimental Details}
\label{app:experiments}
In this section, we explain our experimental setup, hyperparameters, and additional results for the experiments described in \cref{sec:experiments}. Our codebase is fully implemented in JAX~\cite{jax2018github}. We extended the JaxUED~\cite{CowardBF24}, a collection of UED algorithm implementations in JAX, to support problems (levels and tasks) rather than only levels. Our implementation of the partially observable Minigrid environments is based on that by \citet{NikulinKZASK24}, introducing additional complexity compared to the fully observable environments often used in prior formal language-conditioned work and providing a foundation for future research in this direction. Despite their apparent simplicity, executing BabyAI instructions in Minigrid is still challenging; for example, LLMs and VLMs suffer from \emph{``significant shortcomings in the agents’ ability to place objects adjacent to other objects''}~\cite{PaglieriCCPWKPK25}.

The experiments were executed across three different Linux clusters, respectively consisting of (i)~NVIDIA RTX 6000 Ada Generation GPUs and Intel Xeon Gold 6348 CPUs, (ii)~NVIDIA L40S GPUs and Intel Xeon Platinum 8358 CPUs, and (iii)~NVIDIA A100-SXM4-80GB GPUs and Intel Xeon Platinum 8562 CPUs. The software dependencies are listed in the code appendix (see the $\mathtt{requirements.txt}$ file). For all experiments, we reserved 32GB of RAM. Our results report aggregate performance across 5 seeds.

\subsection{Hyperparameters}
\label{app:experiments_hyperparams}
\cref{tab:hyperparameters} shows the table containing the RL-related hyperparameters used in the final experiments, for which we use the Adam optimizer~\cite{KingmaB14}. We refer the reader to \cref{app:architecture} for the architectural details.

To determine the final hyperparameter values, we sampled a random level-conditioned problem set. The problems underwent a solvability check, which we describe in \cref{app:experiments_percentage_solvable}. The RMs in the set are all sequential. The problem set is constituted by 300 samples that result from taking 5 samples for each combination of: number of rooms (1, 2, 4, 6), number of objects and number of transitions (1--5). For each number of rooms, we distinguish three object intervals:
\begin{itemize}
	\item One room: 1--2, 3--4, 5.
	\item Two rooms: 1--3, 4--7, 8--10.
	\item Four rooms: 4--7, 8--11, 12--15.
	\item Six rooms: 7--10, 11--16, 17--20.
\end{itemize} 
This set is exclusively used for validation.

The hyperparameter sweep was as follows. We performed a grid search for \rplr across the buffer capacity $\{25000, 50000\}$, the replay probability $\{0.5, 0.9\}$, the PPO entropy loss coefficient $\{0.0, 0.001, 0.01\}$, and whether to feed the RM embedding into the RNN together with the observation embedding. For ACCEL, we fixed a replay probability of 0.9, a buffer size of 50000, and a maximum number of 10 edits, and repeated the same grid search over the same remaining hyperparameters while sweeping over the minimum number of edits $\{3,5,7,10\}$. In hyperparameter searches preceding the final one, we swept also over the staleness coefficient $\{0.1, 0.3\}$, discount rate $\{0.9, 0.95, 0.98, 0.99\}$, GAE lambda $\{0.9, 0.95, 0.98\}$, sampling temperature $\{0.3, 1.0\}$, learning rate $\{\num{1e-5}, \num{5e-5}\}$, and value loss coefficient $\{0.5, 0.75\}$.

\begin{table}[h]
	\centering
	\begin{tabular}{lr}
		\toprule
		\multicolumn{1}{c}{Parameter} & \multicolumn{1}{c}{Value} \\
		\midrule
		\emph{Environment}            & \\
		~~Maximum episode length         & 512 \\
		\midrule
		\emph{PPO} & \\
		~~\# Environment steps (Seq.) & \num{4194304000}\\
		~~\# Environment steps (DAG)  & \num{6291456000} \\
		~~Discount rate $\disc$       & \num{0.99} \\
		~~GAE $\gae$                  & \num{0.9} \\
		~~\# Parallel environments    & \num{4096} \\
		~~PPO rollout length          & \num{512} \\
		~~PPO epochs                  & \num{4} \\
		~~PPO minibatches per epoch   & \num{128} \\
		~~PPO clip range              & \num{0.2} \\
		~~PPO max. gradient norm      & \num{0.5} \\
		~~Adam learning rate          & \num{5e-5} \\
		~~Adam $\epsilon$             & \num{1e-5} \\
		~~Value loss coefficient      & \num{0.5} \\
		~~Entropy loss coefficient    & \num{0.01} \\
		\midrule
		$\textit{PLR}^\bot$ \\
		~~Buffer capacity             & \num{50000} \\
		~~Prioritization              & rank \\
		~~Temperature                 & \num{1.0} \\
		~~Replay rate                 & \num{0.5} \\
		~~Staleness coefficient       & \num{0.1} \\
		~~Score function              & MaxMC \\
		\midrule
		\emph{ACCEL} \\
		~~Replay rate (full)          & \num{0.9} \\
		~~Number of edits             & $\unidist\{7,10\}$ \\
		\midrule
		\emph{ACCEL-0} \\
		~~Replay rate                 & \num{0.99} \\
		~~Number of edits             & $\unidist\{7,10\}$ \\
		\bottomrule
	\end{tabular}
	\caption{List of environment, PPO, \rplr and ACCEL hyperparameters.}
	\label{tab:hyperparameters}
\end{table}

\subsection{Experimental Setup Details}
\label{app:experiments_setup}
We describe some omitted details from the experimental setup description in \cref{sec:experiments_setup}.

\subsubsection{Level Generation.}
Training levels (see \cref{sec:evaluation_suite_levels}) are generated as follows:
\begin{enumerate}
	\item A \emph{number of rooms} is uniformly sampled from $\{1,2,4,6\}$. Rooms are $5\times5$, surrounded by walls (total dimension $7\times7)$. Two-room levels are $7\times13$, four-room levels are $13\times13$, and six-room levels are $13\times19$.
	Levels with more than one room have doors between each pair of adjacent rooms, always in the middle of their dividing wall.
	\item A \emph{number of objects} is sampled uniformly from a grid-dependent range. One-room levels can have 1--5 objects, two-room levels can have 1--10 objects, four-room levels can have 4--15 objects, and six-room levels can have 7--20 objects. The minimum is determined by the number of doors.
	\item \emph{Non-door} objects are generated by choosing their type, color, and location uniformly at random. \emph{Door} objects are analogously generated by choosing their state (open, closed, locked) and color.
	\item The \emph{agent} is randomly placed in a free position.
\end{enumerate}

\subsubsection{Metrics.}
The CVaR problem set consists of \num{10000} samples, as recommended by \citet{RutherfordBWLHF24}, generated with the \emph{level-conditioned} sampler described in \cref{sec:evaluation_suite_sampling}. Sampling problems conditionally on the level increases the solvability rate (see \cref{app:experiments_percentage_solvable}); hence, although results do not substantially differ, the level-conditioned set covers a more diverse set of problems than if the level and the task were independently sampled. The problems are guaranteed to be \emph{solvable} using the process described in \cref{app:experiments_percentage_solvable}.

\subsection{Fraction of Solvable Randomly Generated Problems}
\label{app:experiments_percentage_solvable}
We analyze the fraction of solvable problems across batches sampled via the different sampling strategies. We report the average solvability and the standard deviation across 5 batches, where each batch consists of 4096 problems, as in our experiments.

To evaluate whether a problem is \emph{solvable}, we decompose the RM task into paths to the accepting state and determine if the formulas along these paths are satisfiable. An RM is considered solvable if the formulas in at least one path are satisfiable. A formula is considered satisfiable if the objects associated with it are reachable by the agent. For example, the formula $\front\_\ball$ is satisfiable if there is a ball within the reach of the agent. Reachability is determined by locked doors---hence, if a given formula cannot be satisfied by the objects within reach, we select a locked door whose color matches a key within reach. This procedure derives a tree where each child node increases the reachability with respect to its parent. Maintaining a tree, keeping different orderings on the opening of the locked doors, is important because only some of them might guarantee solvability. For instance, if there is a single green locked door and the reachability procedure opens it, a subsequent formula $\front\_\door\_\green\_\locked$ will not be satisfiable.

\cref{tab:percent_valid} shows the results for the cases where the \emph{task sampler} generates sequential and directed acyclic graphs, and the \emph{problem sampler} generates levels and tasks independently, or conditionally on the level. The instantiation of the samplers is as explained in \cref{sec:experiments_setup}. We observe that independent sampling produces batches with barely any solvable problems (around 3--4\%), whereas level-conditioned sampling produces a large majority of solvable problems (around 85\%). As shown in \cref{sec:experiments_problem_sampling_ablation_results}, increasing the number of solvable problems per batch results in a sensible performance improvement for DR, which closes the gap with respect to \rplr and ACCEL.

\cref{tab:percent_valid_sequential} decomposes the results for the \emph{sequential} RM sampler across: problem sampler (independent, conditioned), number of transitions (1--5), number of rooms (1, 2, 4, 6), and number of objects (L, M, H). The symbols for the number of objects denote different intervals (low, medium, high) depending on the number of rooms:
\begin{itemize}
	\item One room: L (1--2), M (3--4), H (5).
	\item Two rooms: L (1--3), M (4--7), H (8--10).
	\item Four rooms: L (4--7), M (8--11), H (12--15).
	\item Six rooms: L (7--10), M (11--16), H (17--20).
\end{itemize}
Following the trend from \cref{tab:percent_valid}, we observe that sampling conditionally on the level results in a substantially higher fraction of solvable problems. In the independent case, having fewer transitions and more rooms (and, hence, more objects) results in more solvable problems since there is a higher chance the transitions will refer to objects in the level. The percentage of solvable problems is extremely low, showing the potential of UED approaches in these settings. We emphasize that these results are illustrative and that none of these combinations are exclusively used in training the policies; instead, given a fixed problem sampling strategy (independent, conditioned), we determine the number of transitions, number of rooms, and number of objects uniformly at random.

\begin{table}[]
	\centering
	\begin{tabular}{lrr}
		\toprule
		\multicolumn{1}{c}{Problem Sampling} & \multicolumn{1}{c}{Sequential} & \multicolumn{1}{c}{Directed Acyclic Graph} \\
		\midrule
		Independent                          & $2.7\pm0.3$                        & $3.9\pm0.2$ \\
		Level-Conditioned                    & $83.4\pm0.4$                     & $84.8\pm0.5$ \\
		\bottomrule
	\end{tabular}
	\caption{Percentage of solvable problems when the RMs are sequential and directed acyclic graphs.}
	\label{tab:percent_valid}
\end{table}

\begin{table}[]
	\centering
	\resizebox{\textwidth}{!}{%
		\begin{tabular}{ccrrrrrrrrrrrrrrrr}
			\toprule
			&    & \multicolumn{3}{c}{1} && \multicolumn{3}{c}{2} && \multicolumn{3}{c}{4} && \multicolumn{3}{c}{6}\\
			\cmidrule{3-5} \cmidrule{7-9} \cmidrule{11-13} \cmidrule{15-17}
			&    & \multicolumn{1}{c}{L} & \multicolumn{1}{c}{M} & \multicolumn{1}{c}{H}             && \multicolumn{1}{c}{L} & \multicolumn{1}{c}{M} & \multicolumn{1}{c}{H}             && \multicolumn{1}{c}{L} & \multicolumn{1}{c}{M} & \multicolumn{1}{c}{H}             && \multicolumn{1}{c}{L} & \multicolumn{1}{c}{M} & \multicolumn{1}{c}{H}\\
			\midrule
			\multirow{5}{*}{\rotatebox[origin=c]{90}{Independent}} & 1  & $0.9 \pm 0.1$ & $2.6 \pm 0.2$ & $4.4 \pm 0.3$ && $2.2 \pm 0.1$ & $7.6 \pm 0.3$ & $12.1 \pm 0.2$ && $5.2 \pm 0.3$ & $14.6 \pm 0.6$ & $22.4 \pm 0.3$ && $6.7 \pm 0.3$ & $20.0 \pm 0.4$ & $30.9 \pm 0.8$ \\
			& 2  & $0.0 \pm 0.0$ & $0.0 \pm 0.0$ & $0.2 \pm 0.1$ && $0.1 \pm 0.0$ & $0.6 \pm 0.1$ & $1.5 \pm 0.3$ && $0.3 \pm 0.1$ & $2.5 \pm 0.2$ & $5.6 \pm 0.4$ && $0.6 \pm 0.2$ & $4.4 \pm 0.1$ & $10.6 \pm 0.3$ \\
			& 3  & $0.0 \pm 0.0$ & $0.0 \pm 0.0$ & $0.0 \pm 0.0$ && $0.0 \pm 0.0$ & $0.1 \pm 0.0$ & $0.3 \pm 0.0$ && $0.0 \pm 0.0$ & $0.5 \pm 0.1$ & $1.4 \pm 0.1$ && $0.1 \pm 0.0$ & $1.2 \pm 0.1$ & $4.2 \pm 0.2$ \\
			& 4  & $0.0 \pm 0.0$ & $0.0 \pm 0.0$ & $0.0 \pm 0.0$ && $0.0 \pm 0.0$ & $0.0 \pm 0.0$ & $0.1 \pm 0.0$ && $0.0 \pm 0.0$ & $0.1 \pm 0.1$ & $0.4 \pm 0.1$ && $0.0 \pm 0.0$ & $0.3 \pm 0.0$ & $1.5 \pm 0.1$ \\
			& 5  & $0.0 \pm 0.0$ & $0.0 \pm 0.0$ & $0.0 \pm 0.0$ && $0.0 \pm 0.0$ & $0.0 \pm 0.0$ & $0.0 \pm 0.0$ && $0.0 \pm 0.0$ & $0.0 \pm 0.0$ & $0.1 \pm 0.1$ && $0.0 \pm 0.0$ & $0.2 \pm 0.1$ & $0.6 \pm 0.1$ \\
			\midrule
			\multirow{5}{*}{\rotatebox[origin=c]{90}{Conditioned}} & 1  & $100.0 \pm 0.0$ & $100.0 \pm 0.0$ & $100.0 \pm 0.0$ && $90.5 \pm 0.4$ & $85.1 \pm 0.5$ & $86.6 \pm 0.6$ && $84.6 \pm 0.6$ & $84.0 \pm 0.5$ & $86.0 \pm 0.4$ && $82.3 \pm 0.4$ & $82.1 \pm 0.6$ & $85.8 \pm 0.4$ \\
			& 2  & $100.0 \pm 0.0$ & $100.0 \pm 0.0$ & $100.0 \pm 0.0$ && $87.9 \pm 0.5$ & $79.5 \pm 0.7$ & $80.2 \pm 0.4$ && $79.5 \pm 0.7$ & $78.4 \pm 0.5$ & $81.3 \pm 0.7$ && $76.7 \pm 0.5$ & $76.5 \pm 0.6$ & $80.9 \pm 0.5$ \\
			& 3  & $100.0 \pm 0.0$ & $100.0 \pm 0.0$ & $100.0 \pm 0.0$ && $87.0 \pm 0.4$ & $76.4 \pm 0.8$ & $76.8 \pm 0.7$ && $77.2 \pm 0.7$ & $76.1 \pm 0.8$ & $78.6 \pm 0.6$ && $73.8 \pm 0.6$ & $73.4 \pm 0.5$ & $78.5 \pm 0.1$ \\
			& 4  & $100.0 \pm 0.0$ & $100.0 \pm 0.0$ & $100.0 \pm 0.0$ && $86.5 \pm 0.5$ & $74.4 \pm 0.7$ & $74.6 \pm 0.6$ && $75.8 \pm 0.8$ & $74.1 \pm 0.7$ & $76.7 \pm 0.6$ && $72.0 \pm 0.7$ & $71.4 \pm 0.4$ & $76.8 \pm 0.3$ \\
			& 5  & $100.0 \pm 0.0$ & $100.0 \pm 0.0$ & $100.0 \pm 0.0$ && $86.1 \pm 0.6$ & $73.0 \pm 0.5$ & $73.2 \pm 0.8$ && $74.7 \pm 0.4$ & $72.4 \pm 0.6$ & $75.0 \pm 0.6$ && $70.8 \pm 0.9$ & $69.9 \pm 0.5$ & $75.1 \pm 0.1$ \\
			\bottomrule
		\end{tabular}
	}
	\caption{Percentage of solvable samples produced by the sequential sampler by fixing the problem sampler (independent, level-conditioned), number of transitions (1--5), number of rooms (1, 2, 4, 6), and a range for the number of objects (L, M, H).}
	\label{tab:percent_valid_sequential}
\end{table}

\subsection{Extended Main Results}
\label{app:experiments_main_results}
We provide some more details on the core results described in \cref{sec:experiments_main_results}.

\subsubsection{Performance.}
\cref{fig:solvability_over_time} shows the percentage of solvable problems in the buffer throughout training. As training progresses, \rplr, ACCEL, and ACCEL-0 curate a buffer mostly consisting of solvable problems. The performance for each individual hand-designed problem is reported in \cref{app:experiments_per_prob_results}.

\begin{figure}
	\centering
	\includegraphics[width=0.5\linewidth]{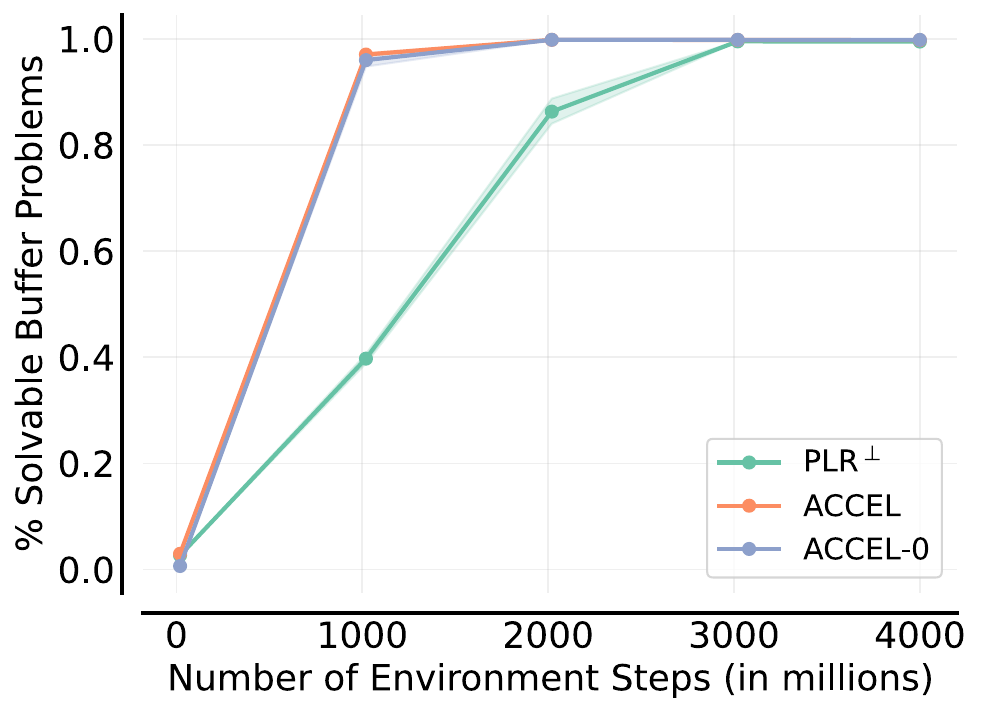}
	\caption{Fraction of solvable buffer problems throughout training in the \emph{independent} sampling setting.}
	\label{fig:solvability_over_time}
\end{figure}

\subsubsection{Curriculum Analysis.}
\cref{fig:gen_prob_dr,fig:gen_prob_plr,fig:gen_prob_accel_full,fig:gen_prob_accel_scratch} show a sequence of training problems for DR, \rplr, ACCEL and ACCEL-0. In the case of the last three approaches, these problems are samples with high probability from the buffer, gradually become more complex, and are often solvable. In the case of DR, the generated problems are of arbitrary complexity and often unsolvable.

\begin{figure}
	\centering
	\begin{subfigure}[b]{.245\textwidth}
		\centering
		\includegraphics[height=4em]{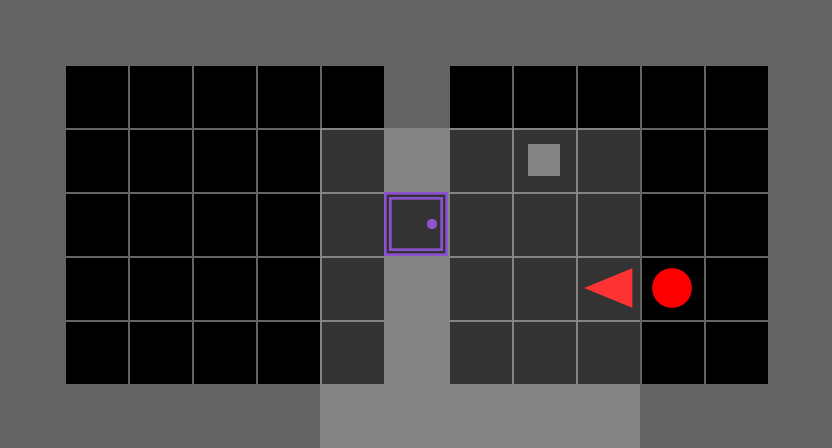}\\[1em]
		\scalebox{0.7}{
			\begin{tikzpicture}[shorten >=1pt,node distance=1.9cm,on grid,auto,every initial by arrow/.style ={-Latex}]
				\node[state,initial,initial text=] (u_0)   {$\rminit$};
				\node[state] (u_1) [below = of u_0]  {$\rmidx{1}$};
				\node[state] (u_2) [below = of u_1]  {$\rmidx{2}$};
				\node[state,accepting] (u_acc) [below = of u_2]  {$\rmacc$};
				
				\path[-Latex] (u_0) edge node[in place] {\small$\langle\text{\icnexthor{\iccolor{\ickey}{icyellow}}{\iccolor{\icdooropen}{icpurple}}},0\rangle$} (u_1);
				\path[-Latex] (u_1) edge node[in place] {\small$\langle\text{\icnexthor{\iccolor{\ickey}{icgray}}{\iccolor{\icdooropen}{icgreen}}},0\rangle$} (u_2);
				\path[-Latex] (u_2) edge node[in place] {\small$\langle\text{\icnexthor{\icnocolor{\icsquare}}{\iccolor{\ickey}{icyellow}}},1\rangle$} (u_acc);
			\end{tikzpicture}
		}
		
		\caption{$\num{1e9}$}
	\end{subfigure}
	\begin{subfigure}[b]{.245\textwidth}
		\centering
		\includegraphics[height=4em]{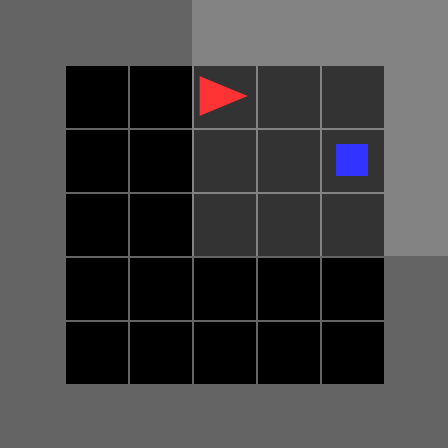}\\[1em]
		\scalebox{0.7}{
			\begin{tikzpicture}[shorten >=1pt,node distance=1.9cm,on grid,auto,every initial by arrow/.style ={-Latex}]
				\node[state,initial,initial text=] (u_0)   {$\rminit$};
				\node[state] (u_1) [below = of u_0]  {$\rmidx{1}$};
				\node[state] (u_2) [below = of u_1]  {$\rmidx{2}$};
				\node[state] (u_3) [below = of u_2]  {$\rmidx{3}$};
				\node[state,accepting] (u_acc) [below = of u_3]  {$\rmacc$};
				
         	   \path[-Latex] (u_0) edge node[in place] {\small$\langle\text{\icnexthor{\iccolor{\icsquare}{icred}}{\iccolor{\icdooropen}{icgreen}}},0\rangle$} (u_1);
				\path[-Latex] (u_1) edge node[in place] {\small$\langle\text{\icnexthor{\iccolor{\iccircle}{icred}}{\iccolor{\ickey}{icyellow}}},0\rangle$} (u_2);
				\path[-Latex] (u_2) edge node[in place] {\small$\langle\text{\icnexthor{\iccolor{\ickey}{icpurple}}{\iccolor{\icdoor}{icblue}}},0\rangle$} (u_3);
				\path[-Latex] (u_3) edge node[in place] {\small$\langle\text{\icnexthor{\icnocolor{\ickey}}{\iccolor{\icdooropen}{icpurple}}},1\rangle$} (u_acc);
			\end{tikzpicture}
		}
		
		\caption{$\num{2e9}$}
	\end{subfigure}
	\begin{subfigure}[b]{.245\textwidth}
		\centering
		\includegraphics[height=7.5em]{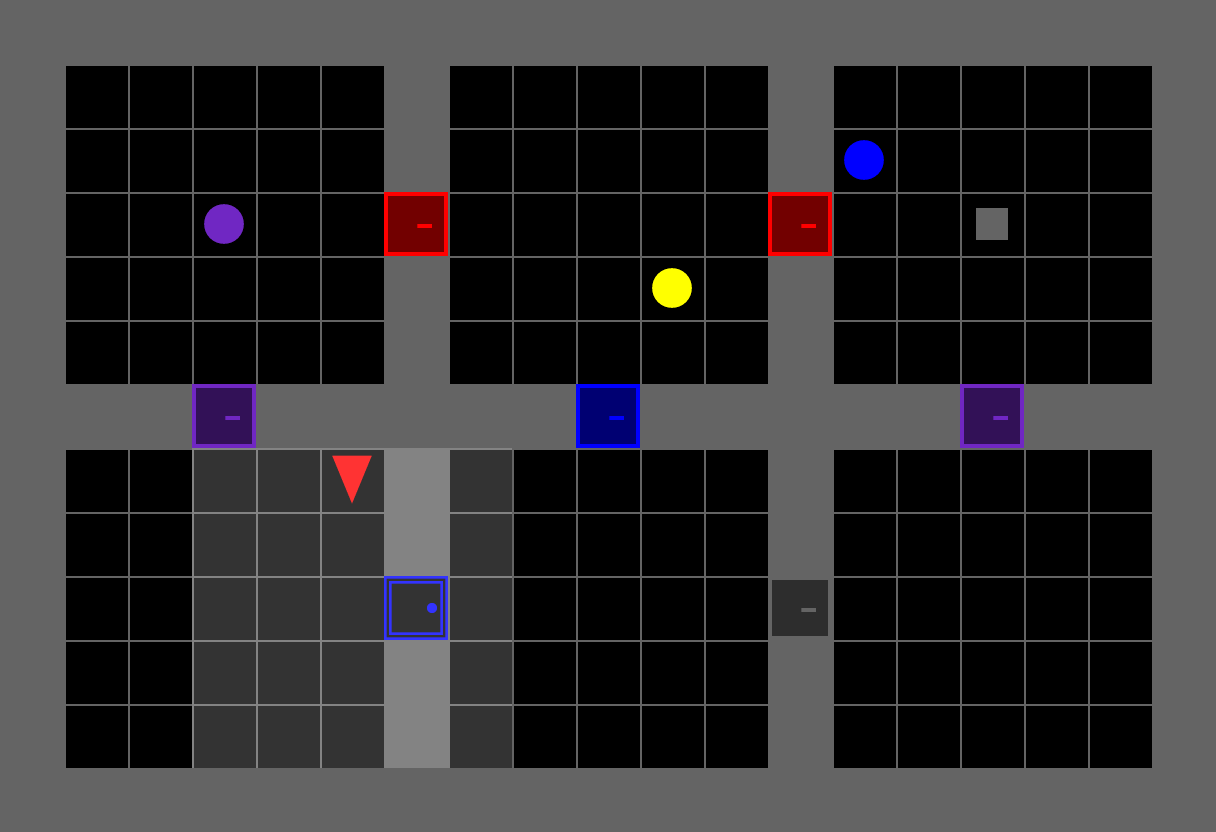}\\[1em]
		\scalebox{0.7}{
			\begin{tikzpicture}[shorten >=1pt,node distance=1.9cm,on grid,auto,every initial by arrow/.style ={-Latex}]
				\node[state,initial,initial text=] (u_0)   {$\rminit$};
				\node[state] (u_1) [below = of u_0]  {$\rmidx{1}$};
				\node[state] (u_2) [below = of u_1]  {$\rmidx{2}$};
				\node[state,accepting] (u_acc) [below = of u_2]  {$\rmacc$};
				
            	\path[-Latex] (u_0) edge node[in place] {\small$\langle\text{\icnexthor{\iccolor{\iccircle}{icblue}}{\iccolor{\icsquare}{icgreen}}},0\rangle$} (u_1);
				\path[-Latex] (u_1) edge node[in place] {\small$\langle\text{\icnexthor{\iccolor{\icsquare}{icblue}}{\iccolor{\ickey}{icgreen}}},0\rangle$} (u_2);
				\path[-Latex] (u_2) edge node[in place] {\small$\langle\text{\icnexthor{\iccolor{\iccircle}{icgray}}{\iccolor{\icdoor}{icyellow}}},1\rangle$} (u_acc);
			\end{tikzpicture}
		}
		\caption{$\num{3e9}$}
	\end{subfigure}
	\begin{subfigure}[b]{.245\textwidth}
		\centering
		\includegraphics[height=4em]{figures/sampling/seq-dr-i/dr-seq-plr-i-2000-7-lvl.png}\\[1em]
		\scalebox{0.7}{
			\begin{tikzpicture}[shorten >=1pt,node distance=1.9cm,on grid,auto,every initial by arrow/.style ={-Latex}]
				\node[state,initial,initial text=] (u_0)   {$\rminit$};
				\node[state] (u_1) [below = of u_0]  {$\rmidx{1}$};
				\node[state,accepting] (u_acc) [below = of u_1]  {$\rmacc$};
				
                \path[-Latex] (u_0) edge node[in place] {\small$\langle\text{\icnexthor{\iccolor{\icsquare}{icred}}{\iccolor{\icdoor}{icgray}}},0\rangle$} (u_1);
				\path[-Latex] (u_1) edge node[in place] {\small$\langle\text{\icnexthor{\iccolor{\icsquare}{icblue}}{\iccolor{\icdoor}{icgray}}},1\rangle$} (u_acc);
			\end{tikzpicture}
		}
		
		\caption{$\num{4e9}$}
	\end{subfigure}
	\caption{Samples of generated problems at different points in time (in number of environment steps) using DR.}
	\label{fig:gen_prob_dr}
\end{figure}

\begin{figure}
	\centering
	\begin{subfigure}[b]{.245\textwidth}
		\centering
		\includegraphics[height=4em]{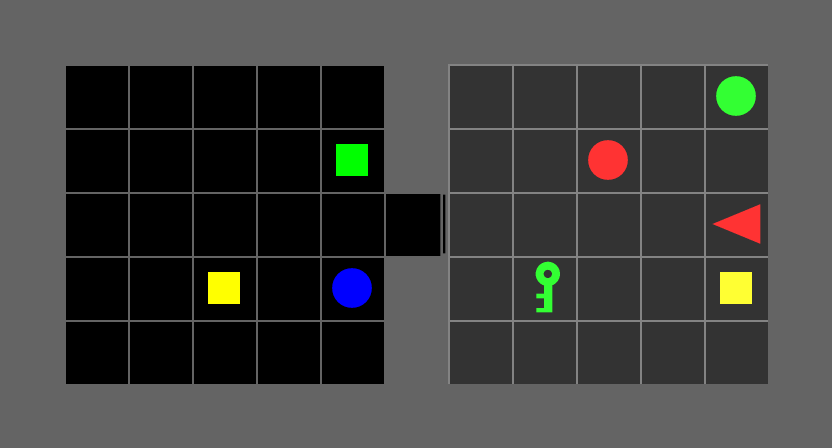}\\[1em]
		\scalebox{0.7}{
			\begin{tikzpicture}[shorten >=1pt,node distance=1.9cm,on grid,auto,every initial by arrow/.style ={-Latex}]
				\node[state,initial,initial text=] (u_0)   {$\rminit$};
				\node[state] (u_1) [below = of u_0]  {$\rmidx{1}$};
				\node[state,accepting] (u_acc) [below = of u_1]  {$\rmacc$};
				
                \path[-Latex] (u_0) edge node[in place] {\small$\langle\text{\icnexthor{\iccolor{\iccircle}{icblue}}{\iccolor{\icsquare}{icgreen}}},0\rangle$} (u_1);
				\path[-Latex] (u_1) edge node[in place] {\small$\langle\text{\icnexthor{\iccolor{\ickey}{icgreen}}{\icnocolor{\icdoorclosed}}},1\rangle$} (u_acc);
			\end{tikzpicture}
		}
		\caption{$\num{1e9}$}
	\end{subfigure}
	\begin{subfigure}[b]{.245\textwidth}
		\centering
		\includegraphics[height=7.5em]{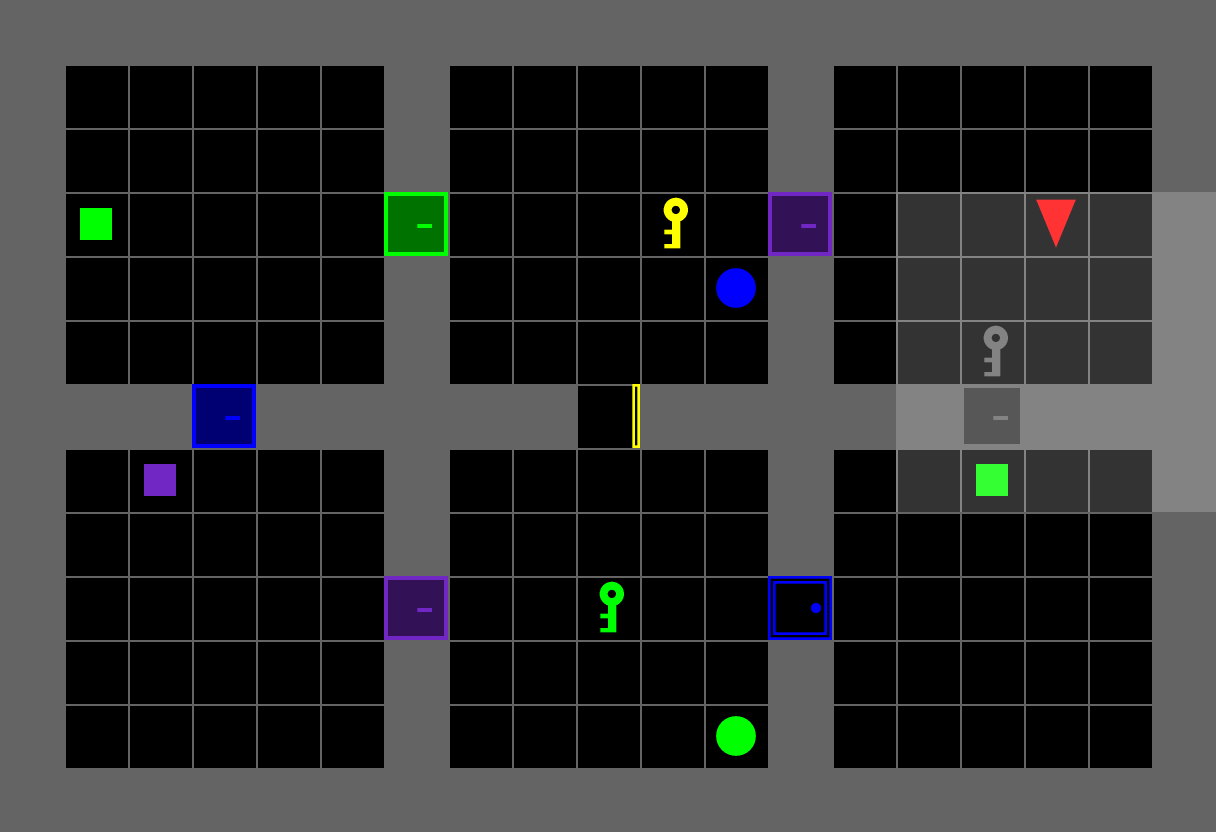}\\[1em]
		\scalebox{0.7}{
			\begin{tikzpicture}[shorten >=1pt,node distance=1.9cm,on grid,auto,every initial by arrow/.style ={-Latex}]
				\node[state,initial,initial text=] (u_0)   {$\rminit$};
				\node[state] (u_1) [below = of u_0]  {$\rmidx{1}$};
				\node[state,accepting] (u_acc) [below = of u_1]  {$\rmacc$};
				
            	\path[-Latex] (u_0) edge node[in place] {\small$\langle\text{\icnexthor{\iccolor{\iccircle}{icblue}}{\icnocolor{\icdoorclosed}}},0\rangle$} (u_1);
				\path[-Latex] (u_1) edge node[in place] {\small$\langle\text{\icnexthor{\iccolor{\iccircle}{icgreen}}{\iccolor{\icdoorlocked}{icgreen}}},1\rangle$} (u_acc);
			\end{tikzpicture}
		}
		\caption{$\num{2e9}$}
	\end{subfigure}
	\begin{subfigure}[b]{.245\textwidth}
		\centering
		\includegraphics[height=7.5em]{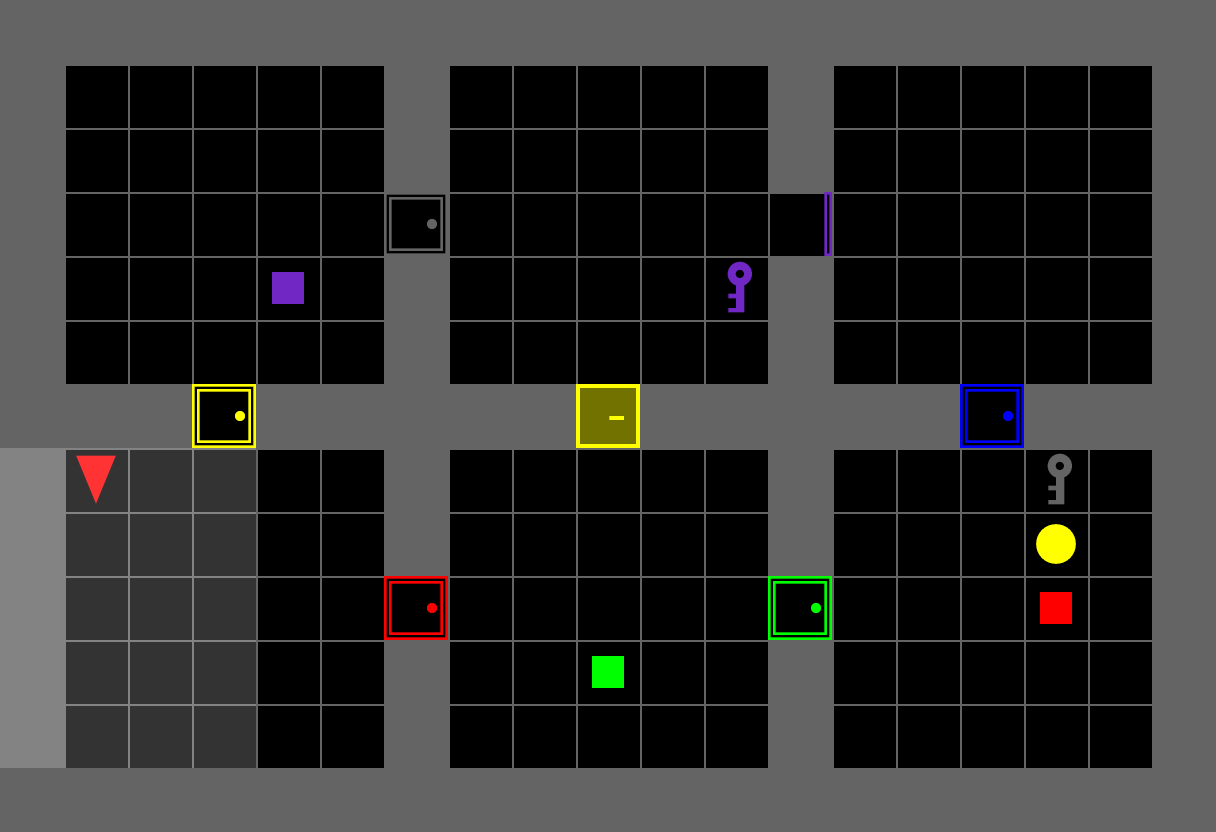}\\[1em]
		\scalebox{0.7}{
			\begin{tikzpicture}[shorten >=1pt,node distance=1.9cm,on grid,auto,every initial by arrow/.style ={-Latex}]
				\node[state,initial,initial text=] (u_0)   {$\rminit$};
				\node[state] (u_1) [below = of u_0]  {$\rmidx{1}$};
				\node[state] (u_2) [below = of u_1]  {$\rmidx{2}$};
				\node[state] (u_3) [below = of u_2]  {$\rmidx{3}$};
				\node[state,accepting] (u_acc) [below = of u_3]  {$\rmacc$};
				
        	    \path[-Latex] (u_0) edge node[in place] {\small$\langle\text{\icnexthor{\icnocolor{\iccircle}}{\iccolor{\icdoor}{icred}}},0\rangle$} (u_1);
				\path[-Latex] (u_1) edge node[in place] {\small$\langle\text{\icnexthor{\iccolor{\ickey}{icpurple}}{\icnocolor{\icdoorclosed}}},0\rangle$} (u_2);
				\path[-Latex] (u_2) edge node[in place] {\small$\langle\text{\icnexthor{\icnocolor{\iccircle}}{\iccolor{\icsquare}{icgreen}}},0\rangle$} (u_3);
				\path[-Latex] (u_3) edge node[in place] {\small$\langle\text{\icnexthor{\iccolor{\icsquare}{icpurple}}{\iccolor{\icdooropen}{icgray}}},1\rangle$} (u_acc);
			\end{tikzpicture}
		}
		\caption{$\num{3e9}$}
	\end{subfigure}
	\begin{subfigure}[b]{.245\textwidth}
		\centering
		\includegraphics[height=7.5em]{figures/sampling/seq-plr-i/seq-plr-i-2000-8-lvl.png}\\[1em]
		\scalebox{0.7}{
			\begin{tikzpicture}[shorten >=1pt,node distance=1.9cm,on grid,auto,every initial by arrow/.style ={-Latex}]
				\node[state,initial,initial text=] (u_0)   {$\rminit$};
				\node[state] (u_1) [below = of u_0]  {$\rmidx{1}$};
				\node[state] (u_2) [below = of u_1]  {$\rmidx{2}$};
				\node[state,accepting] (u_acc) [below = of u_2]  {$\rmacc$};
				
                \path[-Latex] (u_0) edge node[in place] {\small$\langle\text{\icnexthor{\iccolor{\ickey}{icgreen}}{\iccolor{\icdoorclosed}{icred}}},0\rangle$} (u_1);
				\path[-Latex] (u_1) edge node[in place] {\small$\langle\text{\icnexthor{\iccolor{\iccircle}{icyellow}}{\icnocolor{\iccircle}}},0\rangle$} (u_2);
				\path[-Latex] (u_2) edge node[in place] {\small$\langle\text{\icnexthor{\iccolor{\iccircle}{icred}}{\iccolor{\icdooropen}{icred}}},1\rangle$} (u_acc);
			\end{tikzpicture}
		}
		\caption{$\num{4e9}$}
	\end{subfigure}
	\caption{Samples of generated problems at different points in time (in number of environment steps) using \rplr.}
	\label{fig:gen_prob_plr}
\end{figure}

\begin{figure}
	\centering
	\begin{subfigure}[b]{.245\textwidth}
		\centering
		\includegraphics[height=7.5em]{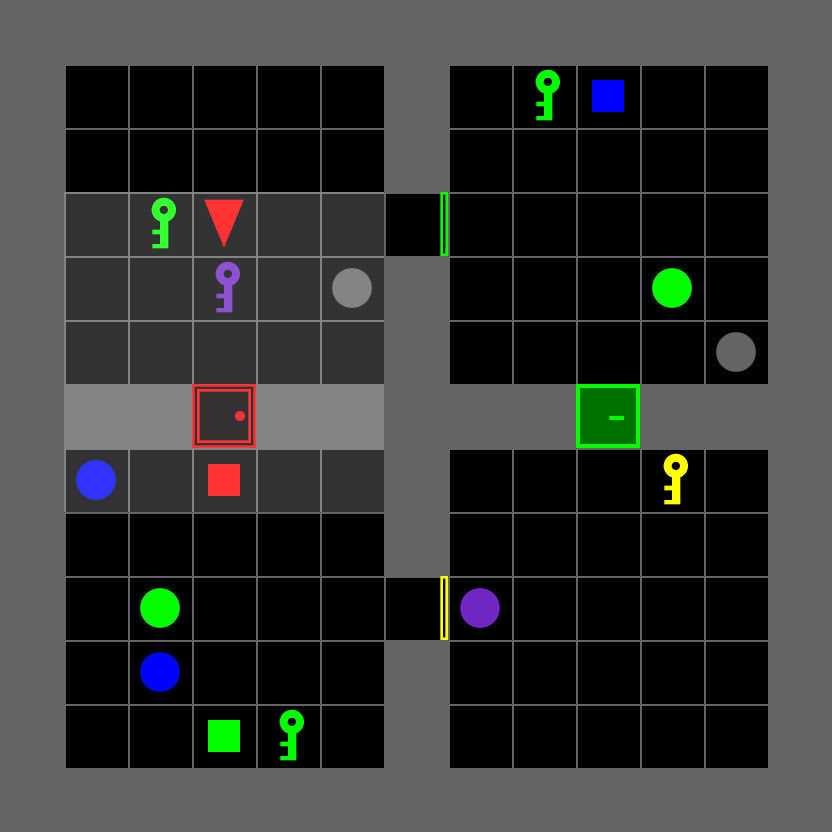}\\[1em]
		\scalebox{0.7}{
			\begin{tikzpicture}[shorten >=1pt,node distance=1.9cm,on grid,auto,every initial by arrow/.style ={-Latex}]
				\node[state,initial,initial text=] (u_0)   {$\rminit$};
				\node[state] (u_1) [below = of u_0]  {$\rmidx{1}$};
				\node[state] (u_2) [below = of u_1]  {$\rmidx{2}$};
				\node[state] (u_3) [below = of u_2]  {$\rmidx{3}$};
				\node[state] (u_4) [below = of u_3]  {$\rmidx{4}$};
				\node[state,accepting] (u_acc) [below = of u_4]  {$\rmacc$};
				
  	            \path[-Latex] (u_0) edge node[in place] {\small$\langle\text{\iccolor{\icsquare}{icblue}},0\rangle$} (u_1);
				\path[-Latex] (u_1) edge node[in place] {\small$\langle\text{\icnexthor{\iccolor{\ickey}{icyellow}}{\iccolor{\icdooropen}{icred}}},0\rangle$} (u_2);
				\path[-Latex] (u_2) edge node[in place] {\small$\langle\text{\icnexthor{\iccolor{\ickey}{icgreen}}{\iccolor{\icdooropen}{icgreen}}},0\rangle$} (u_3);
				\path[-Latex] (u_3) edge node[in place] {\small$\langle\text{\icnexthor{\iccolor{\ickey}{icpurple}}{\icnocolor{\icdoorclosed}}},0\rangle$} (u_4);
				\path[-Latex] (u_4) edge node[in place] {\small$\langle\text{\icnexthor{\iccolor{\ickey}{icgreen}}{\icnocolor{\icdooropen}}},1\rangle$} (u_acc);
			\end{tikzpicture}
		}
		\caption{$\num{1e9}$}
	\end{subfigure}
	\begin{subfigure}[b]{.245\textwidth}
		\centering
		\includegraphics[height=7.5em]{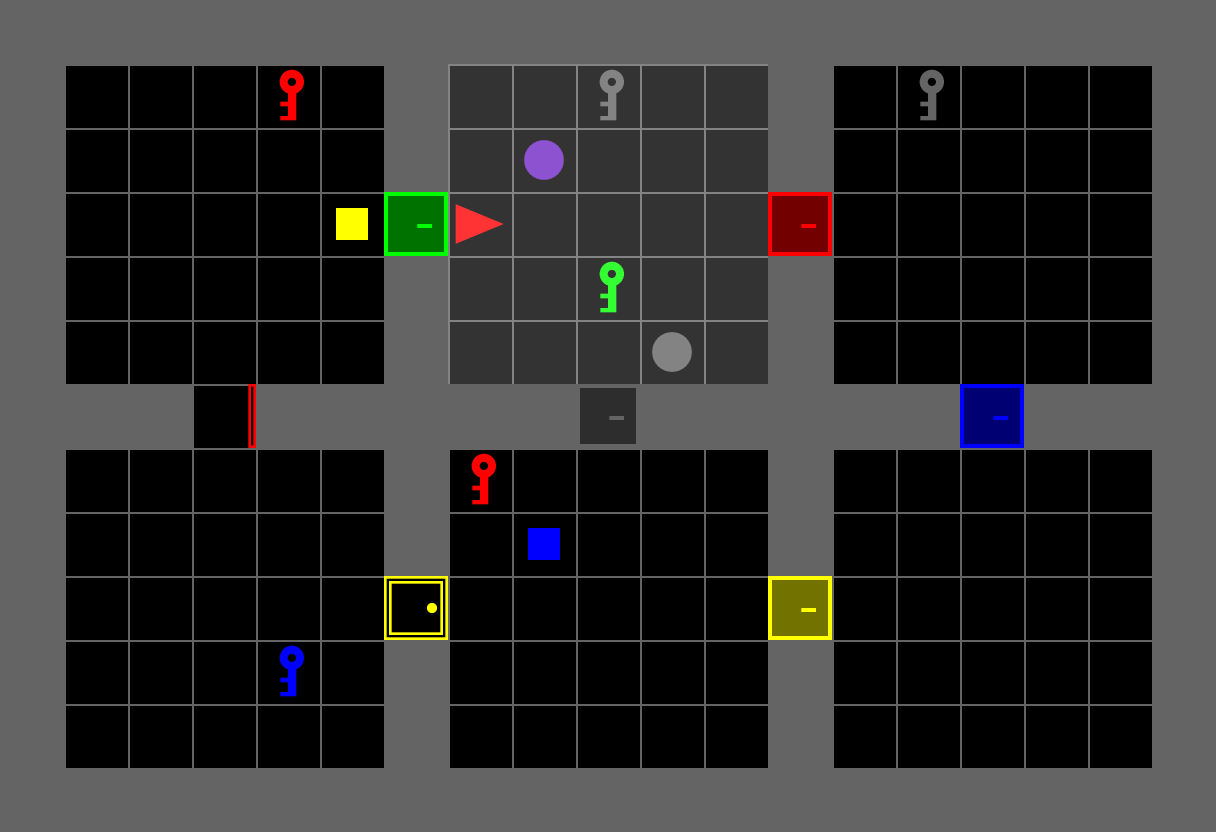}\\[1em]
		\scalebox{0.7}{
			\begin{tikzpicture}[shorten >=1pt,node distance=1.9cm,on grid,auto,every initial by arrow/.style ={-Latex}]
				\node[state,initial,initial text=] (u_0)   {$\rminit$};
				\node[state] (u_1) [below = of u_0]  {$\rmidx{1}$};
				\node[state] (u_2) [below = of u_1]  {$\rmidx{2}$};
				\node[state,accepting] (u_acc) [below = of u_2]  {$\rmacc$};
				
       		     \path[-Latex] (u_0) edge node[in place] {\small$\langle\text{\icnexthor{\iccolor{\iccircle}{icgray}}{\iccolor{\icdooropen}{icgray}}},0\rangle$} (u_1);
				\path[-Latex] (u_1) edge node[in place] {\small$\langle\text{\icnexthor{\iccolor{\ickey}{icred}}{\iccolor{\icdooropen}{icred}}},0\rangle$} (u_2);
				\path[-Latex] (u_2) edge node[in place] {\small$\langle\text{\icnexthor{\iccolor{\iccircle}{icgray}}{\iccolor{\ickey}{icblue}}},1\rangle$} (u_acc);
			\end{tikzpicture}
		}
		\caption{$\num{2e9}$}
	\end{subfigure}
	\begin{subfigure}[b]{.245\textwidth}
		\centering
		\includegraphics[height=7.5em]{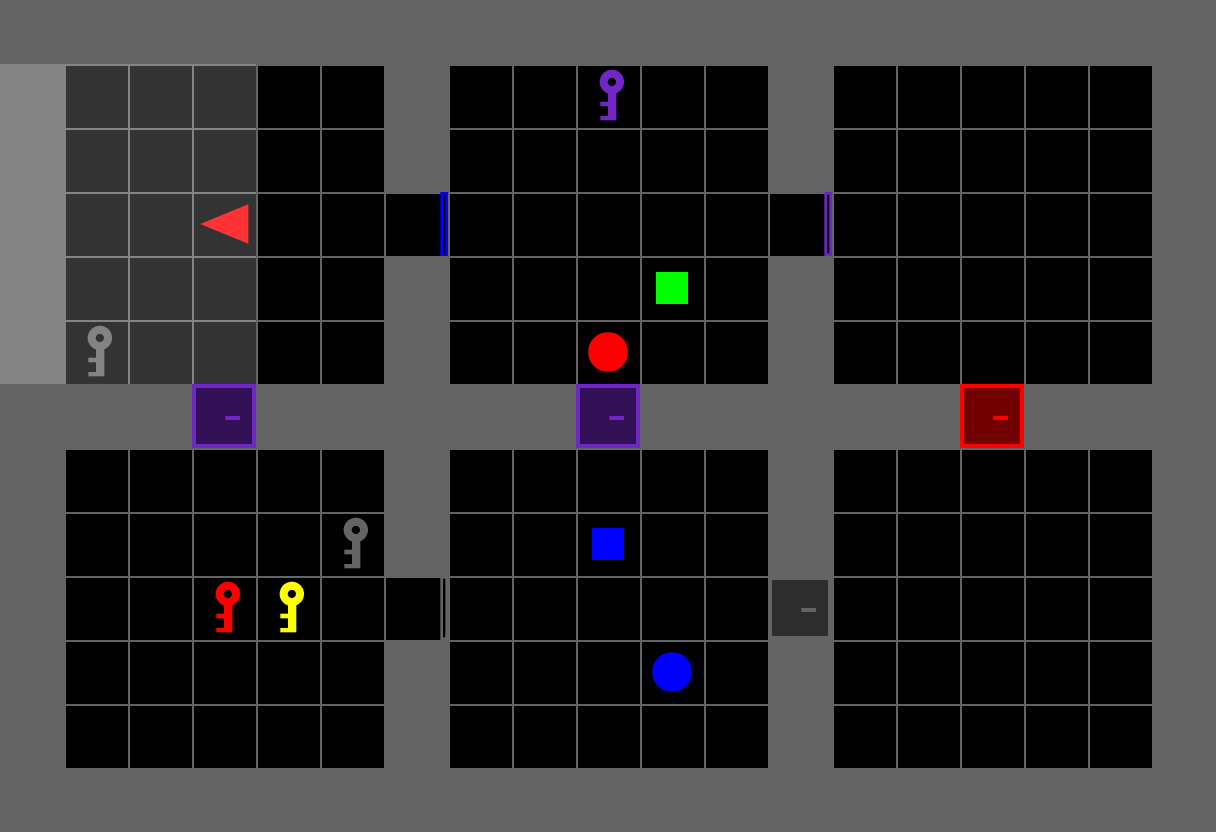}\\[1em]
		\scalebox{0.7}{
			\begin{tikzpicture}[shorten >=1pt,node distance=1.9cm,on grid,auto,every initial by arrow/.style ={-Latex}]
				\node[state,initial,initial text=] (u_0)   {$\rminit$};
				\node[state] (u_1) [below = of u_0]  {$\rmidx{1}$};
				\node[state] (u_2) [below = of u_1]  {$\rmidx{2}$};
				\node[state] (u_3) [below = of u_2]  {$\rmidx{3}$};
				\node[state] (u_4) [below = of u_3]  {$\rmidx{4}$};
				\node[state,accepting] (u_acc) [below = of u_4]  {$\rmacc$};
				
				\path[-Latex] (u_0) edge node[in place] {\small$\langle\text{\icnexthor{\icnocolor{\icsquare}}{\iccolor{\ickey}{icyellow}}},0\rangle$} (u_1);
				\path[-Latex] (u_1) edge node[in place] {\small$\langle\text{\icnexthor{\iccolor{\ickey}{icpurple}}{\icnocolor{\icdoorclosed}}},0\rangle$} (u_2);
				\path[-Latex] (u_2) edge node[in place] {\small$\langle\text{\icnexthor{\iccolor{\iccircle}{icblue}}{\iccolor{\icsquare}{icgreen}}},0\rangle$} (u_3);
				\path[-Latex] (u_3) edge node[in place] {\small$\langle\text{\icnexthor{\iccolor{\iccircle}{icblue}}{\iccolor{\ickey}{icred}}},0\rangle$} (u_4);
				\path[-Latex] (u_4) edge node[in place] {\small$\langle\text{\icnexthor{\icnocolor{\ickey}}{\icnocolor{\ickey}}},1\rangle$} (u_acc);
			\end{tikzpicture}
		}
		\caption{$\num{3e9}$}
	\end{subfigure}
	\begin{subfigure}[b]{.245\textwidth}
		\centering
		\includegraphics[height=7.5em]{figures/sampling/seq-accel_full-i/seq-accel_full-i-2000-9-lvl.png}\\[1em]
		\scalebox{0.7}{
			\begin{tikzpicture}[shorten >=1pt,node distance=1.9cm,on grid,auto,every initial by arrow/.style ={-Latex}]
				\node[state,initial,initial text=] (u_0)   {$\rminit$};
				\node[state] (u_1) [below = of u_0]  {$\rmidx{1}$};
				\node[state] (u_2) [below = of u_1]  {$\rmidx{2}$};
				\node[state] (u_3) [below = of u_2]  {$\rmidx{3}$};
				\node[state,accepting] (u_acc) [below = of u_3]  {$\rmacc$};
				
  	            \path[-Latex] (u_0) edge node[in place] {\small$\langle\text{\icnexthor{\icnocolor{\icsquare}}{\iccolor{\icdoorlocked}{icgray}}},0\rangle$} (u_1);
				\path[-Latex] (u_1) edge node[in place] {\small$\langle\text{\icnexthor{\iccolor{\ickey}{icgreen}}{\iccolor{\icdoor}{icpurple}}},0\rangle$} (u_2);
				\path[-Latex] (u_2) edge node[in place] {\small$\langle\text{\icnexthor{\iccolor{\icsquare}{icblue}}{\iccolor{\icdoorlocked}{icgray}}},0\rangle$} (u_3);
				\path[-Latex] (u_3) edge node[in place] {\small$\langle\text{\icnexthor{\iccolor{\icsquare}{icblue}}{\icnocolor{\ickey}}},1\rangle$} (u_acc);
			\end{tikzpicture}
		}
		\caption{$\num{4e9}$}
	\end{subfigure}
	\caption{Samples of generated problems at different points in time (in number of environment steps) using ACCEL.}
	\label{fig:gen_prob_accel_full}
\end{figure}

\begin{figure}
	\centering
	\begin{subfigure}[b]{.245\textwidth}
		\centering
		\includegraphics[height=7em]{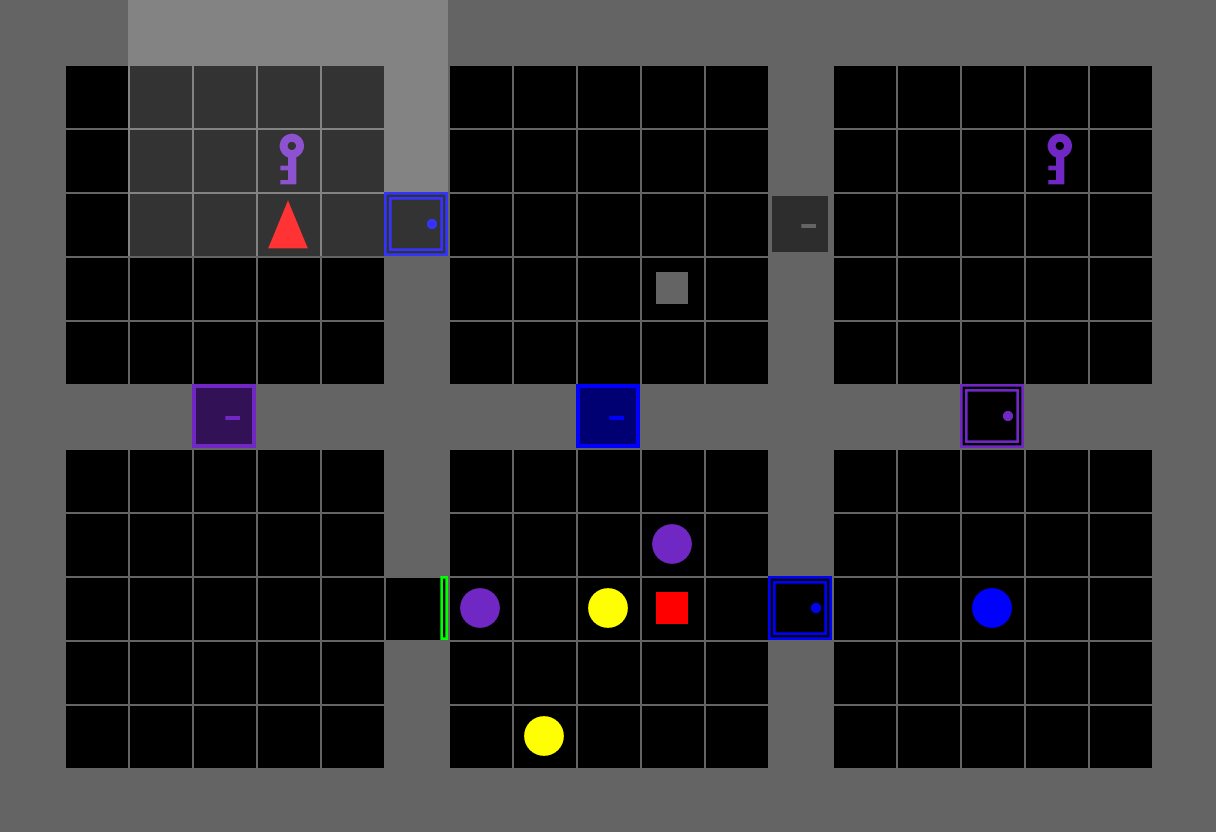}\\[1em]
		\scalebox{0.7}{
			\begin{tikzpicture}[shorten >=1pt,node distance=1.9cm,on grid,auto,every initial by arrow/.style ={-Latex}]
				\node[state,initial,initial text=] (u_0)   {$\rminit$};
				\node[state] (u_1) [below = of u_0]  {$\rmidx{1}$};
				\node[state] (u_2) [below = of u_1]  {$\rmidx{2}$};
				\node[state] (u_3) [below = of u_2]  {$\rmidx{3}$};
				\node[state,accepting] (u_acc) [below = of u_3]  {$\rmacc$};
				
            	\path[-Latex] (u_0) edge node[in place] {\small$\langle\text{\icnexthor{\icnocolor{\iccircle}}{\icnocolor{\ickey}}},0\rangle$} (u_1);
				\path[-Latex] (u_1) edge node[in place] {\small$\langle\text{\icnexthor{\iccolor{\icsquare}{icred}}{\icnocolor{\icdooropen}}},0\rangle$} (u_2);
				\path[-Latex] (u_2) edge node[in place] {\small$\langle\text{\icnexthor{\iccolor{\iccircle}{icpurple}}{\icnocolor{\iccircle}}},0\rangle$} (u_3);
				\path[-Latex] (u_3) edge node[in place] {\small$\langle\text{\icnexthor{\iccolor{\ickey}{icpurple}}{\iccolor{\icdoorclosed}{icblue}}},1\rangle$} (u_acc);
			\end{tikzpicture}
		}
		\caption{$\num{1e9}$}
	\end{subfigure}
	\begin{subfigure}[b]{.245\textwidth}
		\centering
		\includegraphics[height=7em]{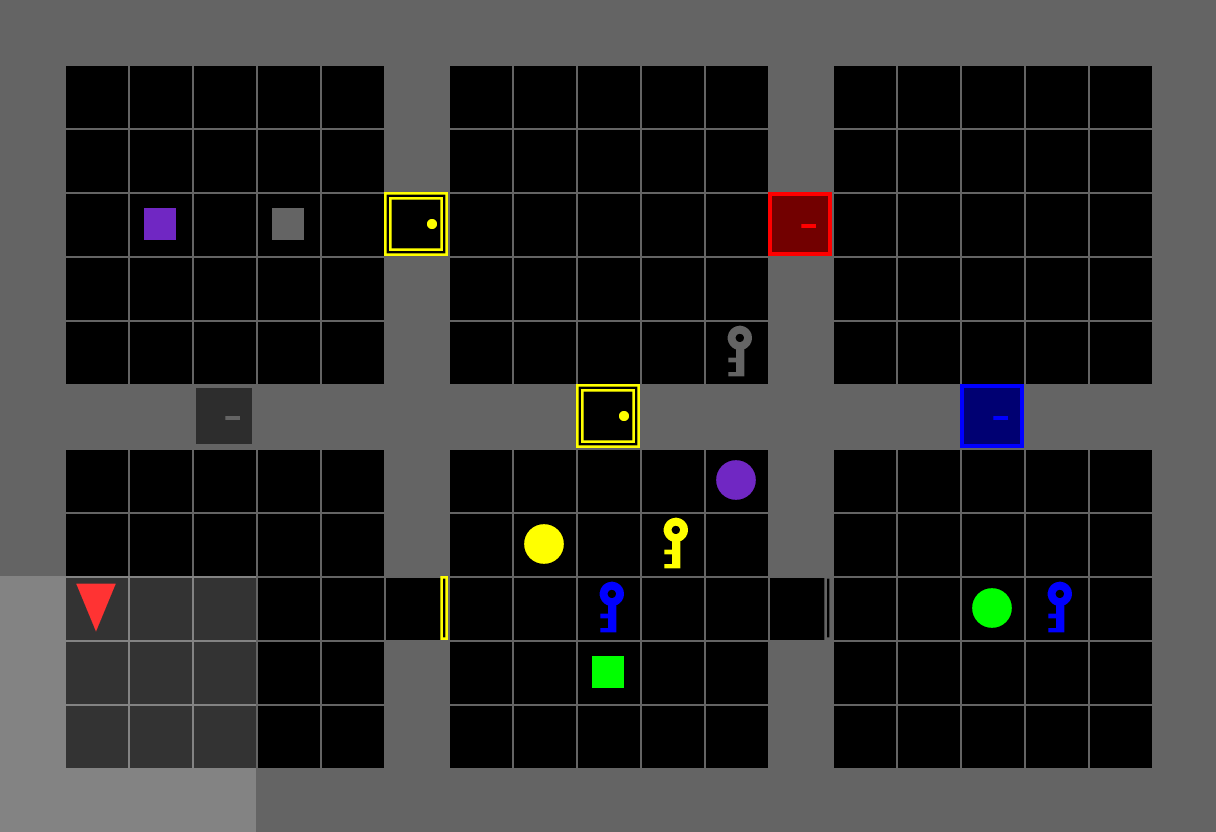}\\[1em]
		\scalebox{0.7}{
			\begin{tikzpicture}[shorten >=1pt,node distance=1.9cm,on grid,auto,every initial by arrow/.style ={-Latex}]
				\node[state,initial,initial text=] (u_0)   {$\rminit$};
				\node[state] (u_1) [below = of u_0]  {$\rmidx{1}$};
				\node[state] (u_2) [below = of u_1]  {$\rmidx{2}$};
				\node[state] (u_3) [below = of u_2]  {$\rmidx{3}$};
				\node[state] (u_4) [below = of u_3]  {$\rmidx{4}$};
				\node[state,accepting] (u_acc) [below = of u_4]  {$\rmacc$};
				
            	\path[-Latex] (u_0) edge node[in place] {\small$\langle\text{\icnexthor{\iccolor{\ickey}{icyellow}}{\iccolor{\icdoorlocked}{icblue}}},0\rangle$} (u_1);
				\path[-Latex] (u_1) edge node[in place] {\small$\langle\text{\icnexthor{\iccolor{\iccircle}{icyellow}}{\iccolor{\ickey}{icgray}}},0\rangle$} (u_2);
				\path[-Latex] (u_2) edge node[in place] {\small$\langle\text{\icnexthor{\iccolor{\icsquare}{icgray}}{\iccolor{\icdooropen}{icgray}}},0\rangle$} (u_3);
				\path[-Latex] (u_3) edge node[in place] {\small$\langle\text{\icnexthor{\icnocolor{\iccircle}}{\iccolor{\ickey}{icblue}}},0\rangle$} (u_4);
				\path[-Latex] (u_4) edge node[in place, pos=0.4] {\small$\langle\text{\iccarrying{\iccolor{\icsquare}{icgreen}}},1\rangle$} (u_acc);
			\end{tikzpicture}
		}
		\caption{$\num{2e9}$}
	\end{subfigure}
	\begin{subfigure}[b]{.245\textwidth}
		\centering
		\includegraphics[height=7em]{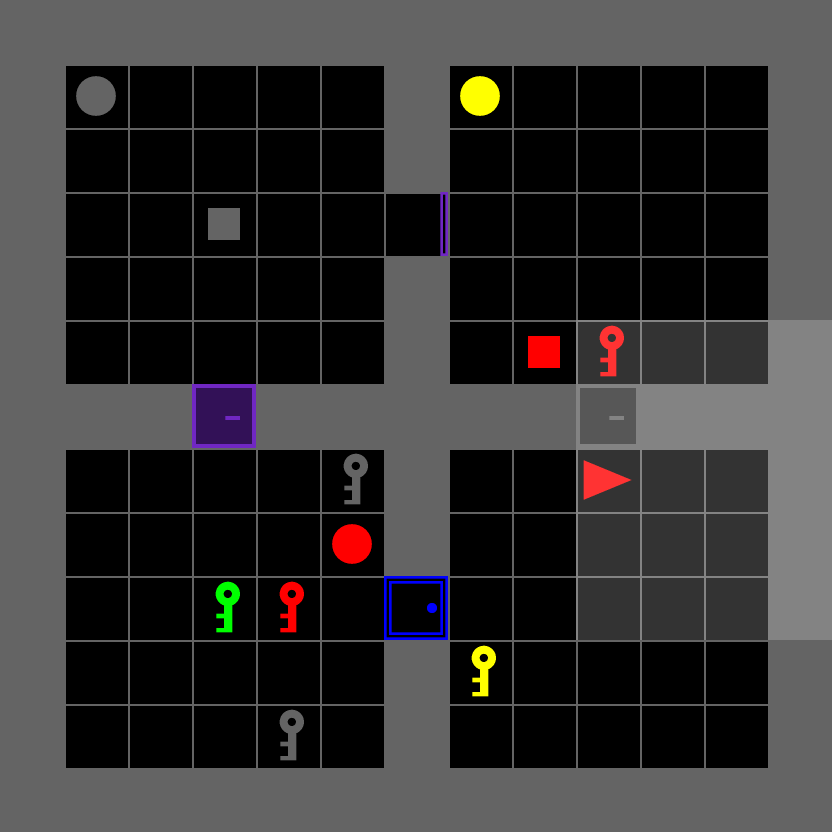}\\[1em]
		\scalebox{0.7}{
			\begin{tikzpicture}[shorten >=1pt,node distance=1.9cm,on grid,auto,every initial by arrow/.style ={-Latex}]
				\node[state,initial,initial text=] (u_0)   {$\rminit$};
				\node[state] (u_1) [below = of u_0]  {$\rmidx{1}$};
				\node[state] (u_2) [below = of u_1]  {$\rmidx{2}$};
				\node[state] (u_3) [below = of u_2]  {$\rmidx{3}$};
				\node[state] (u_4) [below = of u_3]  {$\rmidx{4}$};
				\node[state,accepting] (u_acc) [below = of u_4]  {$\rmacc$};
				
            	\path[-Latex] (u_0) edge node[in place] {\small$\langle\text{\icnexthor{\iccolor{\iccircle}{gray}}{\iccolor{\icsquare}{icred}}},0\rangle$} (u_1);
				\path[-Latex] (u_1) edge node[in place] {\small$\langle\text{\icnexthor{\iccolor{\ickey}{gray}}{\icnocolor{\icdoorclosed}}},0\rangle$} (u_2);
				\path[-Latex] (u_2) edge node[in place] {\small$\langle\text{\icnexthor{\icnocolor{\icsquare}}{\iccolor{\ickey}{icgray}}},0\rangle$} (u_3);
				\path[-Latex] (u_3) edge node[in place] {\small$\langle\text{\icnexthor{\iccolor{\icsquare}{icred}}{\icnocolor{\icdoorclosed}}},0\rangle$} (u_4);
				\path[-Latex] (u_4) edge node[in place] {\small$\langle\text{\icnexthor{\iccolor{\iccircle}{icgray}}{\icnocolor{\icdooropen}}},1\rangle$} (u_acc);
			\end{tikzpicture}
		}
		\caption{$\num{3e9}$}
	\end{subfigure}
	\begin{subfigure}[b]{.245\textwidth}
		\centering
		\includegraphics[height=7em]{figures/sampling/seq-accel_scratch-i/seq-accel_scratch-i-2000-0-lvl.png}\\[1em]
		\scalebox{0.7}{
			\begin{tikzpicture}[shorten >=1pt,node distance=1.9cm,on grid,auto,every initial by arrow/.style ={-Latex}]
				\node[state,initial,initial text=] (u_0)   {$\rminit$};
				\node[state] (u_1) [below = of u_0]  {$\rmidx{1}$};
				\node[state] (u_2) [below = of u_1]  {$\rmidx{2}$};
				\node[state] (u_3) [below = of u_2]  {$\rmidx{3}$};
				\node[state] (u_4) [below = of u_3]  {$\rmidx{4}$};
				\node[state,accepting] (u_acc) [below = of u_4]  {$\rmacc$};
				
				 \path[-Latex] (u_0) edge node[in place] {\small$\langle\text{\icnexthor{\iccolor{\icsquare}{icyellow}}{\iccolor{\icdoorlocked}{icgray}}},0\rangle$} (u_1);
				\path[-Latex] (u_1) edge node[in place] {\small$\langle\text{\icnexthor{\iccolor{\iccircle}{icyellow}}{\icnocolor{\ickey}}},0\rangle$} (u_2);
				\path[-Latex] (u_2) edge node[in place] {\small$\langle\text{\iccolor{\icdooropen}{icgreen}},0\rangle$} (u_3);
				\path[-Latex] (u_3) edge node[in place] {\small$\langle\text{\icnexthor{\iccolor{\ickey}{icgreen}}{\iccolor{\icdooropen}{icgray}}},0\rangle$} (u_4);
				\path[-Latex] (u_4) edge node[in place] {\small$\langle\text{\icnexthor{\iccolor{\iccircle}{icgreen}}{\iccolor{\icdoorclosed}{icgreen}}},1\rangle$} (u_acc);
			\end{tikzpicture}
		}
		\caption{$\num{4e9}$}
	\end{subfigure}
	\caption{Samples of generated problems at different points in time (in number of environment steps) using ACCEL-0.}
	\label{fig:gen_prob_accel_scratch}
\end{figure}

\subsubsection{Mutation Analysis}
\cref{fig:mutations_occurrence} illustrates the evolution of mutations in buffer problems throughout training. Both ACCEL and ACCEL-0 exhibit similar trends. Hindsight edits are barely present (possibly) because they are hard to sample: they are only applicable if the RM state is not $\rminit$ or $\rmacc$, and can only be randomly selected as the first edit in the sequence. Non-hindsight edits have a similar and continued presence in the buffer.

\cref{fig:mutation_lengths} shows the average number of edits that produced each buffer problem from its parent. Non-mutated problems (i.e.,~problems with zero edits) are accounted for. In our experiments, the number of edits is sampled uniformly at random between 7 and 10. For both ACCEL variants, the average number of edits quickly grows close to the maximum, showing that (i)~longer edit sequences are beneficial, and (ii)~the buffer eventually consists mostly of mutated problems.

\begin{figure*}
	\centering
	\begin{subfigure}[b]{0.69\textwidth}
		\includegraphics[width=\textwidth]{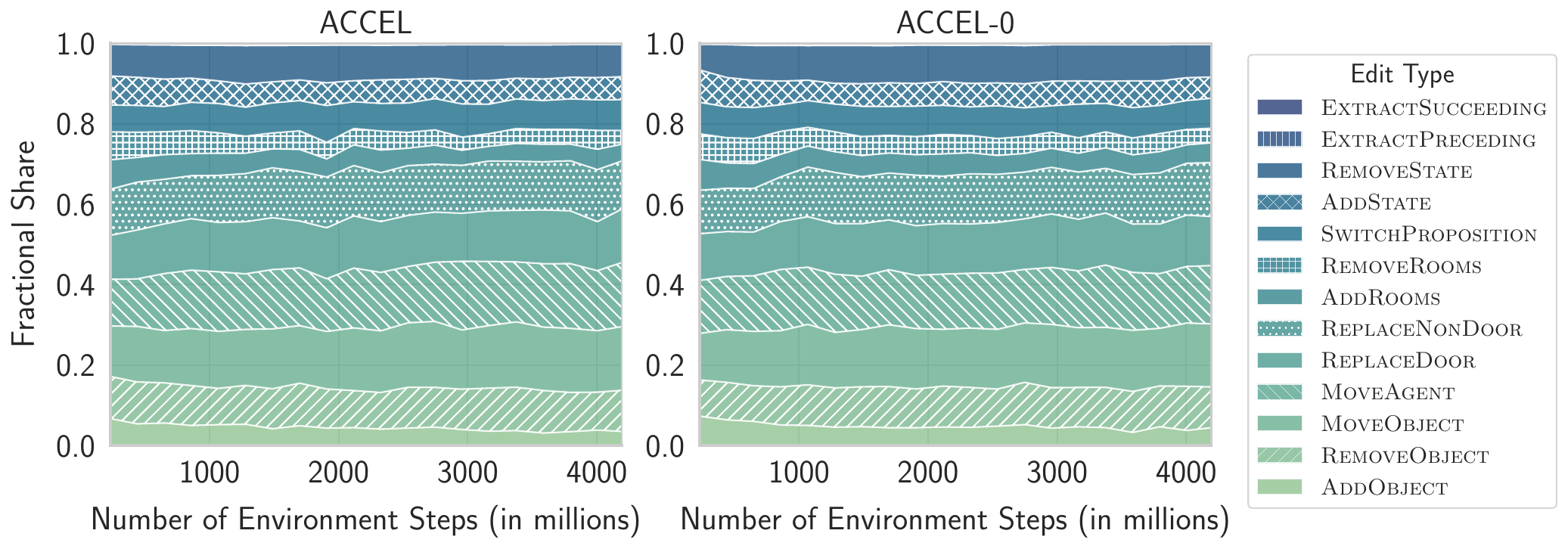}
		\caption{Edit occurrence probability (ignoring non-mutated problems).}
		\label{fig:mutations_occurrence}
	\end{subfigure}
	\hfill
	\begin{subfigure}[b]{0.29\textwidth}
		\centering
		\includegraphics[width=\textwidth]{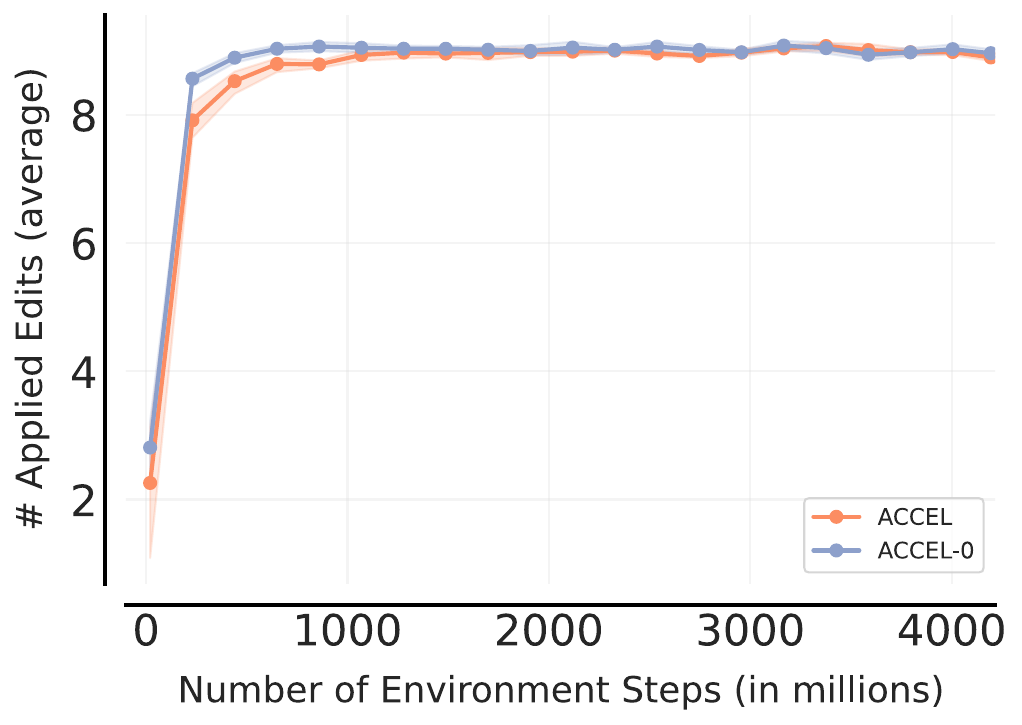}
		\caption{Average number of edits.}
		\label{fig:mutation_lengths}
	\end{subfigure}
	\caption{Evolution of edits in buffer problems throughout training. The frequency is weighted by the sampling probability of the associated problem.}
	\label{fig:mutations}
\end{figure*}

\subsection{Extended Problem Sampling Ablation Results}
\label{app:experiments_problem_sampling_ablation_results}
\cref{fig:ued_cvar_and_test_set_lvl_cond_cvar,fig:ued_cvar_and_test_set_lvl_cond_test} show the performance of DR, \rplr and ACCEL in the \emph{level-conditioned} setting, i.e.~when RM task sampling is conditioned on the level (see \cref{sec:evaluation_suite}). By sampling RMs conditionally, the fraction of solvable problems increases from 2.7\% to 83.4\%. DR is the approach that benefits most from the higher number of solvable problems per batch; indeed, it is competitive with \rplr and ACCEL in terms of robustness (CVaR) and generalization on the hand-designed test set. \cref{fig:ued_cvar_and_test_set_lvl_cond_comparison} illustrates the test performance for independent and level-conditioned sampling at the end of training. Except for DR, all approaches perform similarly for both sampling methods, further emphasizing the robustness of \rplr and ACCEL in settings where solvable problems are difficult to sample.

\begin{figure}
	\centering
	\begin{subfigure}[b]{0.33\linewidth}
		\includegraphics[width=\linewidth]{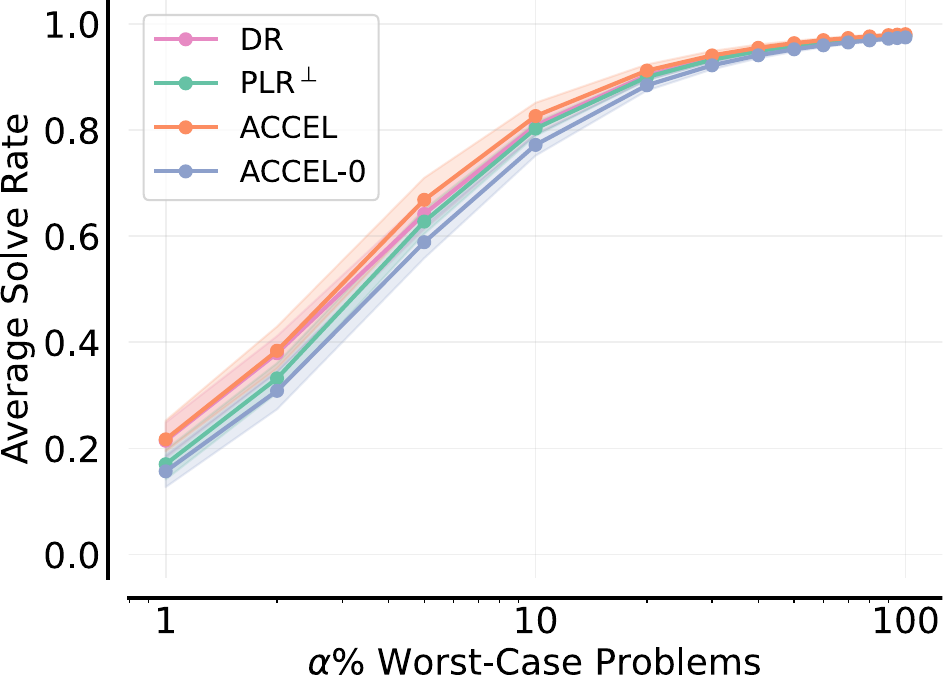}
		\caption{CVaR of the solve rate.}
		\label{fig:ued_cvar_and_test_set_lvl_cond_cvar}
	\end{subfigure}
	\begin{subfigure}[b]{0.33\linewidth}
		\includegraphics[width=\linewidth]{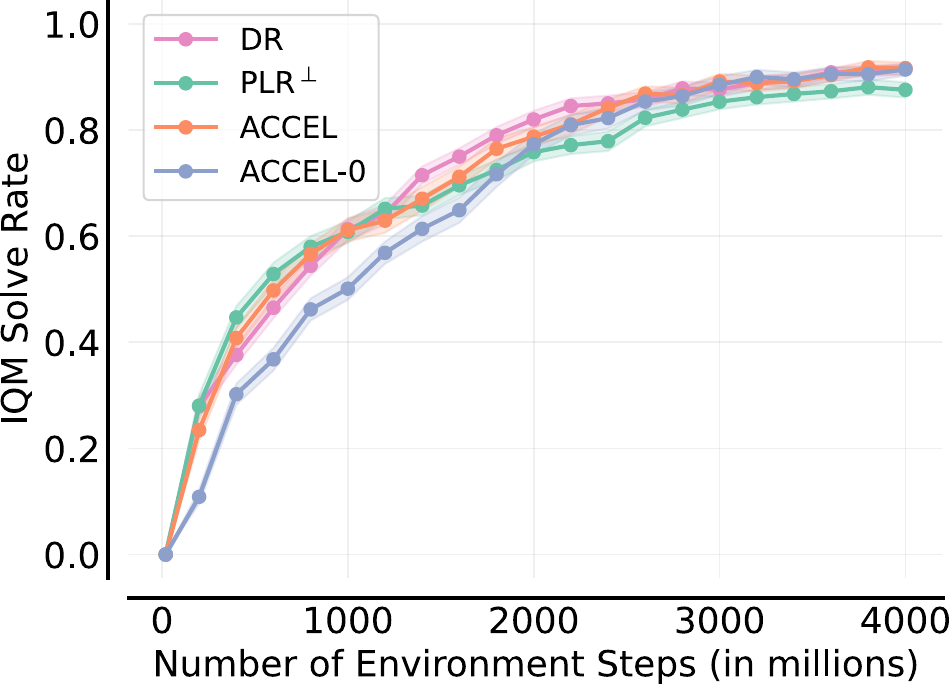}
		\caption{Performance on the hand-designed test set.}
		\label{fig:ued_cvar_and_test_set_lvl_cond_test}
	\end{subfigure}
	\begin{subfigure}[b]{0.32\linewidth}
		\centering
		\includegraphics[width=0.78\linewidth]{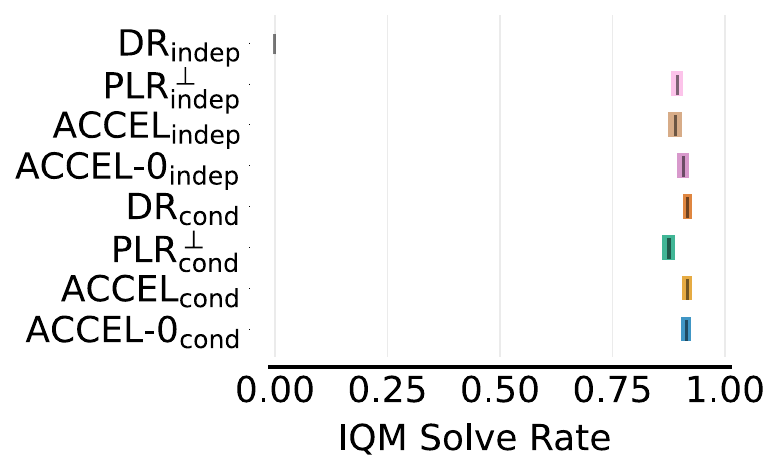}
		\caption{Comparison with independent sampling.}
		\label{fig:ued_cvar_and_test_set_lvl_cond_comparison}
	\end{subfigure}
	
	\caption{Performance of UED approaches with \emph{level-conditioned} problem sampling.}
	\label{fig:ued_cvar_and_test_set_lvl_cond}
\end{figure}

\cref{fig:seq_sampling_curriculum_cond} illustrates the evolution of problems in the buffer throughout training. As for the independent sampling case (see \cref{fig:seq_sampling_curriculum}), we observe that \rplr, ACCEL, and ACCEL-0 start prioritizing simple problems (few states, rooms, and objects) and progressively switch towards harder problems. The curriculum over the number of rooms and objects is similar for both independent and level-conditioned sampling. However, the curriculum over the number of states differs: converging toward sampling RMs with increasingly more states is more challenging in the independent sampling case. This is especially noticeable for \rplr, which generates RMs with 3--4 states (resp. 5--6) for independent sampling (resp. level-conditioned) by the end of training. We hypothesize that enabling the sampling of a higher fraction of solvable problems due to level-conditioning induces this effect on \rplr.

\begin{figure}
	\centering
	\includegraphics[width=\linewidth]{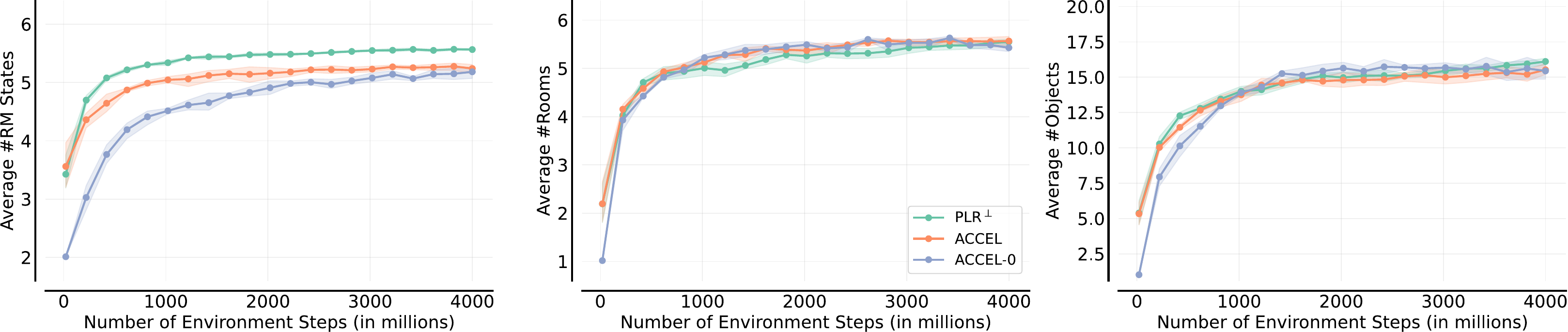}
	\caption{Emergent complexity metrics for problems in the buffer in the \emph{level-conditioned} sampling setting.}
	\label{fig:seq_sampling_curriculum_cond}
\end{figure}

\cref{fig:solvability_over_time_level_cond} shows the fraction of solvable problems in the buffer throughout training. As with independent sampling (\cref{fig:solvability_over_time}, \cref{app:experiments_main_results}), \rplr and ACCEL manage to curate a buffer mostly constituted by solvable problems. We emphasize that although the generator produces level-conditioned samples, the edits are not level-conditioned and may produce unsolvable problems.

\begin{figure}
	\centering
	\includegraphics[width=0.5\linewidth]{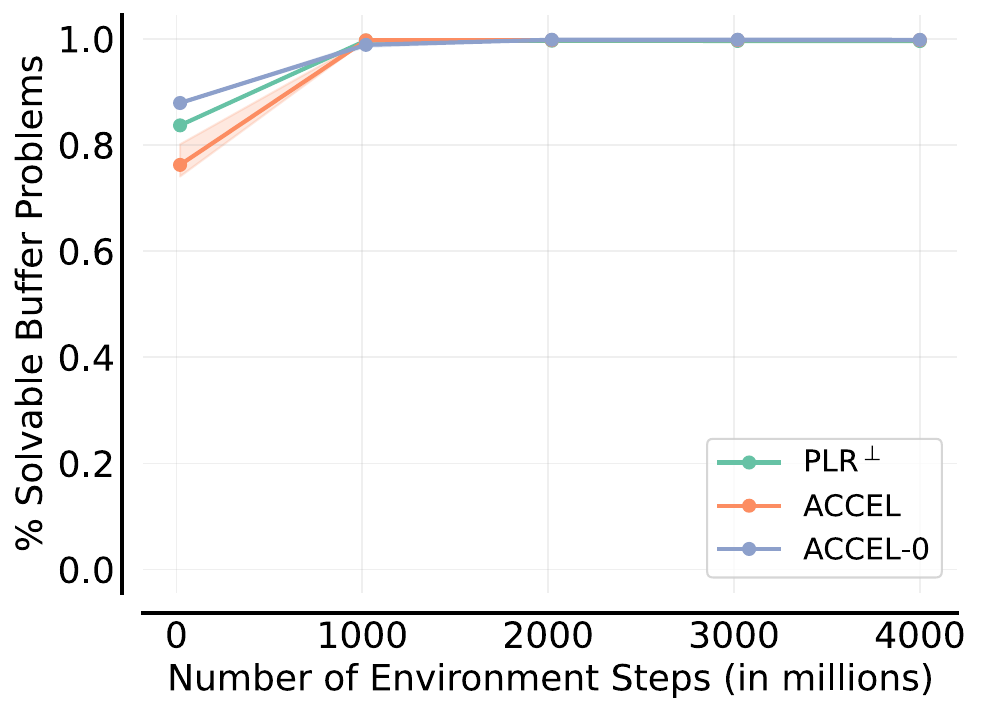}
	\caption{Fraction of solvable buffer problems throughout training in the \emph{level-conditioned} sampling setting.}
	\label{fig:solvability_over_time_level_cond}
\end{figure}

\subsection{Extended Task Sampling Ablation Results}
\label{app:experiments_task_sampling_ablation_results}
\cref{fig:ued_cvar_and_test_set_rw} shows the performance of DR and \rplr in the setting where RM tasks are \emph{directed acyclic graphs} (DAGs) generated with the \emph{random walk-based} sampler. The maximum number of states is set to 6 (like for the sequential sampler), and the maximum number of paths from the initial state $\rminit$ to the accepting state $\rmacc$ is set to 2. We analyze the results with independent sampling and level-conditioned sampling. The CVaR of the solve rate shows that \rplr is far more robust than DR in the independent sampling setting; however, in the level-conditioned sampling setting, \rplr is less robust than DR, albeit they are close. In line with the results on the hand-designed set obtained with sequential sampling, \rplr outperforms DR in the independent sampling setting, which is the most challenging. However, we make two key observations: (i)~the performance is almost half of that obtained with sequential RMs, so increasing the training distribution complexity hinders performance, and (ii)~\rplr performs slightly worse than DR in the level-conditioned setting.

\begin{figure}
	\centering
	\begin{subfigure}[b]{0.49\linewidth}
		\includegraphics[width=\linewidth]{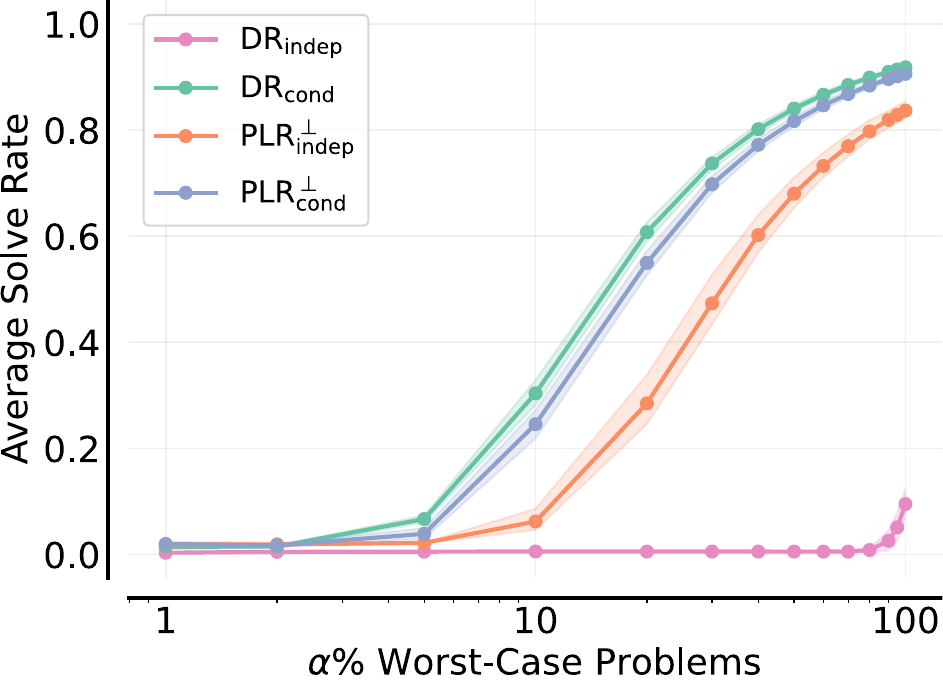}
		\caption{CVaR of the solve rate.}
	\end{subfigure}
	\begin{subfigure}[b]{0.49\linewidth}
		\includegraphics[width=\linewidth]{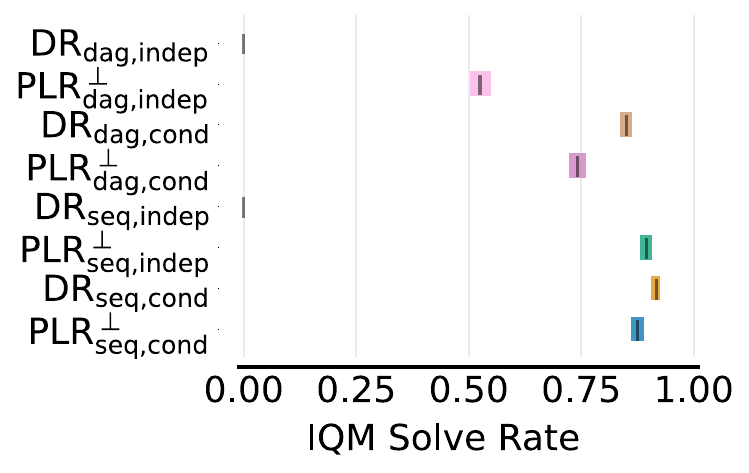}
		\caption{Performance on the hand-designed test set.}
	\end{subfigure}
	\caption{Performance of UED approaches using the \emph{random walk-based} task sampler.}
	\label{fig:ued_cvar_and_test_set_rw}
\end{figure}

\cref{fig:rw_sampling_curriculum} illustrates the evolution of problems in the buffer throughout training. Since, unlike sequential RMs, DAG RMs may consist of several paths from the initial to the accepting state, we analyze whether a curriculum is induced for two new metrics: the \emph{number of paths} and the \emph{average path length}. Both metrics are computed via a topological sort of the RM graph. In line with the sequential setting, we observe that a curriculum is induced over all the metrics; however, some change more drastically with level-conditioned sampling (the number of states and the average path length). \cref{fig:gen_prob_rw_indep_dr,fig:gen_prob_rw_indep_plr} show some problems generated throughout training with DR and \rplr, respectively. Negations to ensure mutual exclusivity (hence, determinism) are omitted for simplicity. While all problems are solvable for \rplr, not all paths are.

\begin{figure*}
	\centering
	\includegraphics[width=\linewidth]{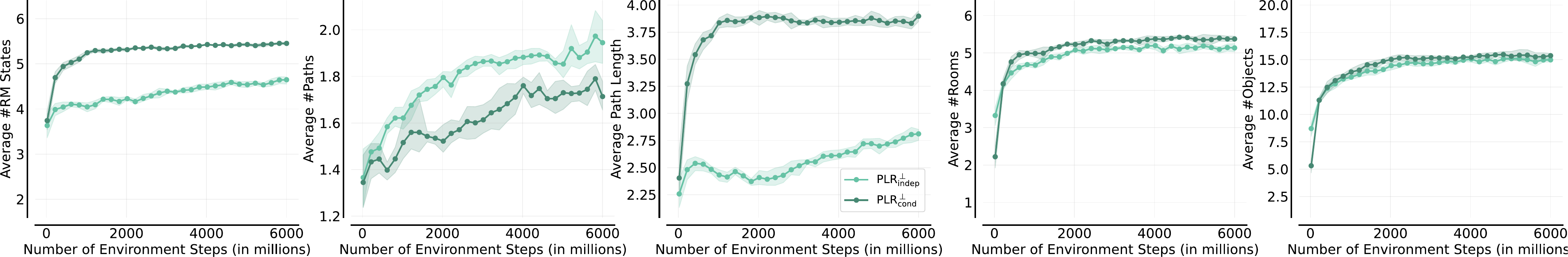}
	\caption{Emergent complexity metrics for problems in the buffer for \emph{independent} and \emph{level-conditioned} sampling settings using the \emph{random-walk based} task sampler to generate DAG RMs.}
	\label{fig:rw_sampling_curriculum}
\end{figure*}

\begin{figure*}
	\begin{subfigure}[b]{0.37\linewidth}
		\centering
		\includegraphics[height=4em]{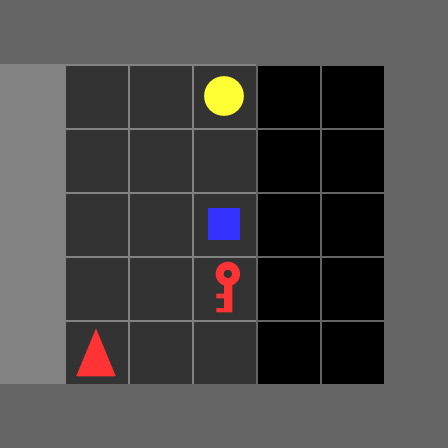}\\[1em]
		\scalebox{0.65}{
			\begin{tikzpicture}[shorten >=1pt,node distance=2.1cm,on grid,auto,every initial by arrow/.style ={-Latex}]
				\node[state,initial,initial text=] (u_0)   {$\rminit$};
				\node[state] (u_1) [below left =3cm of u_0]  {$\rmidx{1}$};
				\node[state] (u_2) [below right =3cm of u_1]  {$\rmidx{2}$};
				\node[state] (u_3) [below = of u_2]  {$\rmidx{3}$};
				\node[state] (u_4) [below left =3cm of u_3]  {$\rmidx{4}$};
				\node[state,accepting] (u_acc) [below right =3cm of u_4]  {$\rmacc$};
				
				\path[-Latex] (u_0) edge[bend right] node[in place, pos=0.6] {\small$\langle\text{\icnexthor{\iccolor{\icsquare}{icpurple}}{\icnocolor{\icdoorclosed}}},0\rangle$} (u_1);
				\path[-Latex] (u_1) edge[bend right] node[in place,pos=0.3] {\small$\langle\text{\icnexthor{\iccolor{\iccircle}{icyellow}}{\iccolor{\icdooropen}{icgreen}}},0\rangle$} (u_2);
				\path[-Latex] (u_0) edge node[in place] {\small$\langle\text{\icnexthor{\iccolor{\icsquare}{icgray}}{\iccolor{\ickey}{icgreen}}},0\rangle$} (u_2);
				\path[-Latex] (u_2) edge node[in place] {\small$\langle\text{\icnexthor{\iccolor{\iccircle}{icblue}}{\iccolor{\icdooropen}{icpurple}}},0\rangle$} (u_3);
				\path[-Latex] (u_3) edge[bend right] node[in place, pos=0.6] {\small$\langle\text{\icnexthor{\iccolor{\ickey}{icred}}{\iccolor{\icdooropen}{icyellow}}},0\rangle$} (u_4);
				\path[-Latex] (u_4) edge[bend right] node[in place, pos=0.3] {\small$\langle\text{\icnexthor{\iccolor{\ickey}{icpurple}}{\iccolor{\icdoorlocked}{icred}}},1\rangle$} (u_acc);
				\path[-Latex] (u_3) edge node[in place] {\small$\langle\text{\icnexthor{\iccolor{\icsquare}{icpurple}}{\iccolor{\icdoorlocked}{icgray}}},1\rangle$} (u_acc);
			\end{tikzpicture}
		}
		
		\caption{\num{2e9}}
	\end{subfigure}
	\begin{subfigure}[b]{0.245\linewidth}
		\centering
		\includegraphics[height=7.5em]{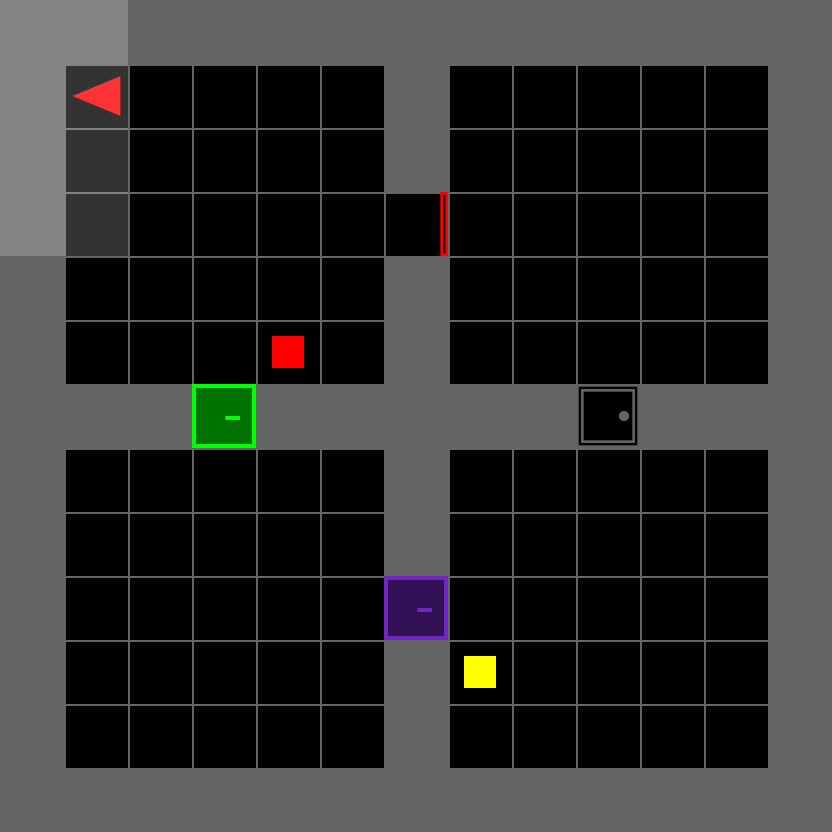}\\[1em]
		\scalebox{0.65}{
			\begin{tikzpicture}[shorten >=1pt,node distance=2.1cm,on grid,auto,every initial by arrow/.style ={-Latex}]
				\node[state,initial,initial text=] (u_0)   {$\rminit$};
				\node[state,accepting] (u_acc) [below = of u_0]  {$\rmacc$};
				
				\path[-Latex] (u_0) edge node[in place, pos=0.5] {\small$\langle\text{\icnexthor{\iccolor{\iccircle}{icred}}{\iccolor{\ickey}{icgray}}},1\rangle$} (u_acc);
			\end{tikzpicture}
		}
		\caption{\num{4e9}}
	\end{subfigure}
	\begin{subfigure}[b]{0.37\linewidth}
		\centering
		\includegraphics[height=7.5em]{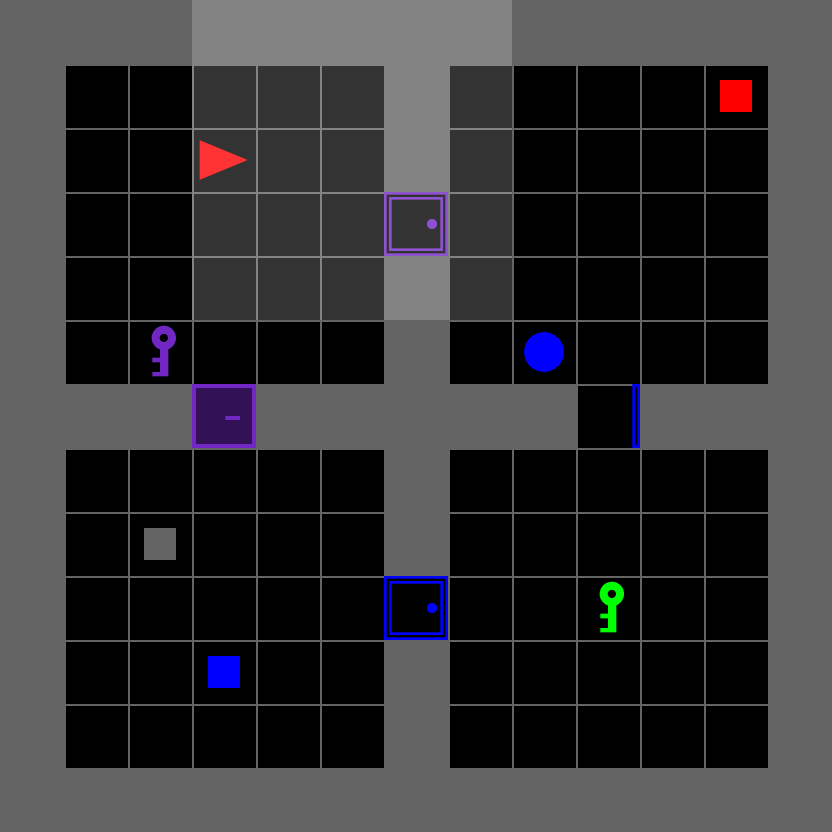}\\[1em]
		\scalebox{0.65}{
			\begin{tikzpicture}[shorten >=1pt,node distance=2.1cm,on grid,auto,every initial by arrow/.style ={-Latex}]
				\node[state,initial,initial text=] (u_0)   {$\rminit$};
				\node[state] (u_1) [below left =3cm of u_0]  {$\rmidx{1}$};
				\node[state] (u_2) [below = of u_1]  {$\rmidx{2}$};
				\node[state] (u_4) [below right =3cm of u_2]  {$\rmidx{4}$};
				\node[state] (u_3) [below =3.25cm of u_0]  {$\rmidx{3}$};
				\node[state,accepting] (u_acc) [below = of u_4]  {$\rmacc$};
				
				\path[-Latex] (u_0) edge[bend right] node[in place, pos=0.6] {\small$\langle\text{\icnexthor{\iccolor{\icsquare}{icgray}}{\icnocolor{\icdoorlocked}}},0\rangle$} (u_1);
				\path[-Latex] (u_1) edge node[in place] {\small$\langle\text{\icnexthor{\iccolor{\icsquare}{icblue}}{\icnocolor{\icdoorclosed}}},0\rangle$} (u_2);
				\path[-Latex] (u_2) edge[bend right] node[in place,pos=0.3] {\small$\langle\text{\icnexthor{\iccolor{\icsquare}{icgray}}{\iccolor{\icdoorclosed}{icpurple}}},0\rangle$} (u_4);
				\path[-Latex] (u_4) edge node[in place] {\small$\langle\text{\icnexthor{\iccolor{\iccircle}{icyellow}}{\icnocolor{\icdoorlocked}}},1\rangle$} (u_acc);
				\path[-Latex] (u_0) edge node[in place] {\small$\langle\text{\icnexthor{\iccolor{\iccircle}{icpurple}}{\iccolor{\iccircle}{icpurple}}},0\rangle$} (u_3);
				\path[-Latex] (u_3) edge node[in place] {\small$\langle\text{\icnexthor{\iccolor{\icsquare}{icgreen}}{\iccolor{\icdooropen}{icgreen}}},0\rangle$} (u_4);
			\end{tikzpicture}
		}
		\caption{\num{6e9}}
	\end{subfigure}
	\caption{Samples of DR-generated problems at different points in time (in number of environment steps) using the \emph{random-walk based} sampler.}
	\label{fig:gen_prob_rw_indep_dr}
\end{figure*}

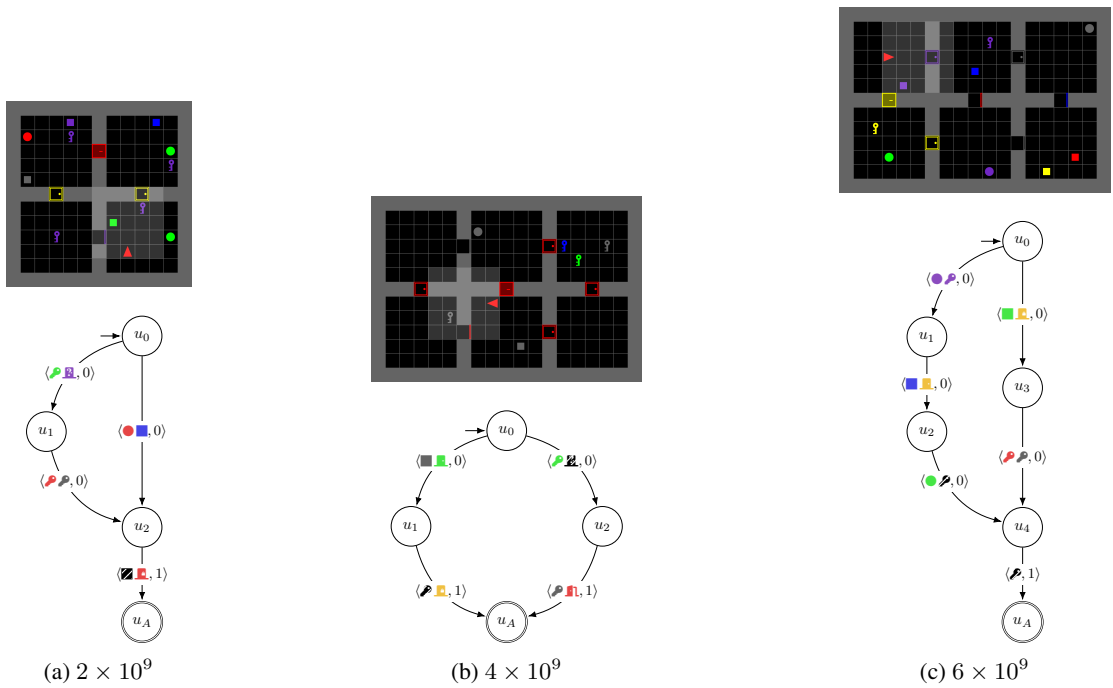
\begin{figure*}
	\begin{subfigure}[b]{0.3\linewidth}
		\centering
		\includegraphics[height=7em]{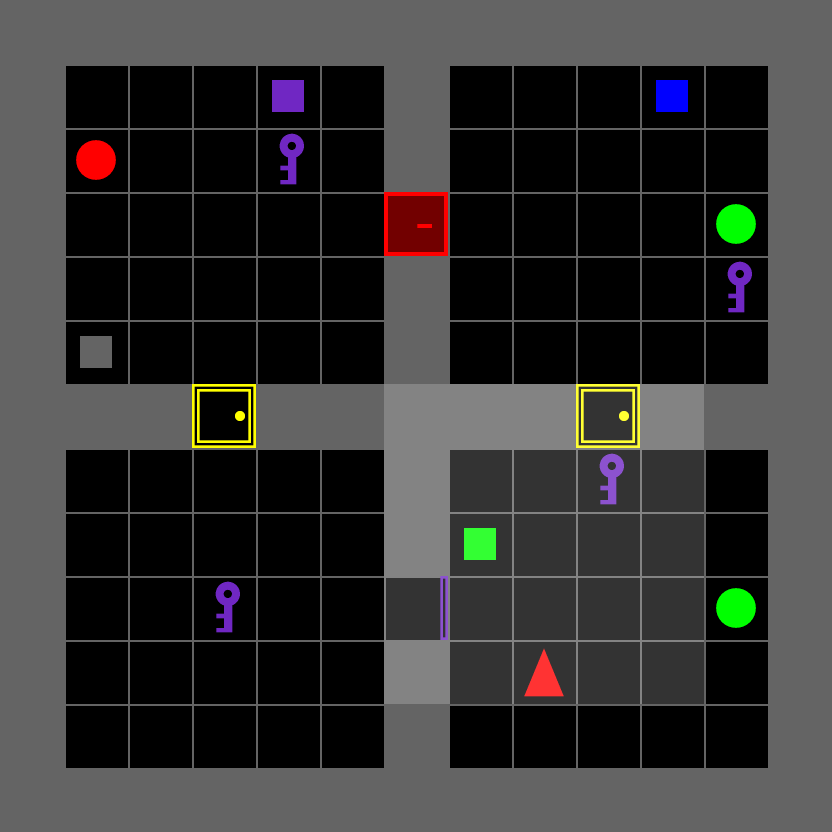}\\[1em]
		\scalebox{0.6}{
			\begin{tikzpicture}[shorten >=1pt,node distance=2.1cm,on grid,auto,every initial by arrow/.style ={-Latex}]
				\node[state,initial,initial text=] (u_0)   {$\rminit$};
				\node[state] (u_1) [below left =3cm of u_0]  {$\rmidx{1}$};
				\node[state] (u_2) [below right =3cm of u_1]  {$\rmidx{2}$};
				\node[state,accepting] (u_acc) [below = of u_2]  {$\rmacc$};
				
				\path[-Latex] (u_0) edge[bend right] node[in place, pos=0.6] {\small$\langle\text{\icnexthor{\iccolor{\ickey}{icgreen}}{\iccolor{\icdoor}{icpurple}}},0\rangle$} (u_1);
				\path[-Latex] (u_1) edge[bend right] node[in place,pos=0.3] {\small$\langle\text{\icnexthor{\iccolor{\ickey}{icred}}{\iccolor{\ickey}{icgray}}},0\rangle$} (u_2);
				\path[-Latex] (u_0) edge node[in place] {\small$\langle\text{\icnexthor{\iccolor{\iccircle}{icred}}{\iccolor{\icsquare}{icblue}}},0\rangle$} (u_2);
				\path[-Latex] (u_2) edge node[in place] {\small$\langle\text{\icnexthor{\icnocolor{\icsquare}}{\iccolor{\icdoorlocked}{icred}}},1\rangle$} (u_acc);
			\end{tikzpicture}
		}
		\caption{\num{2e9}}
	\end{subfigure}
	\begin{subfigure}[b]{0.3\linewidth}
		\centering
		\includegraphics[height=7em]{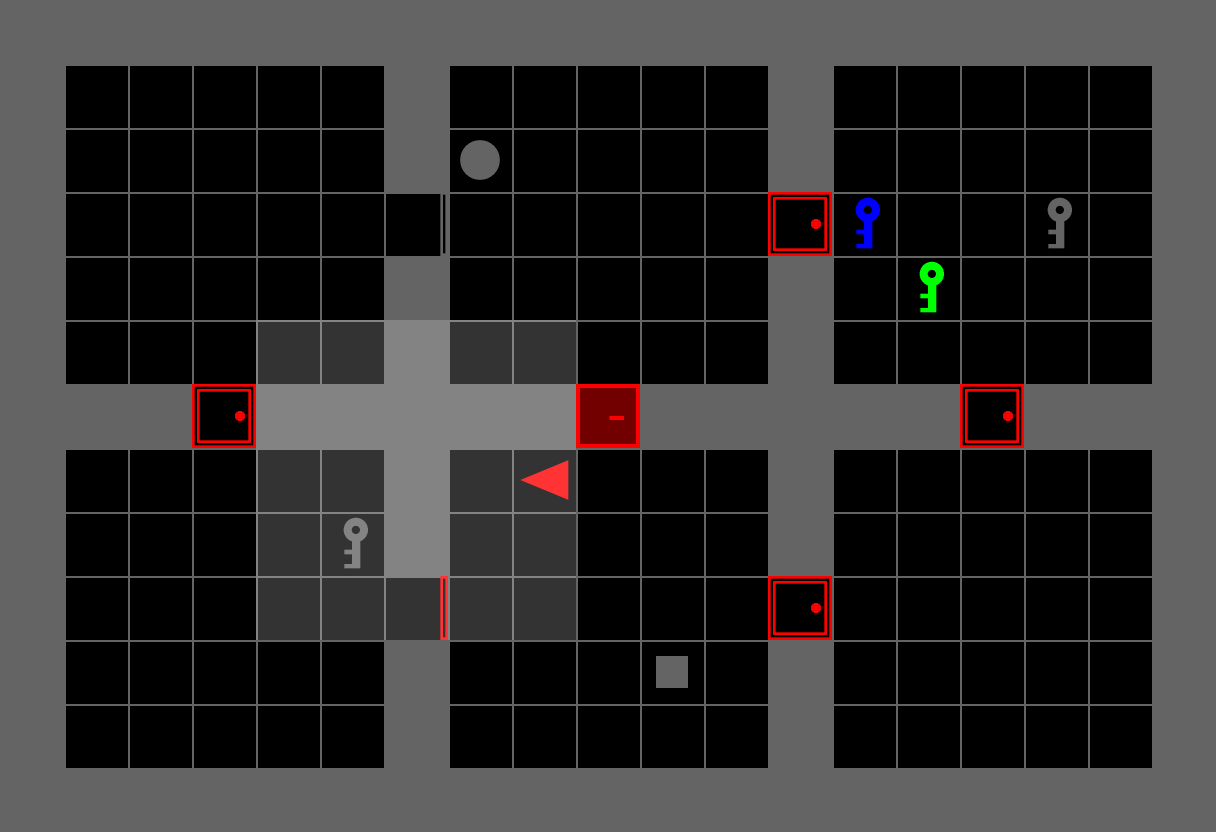}\\[1em]
		\scalebox{0.6}{
			\begin{tikzpicture}[shorten >=1pt,node distance=2.1cm,on grid,auto,every initial by arrow/.style ={-Latex}]
				\node[state,initial,initial text=] (u_0)   {$\rminit$};
				\node[state] (u_1) [below left =3cm of u_0]  {$\rmidx{1}$};
				\node[state] (u_2) [below right =3cm of u_0]  {$\rmidx{2}$};
				\node[state,accepting] (u_acc) [below right =3cm of u_1]  {$\rmacc$};
				
				\path[-Latex] (u_0) edge[bend right] node[in place] {\small$\langle\text{\icnexthor{\iccolor{\icsquare}{icgray}}{\iccolor{\icdoorclosed}{icgreen}}},0\rangle$} (u_1);
				\path[-Latex] (u_0) edge[bend left] node[in place] {\small$\langle\text{\icnexthor{\iccolor{\ickey}{icgreen}}{\icnocolor{\icdoor}}},0\rangle$} (u_2);
				\path[-Latex] (u_1) edge[bend right] node[in place] {\small$\langle\text{\icnexthor{\icnocolor{\ickey}}{\iccolor{\icdoorlocked}{icyellow}}},1\rangle$} (u_acc);
				\path[-Latex] (u_2) edge[bend left] node[in place] {\small$\langle\text{\icnexthor{\iccolor{\ickey}{icgray}}{\iccolor{\icdooropen}{icred}}},1\rangle$} (u_acc);
			\end{tikzpicture}
		}
		\caption{\num{4e9}}
	\end{subfigure}
	\begin{subfigure}[b]{0.39\linewidth}
		\centering
		\includegraphics[height=7em]{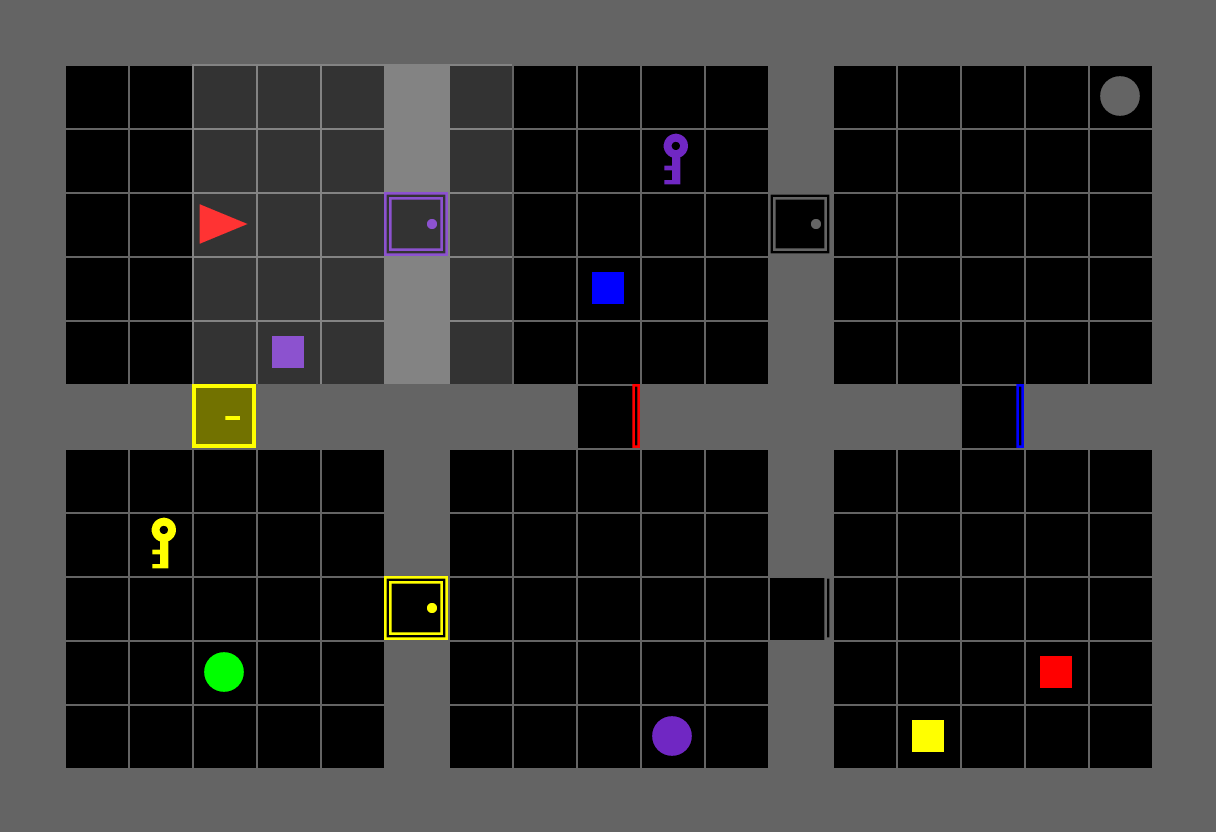}\\[1em]
		\scalebox{0.6}{
			\begin{tikzpicture}[shorten >=1pt,node distance=2.1cm,on grid,auto,every initial by arrow/.style ={-Latex}]
				\node[state,initial,initial text=] (u_0)   {$\rminit$};
				\node[state] (u_1) [below left =3cm of u_0]  {$\rmidx{1}$};
				\node[state] (u_2) [below = of u_1]  {$\rmidx{2}$};
				\node[state] (u_4) [below right =3cm of u_2]  {$\rmidx{4}$};
				\node[state] (u_3) [below =3.25cm of u_0]  {$\rmidx{3}$};
				\node[state,accepting] (u_acc) [below = of u_4]  {$\rmacc$};
				
				\path[-Latex] (u_0) edge[bend right] node[in place, pos=0.6] {\small$\langle\text{\icnexthor{\iccolor{\iccircle}{icpurple}}{\iccolor{\ickey}{icpurple}}},0\rangle$} (u_1);
				\path[-Latex] (u_1) edge node[in place] {\small$\langle\text{\icnexthor{\iccolor{\icsquare}{icblue}}{\iccolor{\icdoorclosed}{icyellow}}},0\rangle$} (u_2);
				\path[-Latex] (u_2) edge[bend right] node[in place,pos=0.3] {\small$\langle\text{\icnexthor{\iccolor{\iccircle}{icgreen}}{\icnocolor{\ickey}}},0\rangle$} (u_4);
				\path[-Latex] (u_4) edge node[in place] {\small$\langle\icnocolor{\ickey},1\rangle$} (u_acc);
				\path[-Latex] (u_0) edge node[in place] {\small$\langle\text{\icnexthor{\iccolor{\icsquare}{icgreen}}{\iccolor{\icdoorlocked}{icyellow}}},0\rangle$} (u_3);
				\path[-Latex] (u_3) edge node[in place] {\small$\langle\text{\icnexthor{\iccolor{\ickey}{icred}}{\iccolor{\ickey}{icgray}}},0\rangle$} (u_4);
			\end{tikzpicture}
		}
		\caption{\num{6e9}}
	\end{subfigure}
	\caption{Samples of \rplr-generated problems at different points in time (in number of environment steps) using the \emph{random-walk based} sampler.}
	\label{fig:gen_prob_rw_indep_plr}
\end{figure*}

\cref{fig:solvability_over_time_rw} shows the fraction of buffer problems that are solvable over time. As in the sequential RM sampling case, \rplr curates a buffer that mostly contains solvable problems.

\begin{figure}
	\centering
	\includegraphics[width=0.5\linewidth]{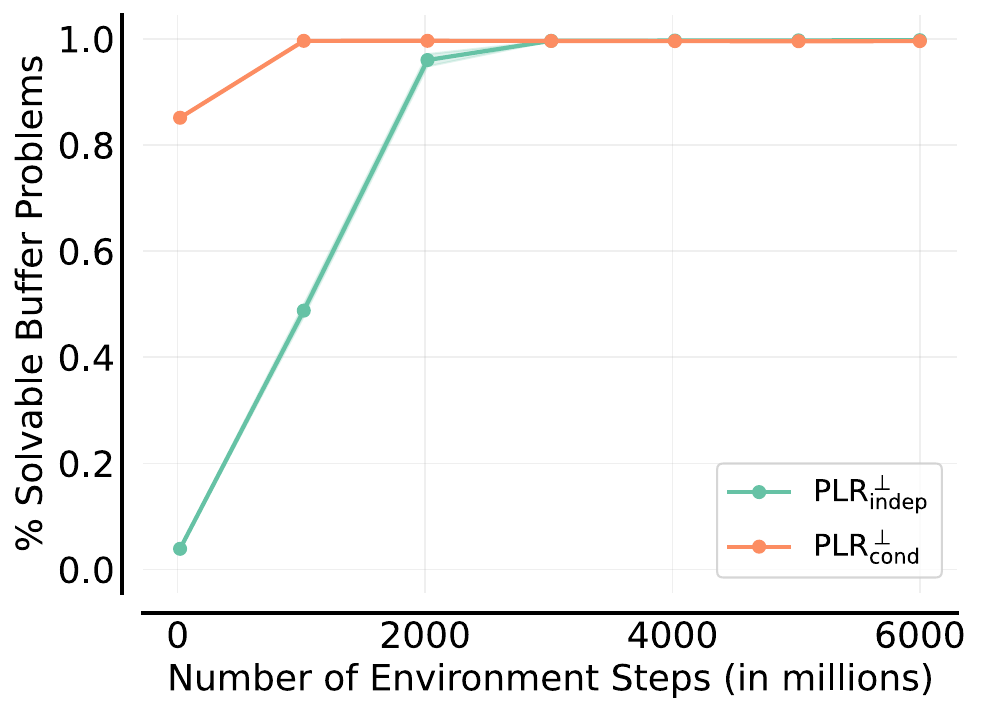}
	\caption{Fraction of solvable buffer problems throughout training using the \emph{random walk-based} task sampler.}
	\label{fig:solvability_over_time_rw}
\end{figure}

\subsection{Extended Mutation Ablation Results}
\label{app:experiments_mutation_ablation_results}
We analyze how performing ablations on the \emph{applicable edit types} and the \emph{edit sequence length} changes the performance of ACCEL and ACCEL-0. \cref{fig:mutation_ablations_types_accel} shows the performance of ACCEL for different combinations of level (L), task (T), and hindsight (H) edits. By default, we perform a combination of all such edits (L+T+H). Hindsight edits are not tested in isolation since they are not always applicable, whereas at least one level/task edit is always applicable. We observe that combining level and task edits sensibly improves performance over the same edits performed in isolation. Hindsight edits enhance performance when combined with level edits alone; however, in general, performance changes induced by hindsight edits seem minimal. As previously observed in \cref{app:experiments_main_results}, hindsight edits have little presence in the buffers.

\cref{fig:mutation_ablations_types_accel_0} illustrates the performance of ACCEL-0 with (default) and without hindsight edits. In this case, we do not ablate level and task edits since they are key to building increasingly complex problems from the sampled ones (one room, one key, one transition). In line with the previous point, hindsight edits do not sensibly improve performance.

\cref{fig:mutation_ablations_edit_seq} shows how performing fewer (1, 3) or more edits (20) than in our default setting changes ACCEL variants' performance. Our default setting uniformly samples between 7 and 10 edits. Decreasing the number of edits severely hinders the performance of ACCEL and ACCEL-0, especially when a single edit is performed. Increasing the number of edits to 20 has barely any effect on performance.

\begin{figure}
	\centering
	\begin{subfigure}[b]{0.32\linewidth}
		\includegraphics[width=\linewidth]{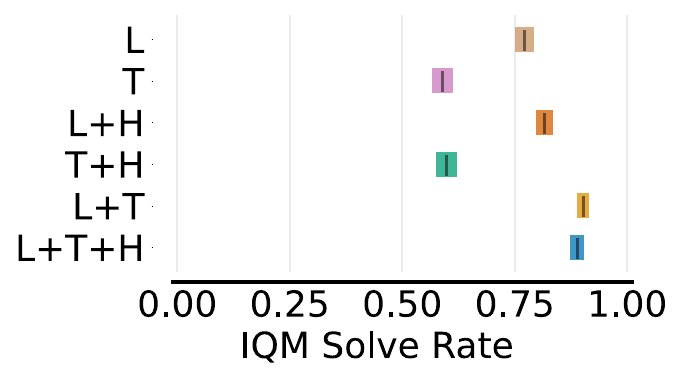}
		\caption{Edit type ablations in ACCEL.}
		\label{fig:mutation_ablations_types_accel}
	\end{subfigure}
	\begin{subfigure}[b]{0.32\linewidth}
		\includegraphics[width=\linewidth]{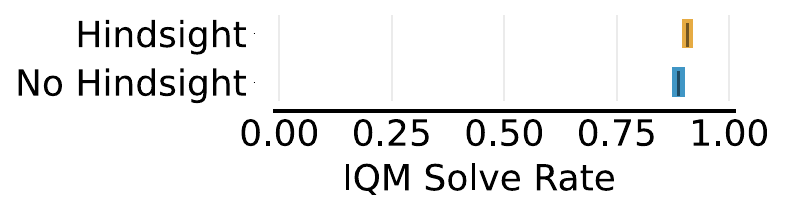}
		\caption{Edit type ablations in ACCEL-0.} 
		\label{fig:mutation_ablations_types_accel_0}
	\end{subfigure}
	\begin{subfigure}[b]{0.32\linewidth}
		\includegraphics[width=\linewidth]{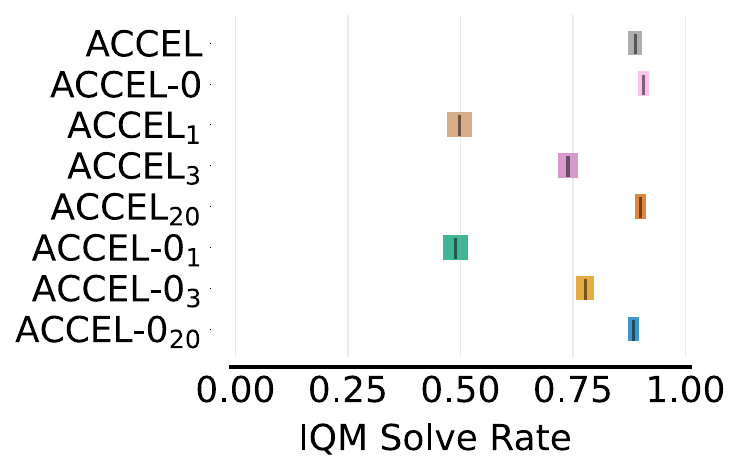}
		\caption{Edit sequence length ablation.}        
		\label{fig:mutation_ablations_edit_seq}
	\end{subfigure}
	\caption{ACCEL and ACCEL-0 performance for ablations on the applicable edit types and length of the edit sequence.}
	\label{fig:mutation_ablations}
\end{figure}

\subsection{Task-Conditioning Ablation Results}
We analyze how the behavior of \rplr changes for some ablations on task-conditioning (see \cref{app:architecture}). The ablations are the following:
\begin{description}
	\item[Vanilla] Condition the policy on the index of the current RM state rather than conditioning on an RM's graph embedding.
	\item[Myopic] Still conditions on the graph embedding, but the GCN has a single layer; hence, each RM state embedding only aggregates information from its immediate neighbors. By default, our GCN has five layers.
	\item[Domain Independent Embeddings (D.I. Embed.)] Instead of exploiting domain-specific knowledge to build the literal embeddings, each literal (proposition or its negation) is embedded differently.
\end{description}
\cref{fig:task_conditioning_ablations} shows the results. We make the following observations. First, the graph embeddings provide a substantial benefit with respect to the \emph{vanilla} embeddings since the latter cannot generalize across different tasks (the RM state index tells nothing about the task being performed). Second, the \emph{myopic} ablation performs close to the default setting, suggesting that the latter has converged to a myopic strategy. Indeed, the performance shown for the \textsc{Myopic} problem (see \cref{fig:test_problem_subset_performance}) is an additional indication: a non-myopic agent could consistently solve the problem. Third, the domain-dependent literal embeddings help improve performance with respect to the domain-agnostic ones.

\begin{figure}
	\centering
	\includegraphics[width=0.5\linewidth]{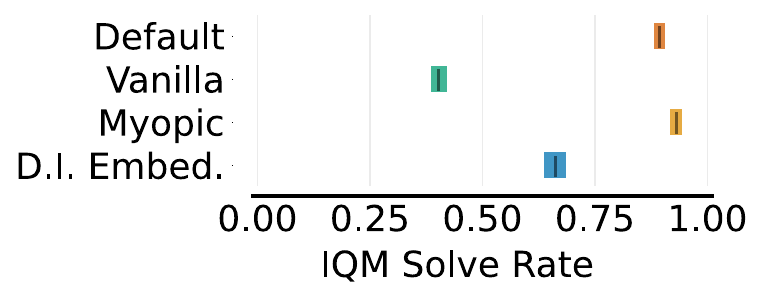}
	\caption{\rplr performance after applying different task-conditioning ablations.}
	\label{fig:task_conditioning_ablations}
\end{figure}

\subsection{Scoring Function Ablation Results}
\label{app:experiments_scoring_fn_ablation_results}
\cref{fig:score_fn_ablation} compares the performance of \rplr, ACCEL and ACCEL-0 using two scoring functions: MaxMC (default) and the \emph{positive value loss}~\cite[PVL;][]{JiangDPFGR21}. MaxMC induces a significantly better performance than PVL for both independent and level-conditioned sampling. This suggests that choosing an appropriate scoring function is key for final performance, or that PVL may require hyperparameter tuning. Performing an in-depth analysis of how scoring functions induce different training distributions is an interesting venue for future work. Along this path, \citet{RutherfordBWLHF24} show that MaxMC and PVL align with success rate rather than actual regret.

\begin{figure}
	\begin{subfigure}[b]{0.5\linewidth}
		\includegraphics[width=\linewidth]{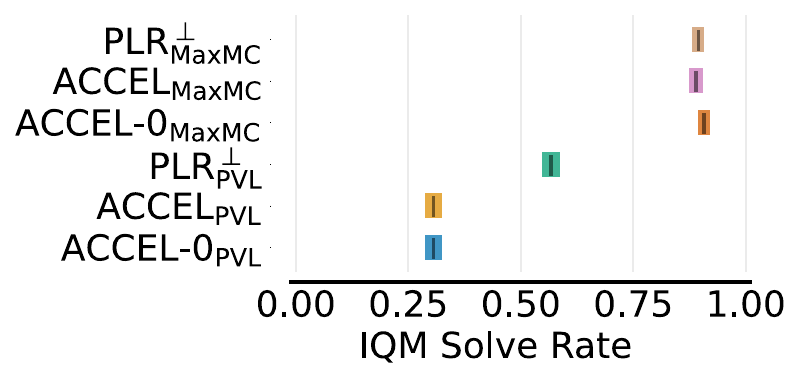}
		\caption{Independent sampling.}
	\end{subfigure}
	\begin{subfigure}[b]{0.5\linewidth}
		\includegraphics[width=\linewidth]{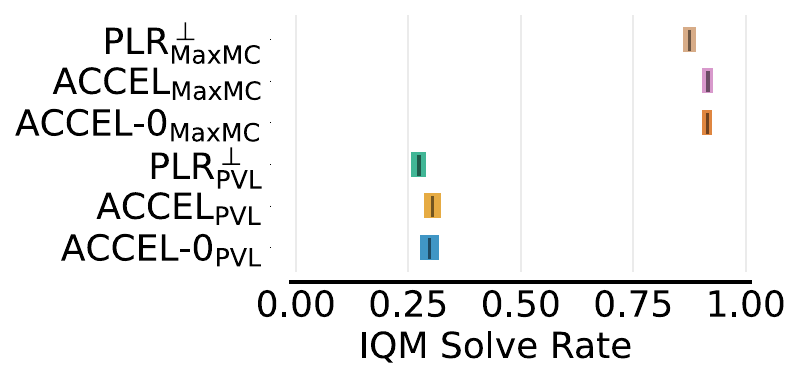}
		\caption{Level-conditioned sampling.}
	\end{subfigure}
	\caption{Performance of \rplr, ACCEL, and ACCEL-0 using the MaxMC (default) and PVL scoring functions.}
	\label{fig:score_fn_ablation}
\end{figure}

\subsection{Individual Problem Results}
\label{app:experiments_per_prob_results}
\cref{tab:per_problem_results_seq_indep,tab:per_problem_results_seq_cond,tab:per_problem_results_rw} report the average solve rate for each of the hand-designed problems in our proposed evaluation set across different task samplers (sequential, random-walk based) and problem sampling strategies (independent, level-conditioned). The problems originally shown in \cref{fig:test_problem_subset_performance} are \textsc{Myopic} ($\mathtt{myopic}$), \textsc{Patrol} ($\mathtt{patrol\_4r\_full\_spec}$), and \textsc{Choice} ($\mathtt{choice\text{-}choice\_three\text{-}next\_to\_square\_b}$).

{\small
\begin{longtable}{lrrrr}
		\label{tab:per_problem_results_seq_indep} \\
		\toprule
		& \multicolumn{1}{c}{DR} & \multicolumn{1}{c}{PLR$^\bot$} & \multicolumn{1}{c}{ACCEL} & \multicolumn{1}{c}{ACCEL-0} \\
		Problem &  &  &  &  \\
		\midrule
		\endfirsthead
		\toprule
		& \multicolumn{1}{c}{DR} & \multicolumn{1}{c}{PLR$^\bot$} & \multicolumn{1}{c}{ACCEL} & \multicolumn{1}{c}{ACCEL-0} \\
		\midrule
		\endhead
		\midrule\\[-1.5em]
		\caption{Solve rates and 95\% CIs (in brackets) for hand-designed problems with \emph{sequential} task sampling and \emph{independent} problem sampling (continued on next page).}
		\endfoot
		\bottomrule\\[-1.5em]
		\caption{Solve rates and 95\% CIs (in brackets) for hand-designed problems with \emph{sequential} task sampling and \emph{independent} problem sampling.}
		\endlastfoot
		$\mathtt{1r\text{-}corner\_balls\text{-}back\_n\_forth}$ & 0.0 (0.0, 0.0) & 1.0 (1.0, 1.0) & 1.0 (1.0, 1.0) & 0.94 (0.9, 0.98) \\
		$\mathtt{1r\text{-}corner\_balls\text{-}criss\_cross}$ & 0.0 (0.0, 0.0) & 1.0 (1.0, 1.0) & 1.0 (1.0, 1.0) & 1.0 (1.0, 1.0) \\
		$\mathtt{1r\text{-}corner\_balls\text{-}move\_purple\_key}$ & 0.0 (0.0, 0.0) & 1.0 (1.0, 1.0) & 0.9 (0.6, 1.0) & 1.0 (1.0, 1.0) \\
		$\mathtt{1r\text{-}corner\_balls\text{-}withholding}$ & 0.06 (0.0, 0.24) & 1.0 (1.0, 1.0) & 1.0 (1.0, 1.0) & 1.0 (1.0, 1.0) \\
		$\mathtt{1r\text{-}item\_cluster\text{-}all\_keys}$ & 0.0 (0.0, 0.0) & 1.0 (1.0, 1.0) & 1.0 (1.0, 1.0) & 1.0 (1.0, 1.0) \\
		$\mathtt{1r\text{-}item\_cluster\text{-}arrangement}$ & 0.0 (0.0, 0.0) & 1.0 (1.0, 1.0) & 0.9 (0.6, 1.0) & 0.96 (0.84, 1.0) \\
		$\mathtt{1r\text{-}item\_cluster\text{-}arrangement\_a}$ & 0.0 (0.0, 0.0) & 0.96 (0.92, 1.0) & 0.8 (0.44, 0.96) & 0.98 (0.92, 1.0) \\
		$\mathtt{1r\text{-}item\_cluster\text{-}arrangement\_b}$ & 0.0 (0.0, 0.0) & 0.98 (0.92, 1.0) & 0.76 (0.2, 1.0) & 0.92 (0.76, 1.0) \\
\end{longtable}}

{
	\small
	\begin{longtable}{lrrrr}
		\label{tab:per_problem_results_seq_cond} \\
		\toprule
		& \multicolumn{1}{c}{DR} & \multicolumn{1}{c}{PLR$^\bot$} & \multicolumn{1}{c}{ACCEL} & \multicolumn{1}{c}{ACCEL-0} \\
		Problem &  &  &  &  \\
		\midrule
		\endfirsthead
		\toprule
		& \multicolumn{1}{c}{DR} & \multicolumn{1}{c}{PLR$^\bot$} & \multicolumn{1}{c}{ACCEL} & \multicolumn{1}{c}{ACCEL-0} \\
		Problem &  &  &  &  \\
		\midrule
		\endhead
		\midrule\\[-1.5em]
		\caption{Solve rates and 95\% CIs (in brackets) for hand-designed problems with \emph{sequential} task sampling and \emph{level-conditioned} problem sampling (continued on next page).}
		\endfoot
		\bottomrule\\[-1.5em]
		\caption{Solve rates and 95\% CIs (in brackets) for hand-designed problems with \emph{sequential} task sampling and \emph{level-conditioned} problem sampling.}
		\endlastfoot
		$\mathtt{1r\text{-}corner\_balls\text{-}back\_n\_forth}$ & 1.0 (1.0, 1.0) & 1.0 (1.0, 1.0) & 1.0 (1.0, 1.0) & 1.0 (1.0, 1.0) \\
		$\mathtt{1r\text{-}corner\_balls\text{-}criss\_cross}$ & 1.0 (1.0, 1.0) & 1.0 (1.0, 1.0) & 1.0 (1.0, 1.0) & 1.0 (1.0, 1.0) \\
		$\mathtt{1r\text{-}corner\_balls\text{-}move\_purple\_key}$ & 1.0 (1.0, 1.0) & 1.0 (1.0, 1.0) & 1.0 (1.0, 1.0) & 1.0 (1.0, 1.0) \\
		$\mathtt{1r\text{-}corner\_balls\text{-}withholding}$ & 1.0 (1.0, 1.0) & 1.0 (1.0, 1.0) & 1.0 (1.0, 1.0) & 1.0 (1.0, 1.0) \\
		$\mathtt{1r\text{-}item\_cluster\text{-}all\_keys}$ & 1.0 (1.0, 1.0) & 1.0 (1.0, 1.0) & 1.0 (1.0, 1.0) & 1.0 (1.0, 1.0) \\
		$\mathtt{1r\text{-}item\_cluster\text{-}arrangement}$ & 1.0 (1.0, 1.0) & 0.88 (0.52, 1.0) & 1.0 (1.0, 1.0) & 0.96 (0.84, 1.0) \\
		$\mathtt{1r\text{-}item\_cluster\text{-}arrangement\_a}$ & 1.0 (1.0, 1.0) & 1.0 (1.0, 1.0) & 1.0 (1.0, 1.0) & 0.94 (0.84, 1.0) \\
		$\mathtt{1r\text{-}item\_cluster\text{-}arrangement\_b}$ & 1.0 (1.0, 1.0) & 0.98 (0.92, 1.0) & 1.0 (1.0, 1.0) & 0.96 (0.92, 1.0) \\
\end{longtable}
}

{
	\small
	\begin{longtable}{lrrrr}
		\label{tab:per_problem_results_rw} \\
		\toprule
		& \multicolumn{1}{c}{DR$_\text{indep}$} & \multicolumn{1}{c}{PLR$^\bot_\text{indep}$} & \multicolumn{1}{c}{DR$_\text{cond}$} & \multicolumn{1}{c}{PLR$^\bot_\text{cond}$} \\
		Problem &  &  &  &  \\
		\midrule
		\endfirsthead
		\toprule
		& \multicolumn{1}{c}{DR$_\text{indep}$} & \multicolumn{1}{c}{PLR$^\bot_\text{indep}$} & \multicolumn{1}{c}{DR$_\text{cond}$} & \multicolumn{1}{c}{PLR$^\bot_\text{cond}$} \\
		Problem &  &  &  &  \\
		\midrule
		\endhead\\[-1.5em]
		\midrule
		\caption{Solve rates and 95\% CIs (in brackets) for hand-designed problems with \emph{random walk-based} task sampling (continued on next page).}
		\endfoot
		\bottomrule\\[-1.5em]
		\caption{Solve rates and 95\% CIs (in brackets) for hand-designed problems with \emph{random walk-based} task sampling.}
		\endlastfoot
		$\mathtt{1r\text{-}corner\_balls\text{-}back\_n\_forth}$ & 0.0 (0.0, 0.0) & 0.98 (0.92, 1.0) & 1.0 (1.0, 1.0) & 1.0 (1.0, 1.0) \\
		$\mathtt{1r\text{-}corner\_balls\text{-}criss\_cross}$ & 0.0 (0.0, 0.0) & 1.0 (1.0, 1.0) & 1.0 (1.0, 1.0) & 1.0 (1.0, 1.0) \\
		$\mathtt{1r\text{-}corner\_balls\text{-}move\_purple\_key}$ & 0.0 (0.0, 0.0) & 1.0 (1.0, 1.0) & 1.0 (1.0, 1.0) & 1.0 (1.0, 1.0) \\
		$\mathtt{1r\text{-}corner\_balls\text{-}withholding}$ & 0.16 (0.0, 0.36) & 1.0 (1.0, 1.0) & 1.0 (1.0, 1.0) & 1.0 (1.0, 1.0) \\
		$\mathtt{1r\text{-}item\_cluster\text{-}all\_keys}$ & 0.0 (0.0, 0.0) & 1.0 (1.0, 1.0) & 1.0 (1.0, 1.0) & 1.0 (1.0, 1.0) \\
		$\mathtt{1r\text{-}item\_cluster\text{-}arrangement}$ & 0.12 (0.0, 0.48) & 1.0 (1.0, 1.0) & 1.0 (1.0, 1.0) & 1.0 (1.0, 1.0) \\
		$\mathtt{1r\text{-}item\_cluster\text{-}arrangement\_a}$ & 0.0 (0.0, 0.0) & 1.0 (1.0, 1.0) & 1.0 (1.0, 1.0) & 1.0 (1.0, 1.0) \\
		$\mathtt{1r\text{-}item\_cluster\text{-}arrangement\_b}$ & 0.0 (0.0, 0.0) & 1.0 (1.0, 1.0) & 1.0 (1.0, 1.0) & 1.0 (1.0, 1.0) \\
\end{longtable}
}
\fi

\end{document}